\documentclass[journal]{IEEEtran}
\usepackage[
            pdfstartview=FitH,
            CJKbookmarks=true,
            bookmarksnumbered=true,
            bookmarksopen=true,
            colorlinks, 
            pdfborder=001,   
            linkcolor=green,
            anchorcolor=green,
            citecolor=green
            ]{hyperref}
\usepackage{multirow}
\usepackage[pdftex]{graphicx}
\usepackage[cmex10]{amsmath}
\usepackage{cite}
\usepackage{url}
\usepackage{enumerate}
\usepackage{epstopdf}
\usepackage{subfigure}
\usepackage{extarrows}
\usepackage{amsfonts}

\def\ie{{\em i.e.}}
\def\eg{{\em e.g.}}
\def\etal{{\em et al.}}

\ifCLASSINFOpdf
\else
\fi

\begin{document}

\title{A New Evaluation Protocol and Benchmarking Results for Extendable Cross-media Retrieval}

\author{Ruoyu Liu,
        Yao Zhao, 
        Liang Zheng,
        Shikui Wei,
        and Yi Yang

\IEEEcompsocitemizethanks{
\IEEEcompsocthanksitem R. Liu and S. Wei are with the Institute of Information Science, Beijing Jiaotong University, Beijing Key Laboratory of Advanced Information Science and Network Technology, Beijing 100044, China. \protect\\
E-mail: 12112062@bjtu.edu.cn, shkwei@bjtu.edu.cn
\IEEEcompsocthanksitem Y. Zhao is with the State Key Laboratory of Rail Traffic Control and Safety, Beijing Jiaotong University, Beijing 100044, China, and the Institute of Information Science, Beijing Jiaotong University, Beijing 100044, China. \protect\\
E-mail: yzhao@bjtu.edu.cn
\IEEEcompsocthanksitem L. Zheng and Y. Yang are with the
Centre for Quantum Computation and Intelligent Systems, University of Technology at Sydney, NSW, Australia. \protect\\
E-mail: liangzheng06@gmail.com, yee.i.yang@gmail.com
}
}

\maketitle

\begin{abstract}
This paper proposes a new evaluation protocol for cross-media retrieval which better fits the real-word applications. Both image-text and text-image retrieval modes are considered. Traditionally, class labels in the training and testing sets are identical. That is, it is usually assumed that the query falls into some pre-defined classes. However, in practice, the content of a query image/text may vary extensively, and the retrieval system does not necessarily know in advance the class label of a query. Considering the inconsistency between the real-world applications and laboratory assumptions, we think that the existing protocol that works under identical train/test classes can be modified and improved.

This work is dedicated to addressing this problem by considering the protocol under an extendable scenario, \ie, the training and testing classes do not overlap. We provide extensive benchmarking results obtained by the existing protocol and the proposed new protocol on several commonly used datasets.  We demonstrate a noticeable performance drop when the testing classes are unseen during training. Additionally, a trivial solution, \ie, directly using the predicted class label for cross-media retrieval, is tested. We show that the trivial solution is very competitive in traditional non-extendable retrieval, but becomes less so under the new settings. The train/test split, evaluation code, and benchmarking results are publicly available on our website\footnote{Codes are released on our website: liuruoyu.github.io}.

\end{abstract}


\IEEEpeerreviewmaketitle

\section{Introduction and Related Work}
\label{sec:introduction}

This paper focuses on the cross-media retrieval between images and texts. In this task, given a query text/image, we aim to retrieve the relevant images/texts from the gallery (database). Since the two modalities are located in different feature spaces, the challenge of cross-media retrieval consists in the similarity measurement between the heterogeneous data. An effective solution learns a unified representation for different modalities so that the common distances can be employed for similarity measurement.

\textbf{Related Work.} The research of multimedia retrieval has two diverse branches: single-media retrieval and cross-media retrieval. The former branch, such as image retrieval \cite{zheng2014packing,zheng2016accurate,zheng2016sift} and video retrieval \cite{song2013effective}, uses the homogeneous queries to perform image-to-image or video-to-video retrieval. However, the query and gallery in cross-media retrieval are heterogeneous and their similarities cannot be directly measured.

The concept of cross-media retrieval is firstly defined by Wu \emph{et al.} \cite{wu2005understanding} and they also propose the earliest cross-media model: multimedia document (MMD). The media objects of different modalities that carry the 
same semantic (like the image, text and audio in the same web page) are collected together as an MMD. Then, the distance between two MMDs is calculated from the distances of the media objects in each modality. After \cite{wu2005understanding}, Yang \emph{et al.} propose a series of methods to tackle cross-media retrieval by using MMD \cite{zhuang2008mining,yang2008harmonizing,yang2009ranking,yang2010cross}. The shortcoming of MMD is that it is not very flexible, because it handles the set of component media objects as a whole. 

The main body of cross-media methods is based on learning the common representation. The milestone work is~\cite{rasiwasia2010new} proposed by Rasiwasia~\etal, which employs the canonical correlation analysis (CCA)~\cite{hotelling1936relations} and multi-class logistic regression to learn the descriptors for the heterogeneous data. Inspired by~\cite{rasiwasia2010new}, many approaches have been proposed to learn the common representations, which can be classified into two groups: real-valued and binary representations.

The real-valued representations map the heterogeneous data into a common continuous feature space. Shallow methods learn two linear functions~\cite{rasiwasia2010new,sharma2012generalized,wang2013learning} or simple nonlinear functions~\cite{gong2014multi,rasiwasia2014cluster} to maximize the correlations between the pairwise data or further improve feature discrimination by using the category labels. With the introduction of deep learning technique, deep networks have also been employed in cross-media retrieval, which learns more complex projections. The feasible networks include fully connected networks~\cite{andrew2013deep,wang2016learning}, convolutional neural networks~\cite{yan2015deep,wei2016cross,he2016cross}, recursive neural networks~\cite{socher2014grounded}, recurrent neural networks~\cite{kiros2014unifying}, auto-encoders~\cite{feng2014cross,vukotic2016bidirectional} and adversarial networks~\cite{park2016image}. These deep methods have shown their superiority in retrieval accuracy to the shallow methods.

The binary representations, on the other hand, map the heterogeneous data into a discrete space, where the entries of the features consist of two common values: $ \{0,1\} $. These methods are also called cross-media hashing, and they focus on large-scale retrieval. In this scenario, these methods use Hamming distance to accelerate the search process. Most of the binary methods are shallow models which relax the problem into a real-valued case~\cite{kumar2011learning,song2013inter,ding2014collective,zhang2014large,zhou2014latent,yu2014discriminative,xu2016discriminant} or optimize to learn the hash codes directly~\cite{bronstein2010data,zhen2012probabilistic,lin2015semantics}. Some recent works have also employed deep learning to learn better hash codes~\cite{masci2014multimodal,jiang2016deep,cao2016deep}.

There are some methods using deep networks to tackle cross-media retrieval from other perspectives. For example, \cite{frome2013devise,mao2014explain,karpathy2014deep,ma2015multimodal} train the networks to learn the similarities between the heterogeneous data directly, which can be viewed as extensions of metric learning. Several image caption methods also show their capability of performing cross-media retrieval~\cite{chen2015mind,karpathy2015deep}, since they use a part of their networks to embed images and texts into a common feature space.

Currently, a majority line of cross-media research relies on the class labels in training \cite{gong2014multi,wei2016cross,xu2016discriminant}. For example, Wei \emph{et al.} \cite{wei2016cross} fine-tune a convolutional neural network (CNN) on the text and image domains using the class supervision end extracts the softmax layer for retrieval. Gong \emph{et al.} \cite{gong2014multi} introduce supervision by treating the class labels as the third domain. On the end of performance evaluation, classic datasets, \eg, Wikipedia \cite{rasiwasia2010new} and NUS-WIDE \cite{chua2009nus}, define their ground-truths based on the category labels. They aim to search for texts/images belonging to the same class with the query image/text. Usually, the heterogeneous data is treated as a true match if it has at least one common category label with the query. This task is called \textbf{class(-level) retrieval} in this paper, and this type of  ground-truth is called the \textbf{class ground-truth}.

To evaluate the performance of cross-media retrieval, many datasets have been built or employed. Classical datasets are featured by the following aspects: they generally have two media types, \ie, images and texts, their data are labeled by several category labels, and they perform class retrieval using the class ground-truths. Wikipedia~\cite{rasiwasia2010new} is the first such dataset, and other datasets include NUS-WIDE~\cite{chua2009nus}, MIRFlickr25K~\cite{huiskes2008mir}, Pascal VOC (2007)~\cite{everingham2010pascal}, Web Queries~\cite{krapac2010improving} and Pascal Sentence~\cite{rashtchian2010collecting}. Most of the recent works use the datasets employed in the field of image captioning \cite{mao2014explain,karpathy2014deep,ma2015multimodal,chen2015mind,karpathy2015deep,yan2016image}. Such (image-sentence) datasets include Flickr8K~\cite{hodosh2013framing}, Flickr30K~\cite{young2014image}, MSCOCO~\cite{lin2014microsoft} and SBU~\cite{ordonez2011im2text}. These datasets define the true matches of a query as the heterogeneous data describing it. In spite of the popularity of these new datasets, it is still meaningful to re-evaluate some popular methods on the traditional datasets. One import reason is that their texts cover more types (article, tag, surrounding words and sentence). A recent specialized cross-media dataset, XMedia~\cite{zhai2014learning}, consists of five media types in total and still uses the class ground-truths.

In this paper, we propose a new evaluation protocol to re-evaluate the existing cross-media methods on\textbf{ the extendability to unseen classes}. This protocol is designed for \textbf{extendable cross-media retrieval}, and it can reflect the more realistic performance compared to the existing protocol.

\section{the Evaluation Protocol}
\label{sec:evaprotos}

\begin{figure}[t]
\begin{center}
\includegraphics[width=1.0\linewidth]{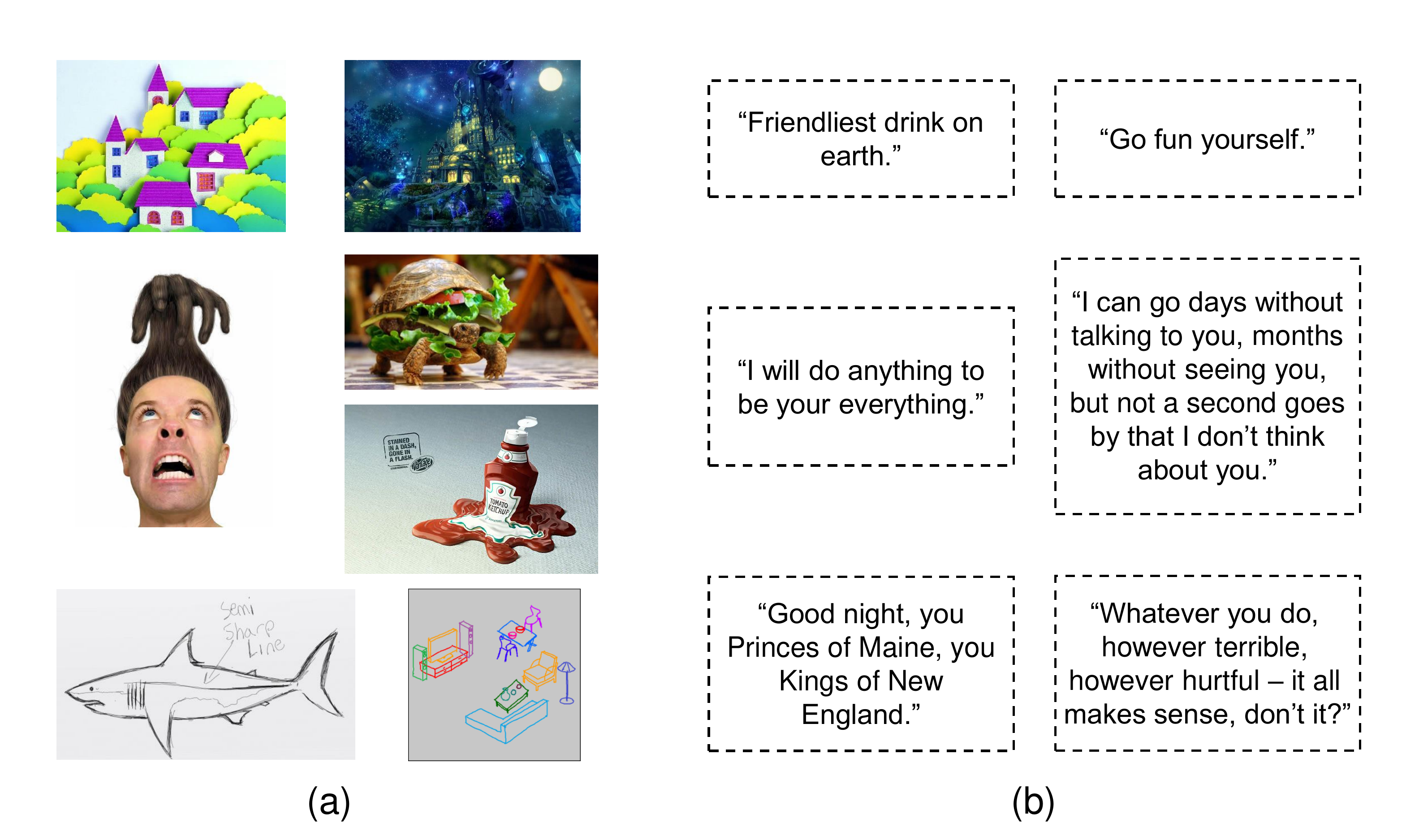}
\end{center}
	\caption{Some examples of (a) image and (b) text queries exhibiting various content. These images and texts are hard to be taken into consideration when training cross-media models.}
	\label{fig:QueryExample}
\end{figure}

In the cross-media retrieval community, the current mainstream methods employ the same set of classes in both the training and testing steps, \eg, the 10 classes in Wikipedia, the most frequent 10 classes in NUS-WIDE. This train/test protocol assumes that a query always belongs to one of the pre-defined classes. Yet, this assumption does not always hold in practice, because the query text/image may exhibit various content and it is challenging for the training process to take into account all the variety in query types. Some examples are illustrated in Fig. \ref{fig:QueryExample}. Moreover, in the field of learning to hash, Sablayrolles \emph{et al.} \cite{sablayrolles2016should} suggests that there exists a trivial solution to this problem: a classifier is trained for each class, class predictions are made for the query and the gallery items, and the retrieval process is equivalent to finding the relevant images with the same class predictions with the query; so performing an explicit hashing-based retrieval process does not seem necessary. It is therefore not well-grounded for an evaluation protocol to assume that the train/test data have the same set of classes.

Our main point is that \emph{cross-media retrieval can be in effect viewed as an extendable problem, in which the query class is ``unseen'' during training.} On the one hand, this setting meets closely with the reality. On the other hand, the extendable assumption has also been used by default in several other retrieval problems, such as generic instance retrieval \cite{zheng2016sift,radenovic2016cnn}, person re-identification \cite{zheng2015scalable, zheng2016person} and vehicle re-identification \cite{liu2016deep,liu2016large}. A model is usually learned on the training set and tested for the unseen query and gallery. Under this scenario, the effectiveness of previous learning methods should be re-evaluated; it is possible that a method that works well on the non-extendable problem may exhibit low generalization ability under the extendable setting. More insights need to be gained.

Another problem associated with the existing protocol is that the same data is used for training and gallery. This is potentially problematic because in practice, the gallery may be very large, and it is infeasible to label all the gallery data. As a consequence, for practical evaluation, our second point is that \emph{it would be best to separate the training set and the testing set (composed of the gallery and query).}

Considering the above two points, \ie, the extendable nature and the separation of train/test splits, we proposes a new evaluation protocol on the currently available datasets. 

\textbf{Dataset Splitting.} Motivated by \cite{sablayrolles2016should}, we propose a new train/test splitting for cross-media retrieval. It separates the training and testing data so that the training and testing sets each has 50\% of the categories, \ie, there is no class overlap between them. Models are learned on the training classes only and are directly tested on the testing set (gallery+query). The new splitting is in accordance with practical usage. Under this circumstance, the trivial solution does not produce competitive performance (to be shown in Section \ref{sec:ExpResults}).

\begin{figure}[t]
\begin{center}
\includegraphics[width=1.0\linewidth]{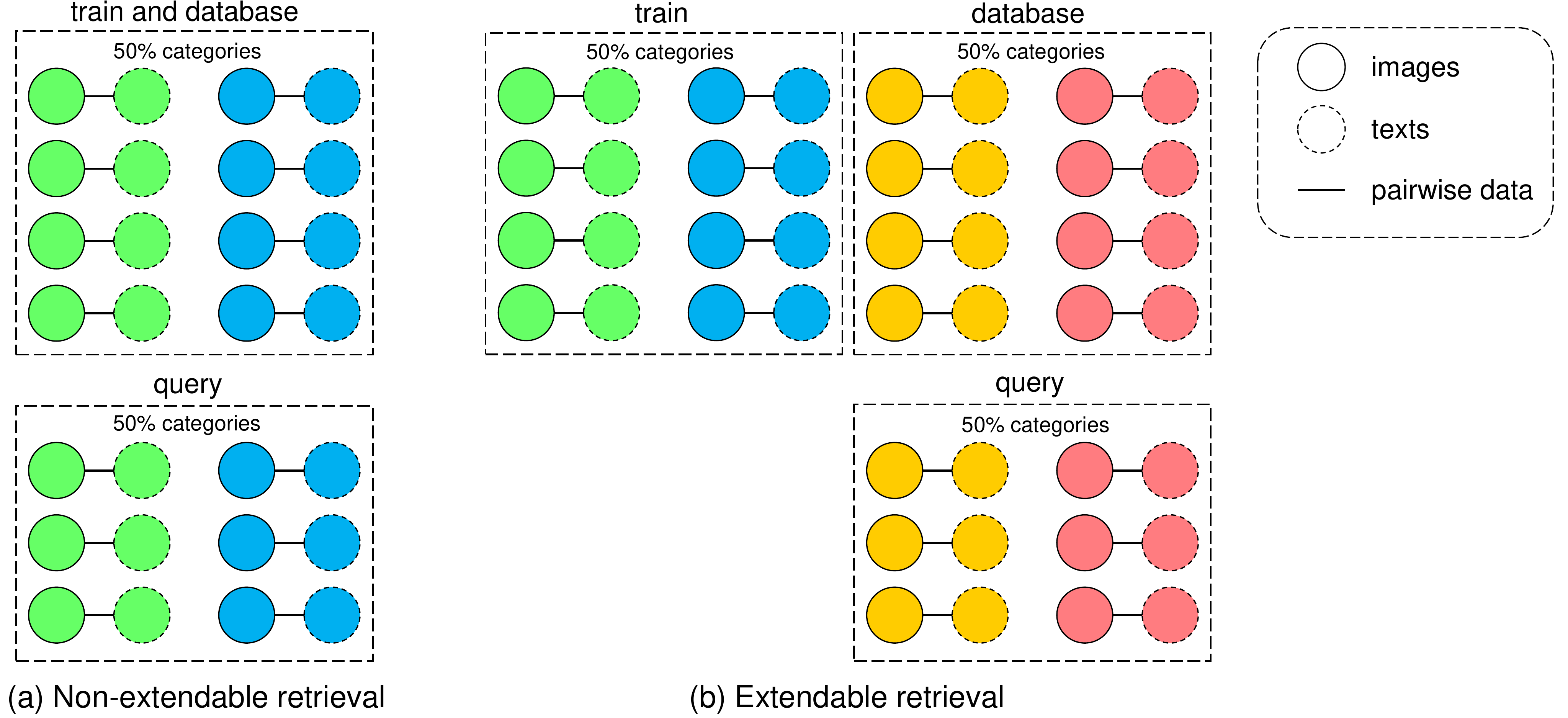}
\end{center}
	\caption{Train/test splitting of two retrieval tasks. (a) Non-extendable retrieval: the training and testing data (query and database) are from the same $ 50\% $ classes, the database subset is used for training. (b) Extendable retrieval: it uses the same training data in traditional retrieval, and uses the data of the other $ 50\% $ classes as the testing data (query and database). Classes are best viewed in color.
	}
	\label{fig:DatasetSeperation}
\end{figure}

Specifically, each dataset is separated into two parts: a training set consisting of the data from half of the categories, and a testing set consisting of the other half categories. Each set is further separated into two subsets: a database subset and a query subset. Using the four subsets, we evaluate cross-media retrieval on two tasks as illustrated in Fig. \ref{fig:DatasetSeperation}:

(1) \textbf{Non-extendable (non-XTD) retrieval}: In this task, we use the database subset of the training set to train the methods. Then, each sample in the query subset of the training set is used as a query to search its relevant heterogeneous data in the training subset of the training set. The train/text classes are identical, and it evaluates the performance of traditional non-xtd cross-media retrieval.

(2) \textbf{Extendable (XTD) retrieval}: In this task, we still use the database subset of the training set for training. But different from non-xtd retrieval, we use the samples of the query subset of the testing set as the queries to search their relevant heterogeneous data in the database subset of the testing set. There is no class overlap between the training and testing data, and in this task it evaluates the extendability to new datasets. 

To balance the influences of the different class splits, we shuffle the categories and use $ N $ folds to define $ N $ such class splits. The performances are averaged over the $ N $ folds to get the final metric scores. In our experiments, we set $ N = 5 $.

\textbf{Evaluation metrics.} Two evaluation metrics are employed: CMC curve and MAP.

We still use mean average precision (MAP) as a metric of performance. MAP is the mean value of the average precision (AP) scores of the whole queries, which can be formulated as follows:
\begin{align}
\label{equation:MAP}
\text{MAP} = \dfrac{\sum_{q=1}^{Q}\,\text{AP}(q)}{Q}
\end{align}

AP computes the average value of the precisions along with the variation of the recall, which is the area under the precision-recall curve. In practice, the integral is replaced with a finite sum over all the positions in the ranked sequence of the retrieved documents. Given a query $ q $, we define an indicator $ \delta(q,i) = 1 $ if the $ i $-th retrieved document is positive, and $ 0 $ otherwise. The precision at the $ k^\text{th} $ rank is given by $ P(q,k) = \frac{1}{k}\sum_{i=1}^{k}\,\delta(q,i) $. Denote $ \text{cl}(q) = \sum_{i=1}^{N}\,\delta(q,i) $ as the total number of positive documents in the database, then the average precision at $ k $ is:
\begin{align}
\label{equation:AP}
\text{AP}(q,k) = \dfrac{1}{\text{cl}(q)}\sum_{i=1}^{k}\,\delta(q,i)P(q,i)
\end{align}

Generally, we set $ k $ as the volume of the database so that to omit the second parameter $ k $ in Eq.~\ref{equation:AP} for simplification as Eq.~\ref{equation:MAP}.

MAP is a common metric of retrieval, which can reflect the overall performance of the methods. However, it lacks the insights into the details of the retrieval results. To overcome this shortage, we use an additional metric: cumulative matching characteristics (CMC) curve.

CMC curve is a common evaluation metric used in person re-identification~\cite{zheng2016person}, which represents the probability that the positive results can be found within the top $ n $ ranks of the returned list. No matter how many ground-truth matches are there in the database, only the first match is counted in the calculation. Compared to MAP, CMC curve is a fine-grained metric, which shows the variation of precision with the ranks. CMC curve is a good complementary metric for MAP.

\setlength{\tabcolsep}{9.4pt}
\begin{table}[h]
\begin{center}
\caption{Summarization of the Benchmark Datasets}
\label{Table:SummBenchmark}
\begin{tabular}{l|lcc}
\hline
Dataset & Media Types & Capacity & \# Categories \\
\hline
Wikipedia & image/article & 2,866 & 10 \\
Pascal Sentence & image/sentence & 1,000 & 20 \\
NUS-WIDE & image/tags & 67,994 & 10 \\
\hline
\end{tabular}
\end{center}
\end{table}

\section{Experimental Results}
\label{sec:ExpResults}

In this section, we evaluate some methods on three benchmark datasets under the existing protocol and the new protocol. These results can serve as the baselines for the future works.  

\subsection{Datasets}

In our experiments, we employ three datasets: Wikipedia, Pascal Sentence and NUS-WIDE. We summarize the three datasets in Table~\ref{Table:SummBenchmark} and provide dataset details as follows:

Wikipedia~\cite{rasiwasia2010new} contains 2,866 image-article pairs, and each pair is labeled by one of its 10 categories. It has a training set of 2,173 pairwise data and a testing set of the rest 693 pairs. In our experiments, we separate the database into the four subsets partially based on its original separation. That is, we build the database subsets from the original training set, and build the query subsets from the original testing set.

Pascal Sentence~\cite{rashtchian2010collecting} is a subset of Pascal VOC~\cite{everingham2010pascal}, which contains 1,000 images of 20 categories (50 images for each category). Each image is described by 5 sentences. In our experiments, we treat the 5 sentences of an image together as its corresponding texts. Besides, we use $ 80 \% $ data of each category to construct the database subsets and use the rest $ 20 \% $ data to construct the query subsets.

NUS-WIDE~\cite{chua2009nus} is a dataset that contains 269,648 images with their associated tags. Each image belongs to at least one of the 81 categories. In our experiments, we retain the images belonging to the most 10 categories. Finally, we obtain a set of 67,994 images. We use $ 60 \% $ data of this set to construct the database subsets and the rest $ 40 \% $ data to construct the query subsets.

\begin{figure*}[t]
\begin{center}
\subfigure[Non-XTD retrieval: I$ \rightarrow $T]{\includegraphics[width=0.24\linewidth]{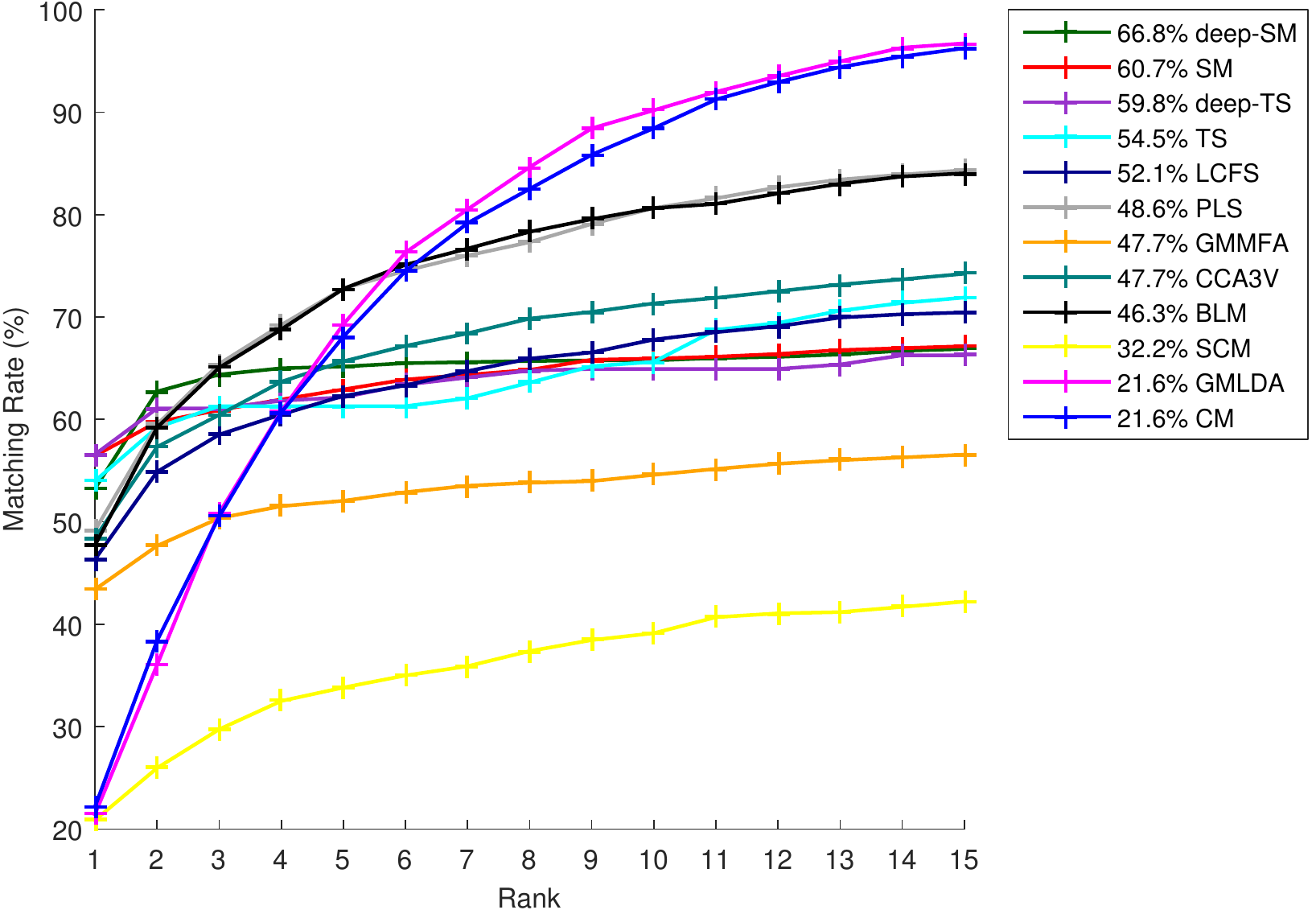}\label{subfig:wiki_i2t1_rl}}
\subfigure[XTD retrieval: I$ \rightarrow $T]{\includegraphics[width=0.24\linewidth]{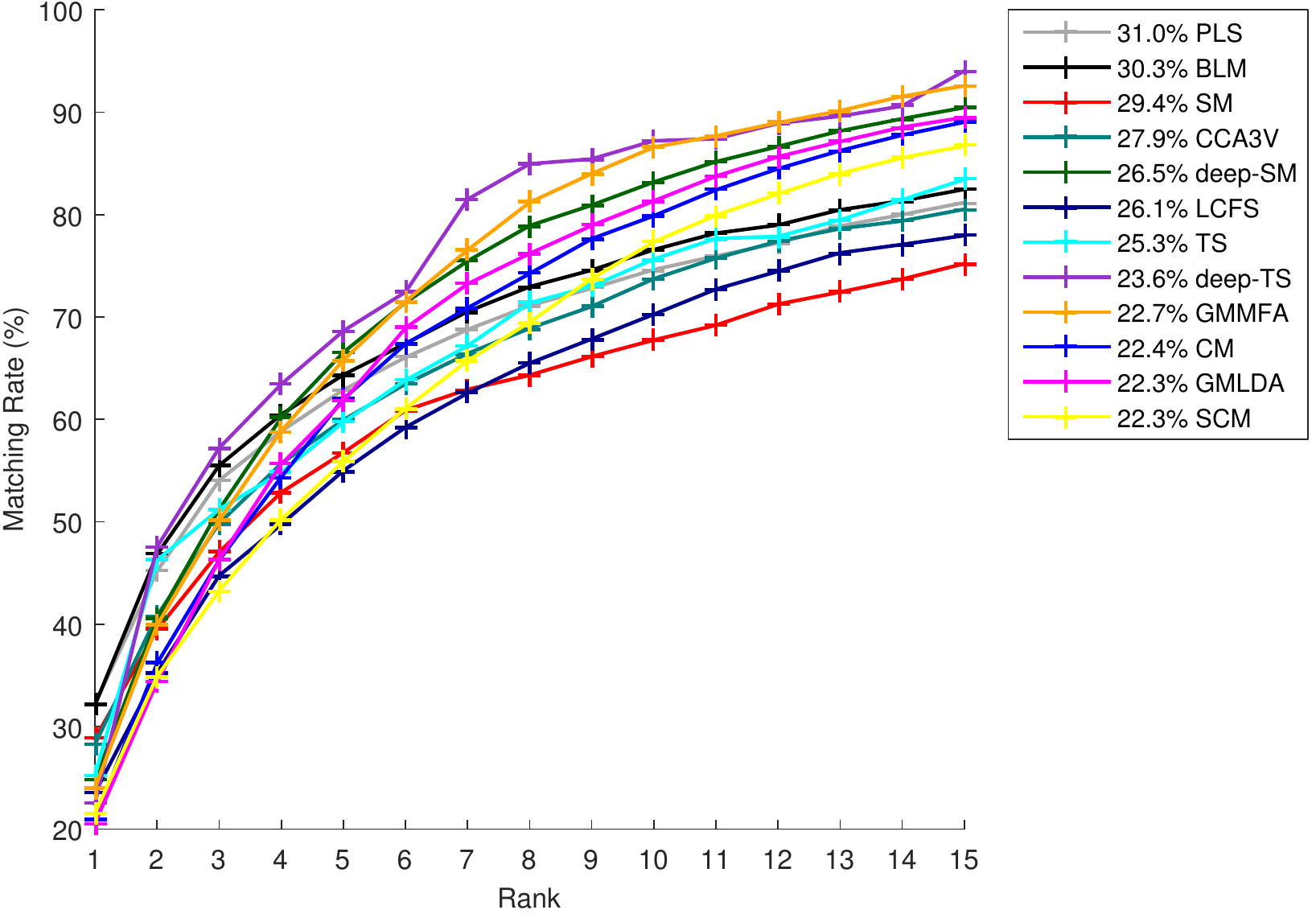}\label{subfig:wiki_i2t2_rl}}
\subfigure[Non-XTD retrieval: T$ \rightarrow $I]{\includegraphics[width=0.24\linewidth]{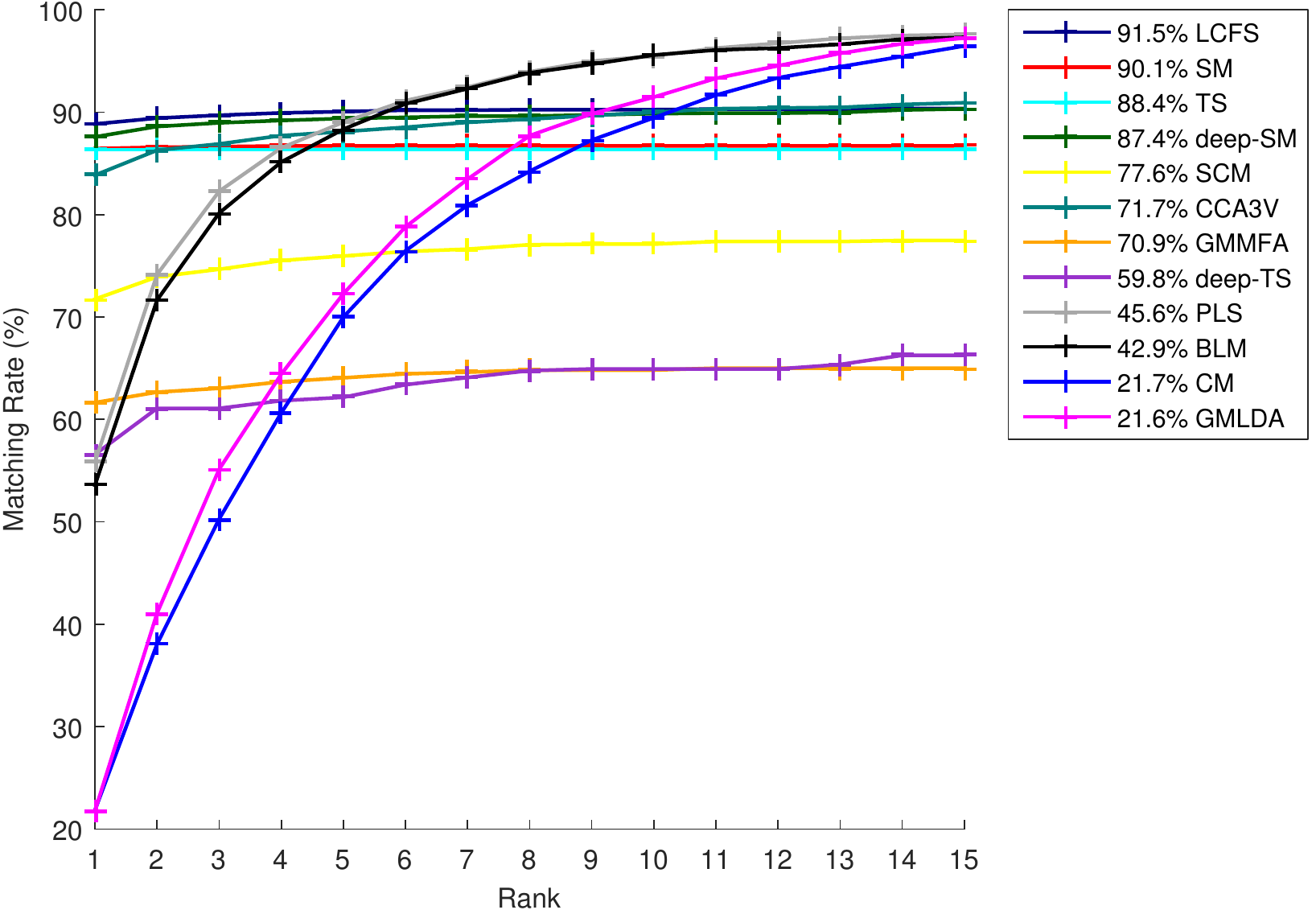}}
\subfigure[XTD retrieval: T$ \rightarrow $I]{\includegraphics[width=0.24\linewidth]{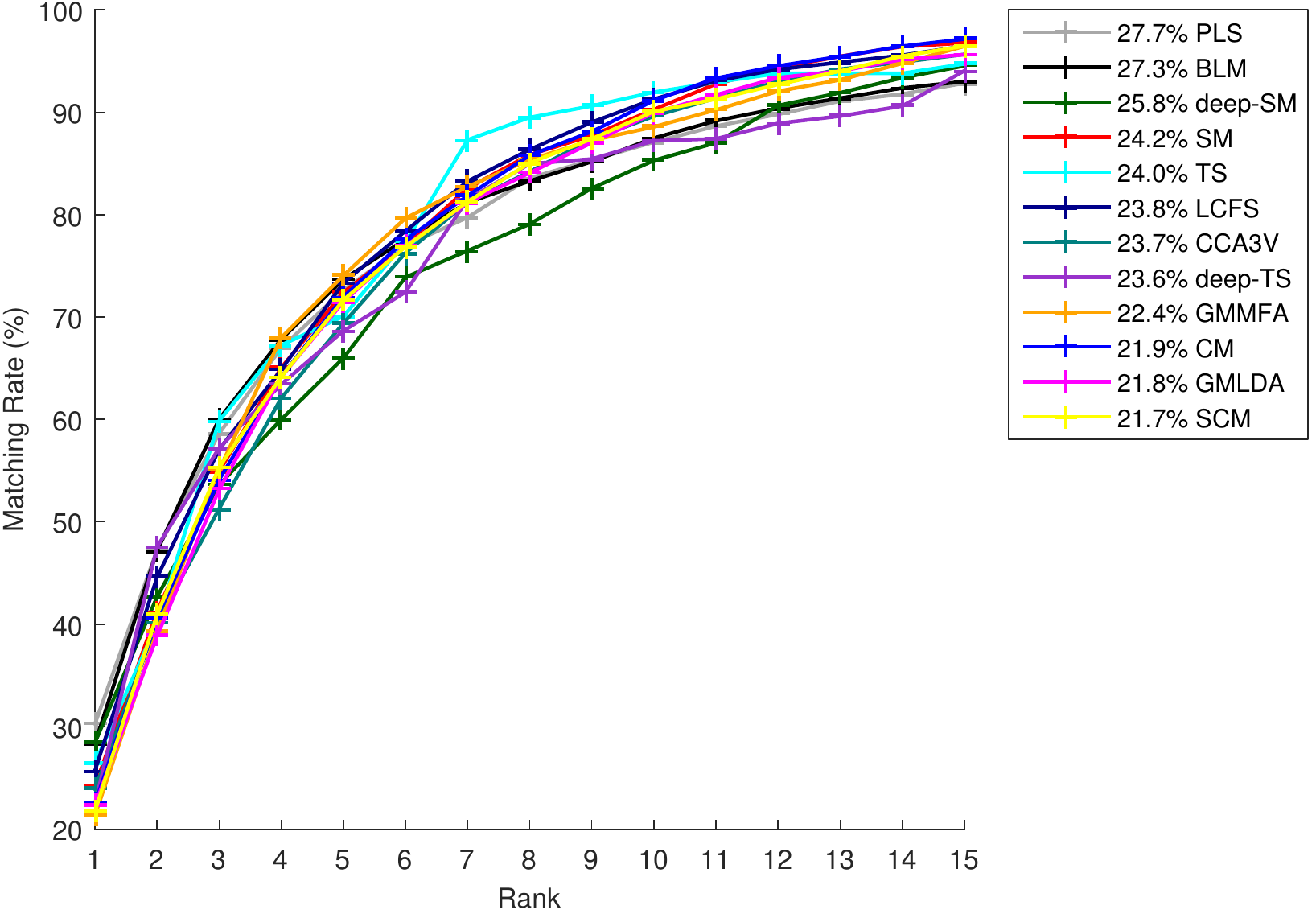}}
\end{center}
	\caption{Evaluation results of real-valued representations on Wikipedia. CMC curves are shown. MAP is shown before the name of each method. (a) and (c) represent the non-extendable retrieval results, while (b) and (d) are the extendable retrieval results. (a) and (b) denote image-to-text retrieval, while (c) and (d) denote text-to-image retrieval.}
	\label{fig:rlWiki}
\end{figure*}

\begin{figure*}[!t]
\begin{center}
\subfigure[Non-XTD retrieval: I$ \rightarrow $T]{\includegraphics[width=0.24\linewidth]{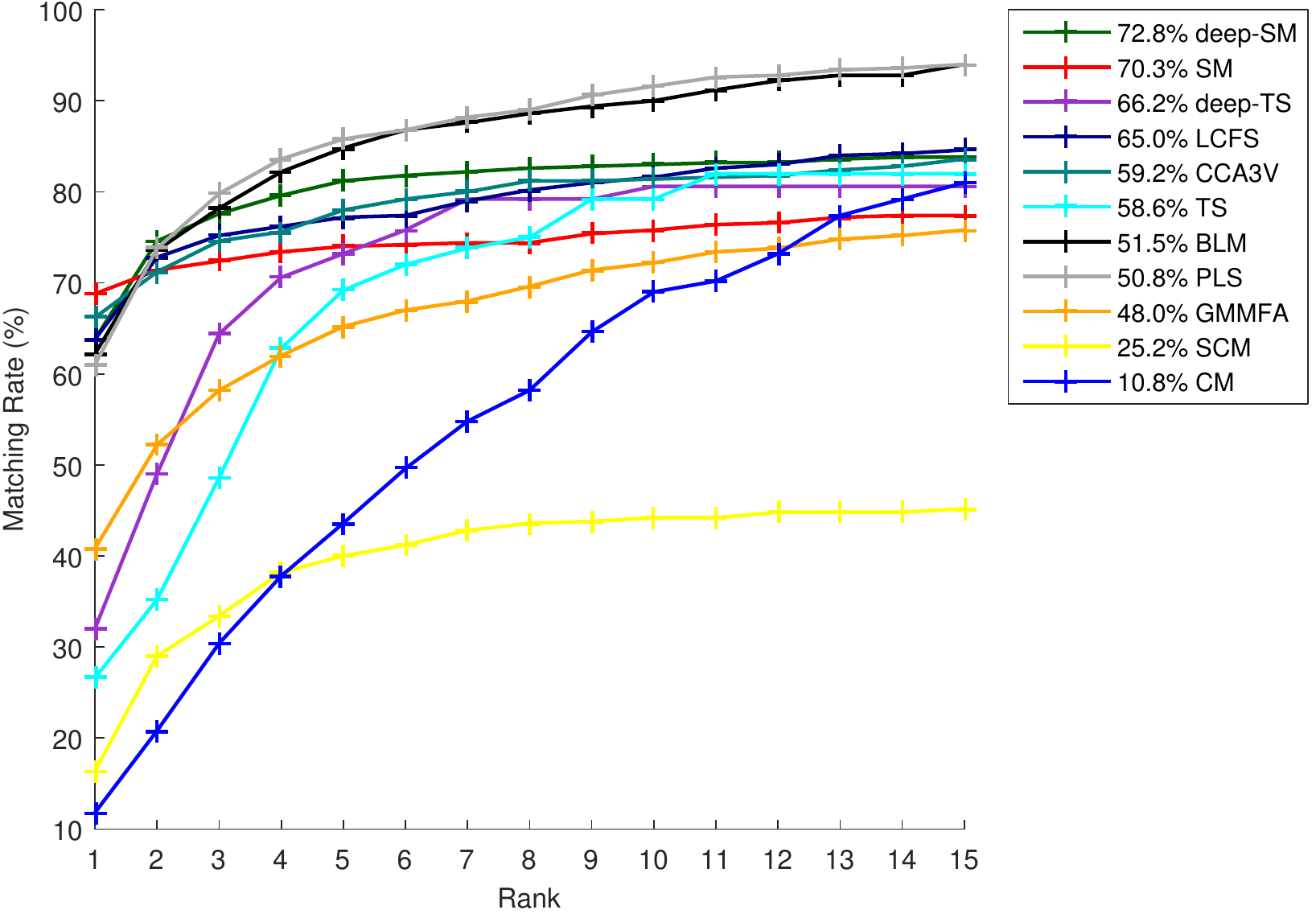}}
\subfigure[XTD retrieval: I$ \rightarrow $T]{\includegraphics[width=0.24\linewidth]{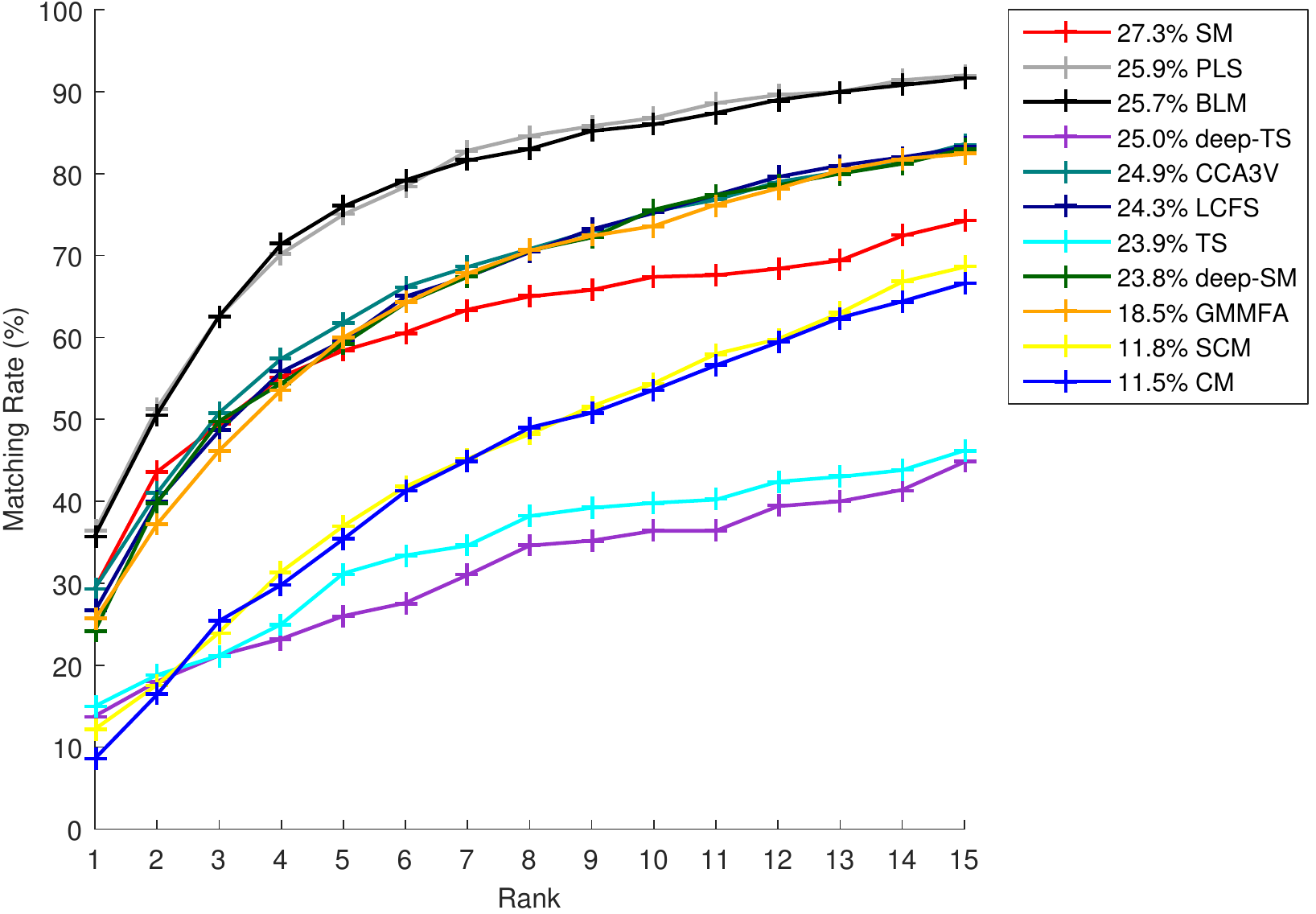}}
\subfigure[Non-XTD retrieval: T$ \rightarrow $I]{\includegraphics[width=0.24\linewidth]{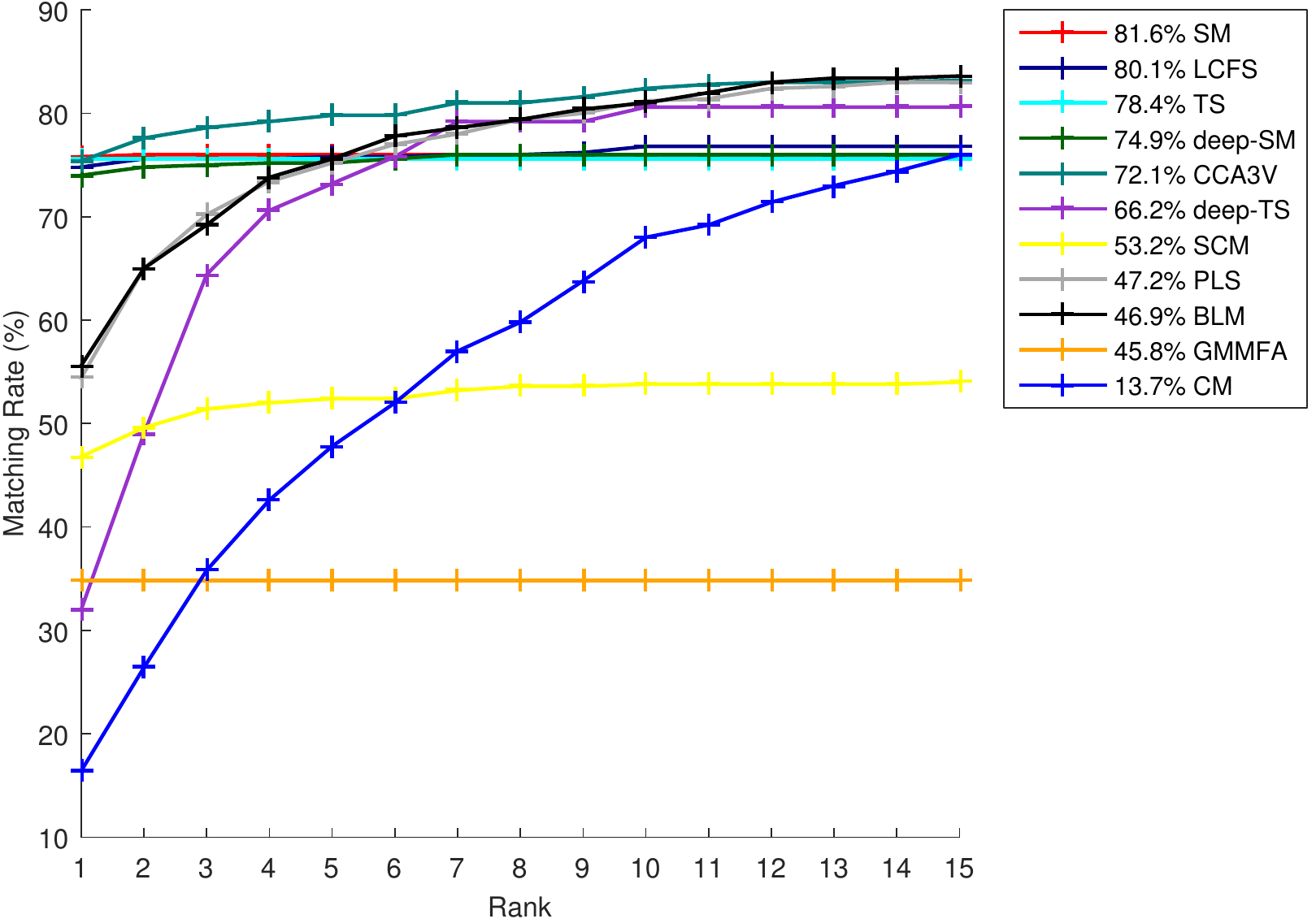}}
\subfigure[XTD retrieval: T$ \rightarrow $I]{\includegraphics[width=0.24\linewidth]{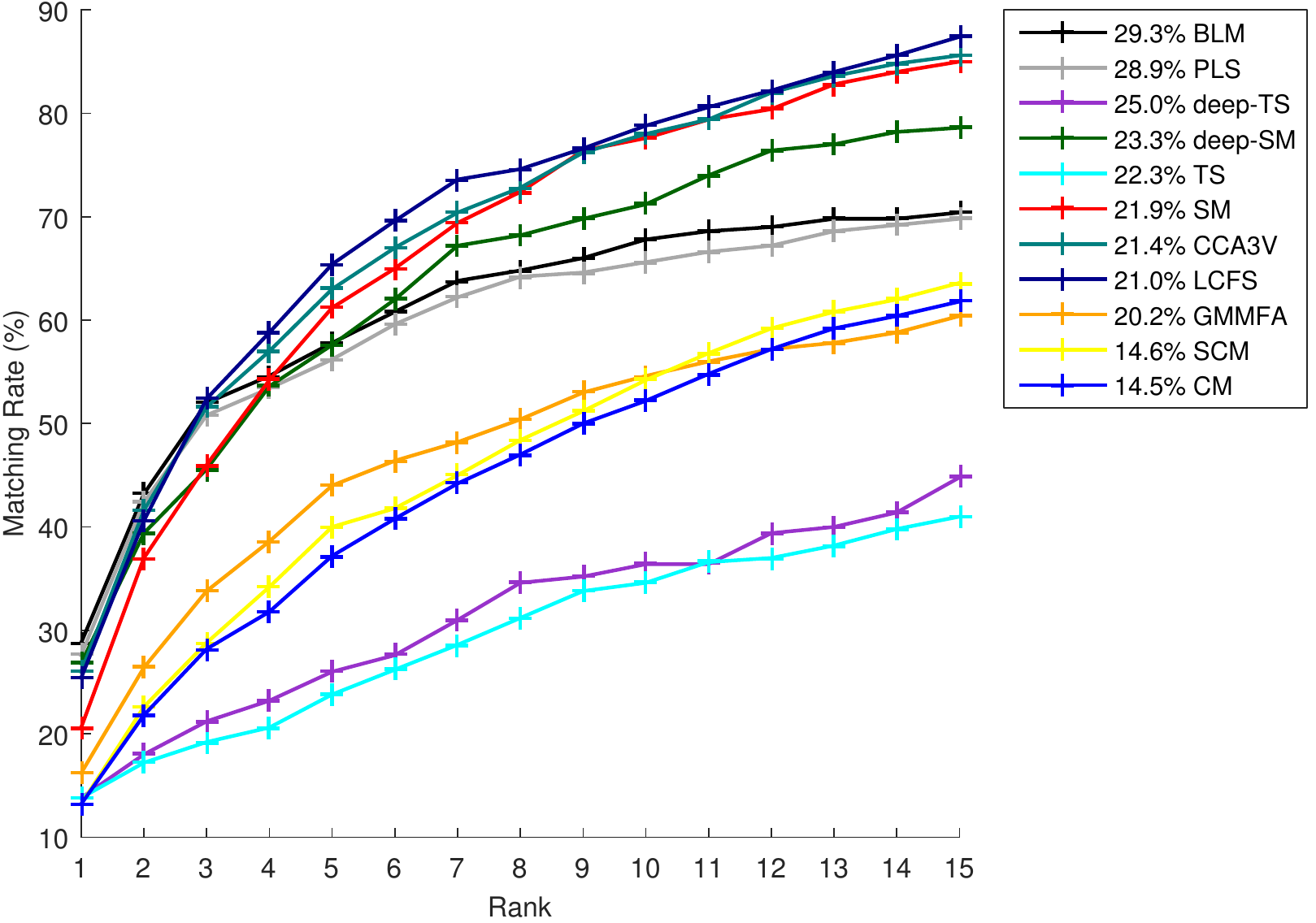}}
\end{center}
	\caption{Evaluation results of real-valued representations on Pascal Sentence. CMC curves are shown. MAP is shown before the name of each method. Retrieval modes in (a)(b)(c)(d) are the same with Fig. \ref{fig:rlWiki}.}
	\label{fig:rlPascal}
\end{figure*}

\begin{figure*}[!t]
\begin{center}
\subfigure[Non-XTD retrieval: I$ \rightarrow $T]{\includegraphics[width=0.24\linewidth]{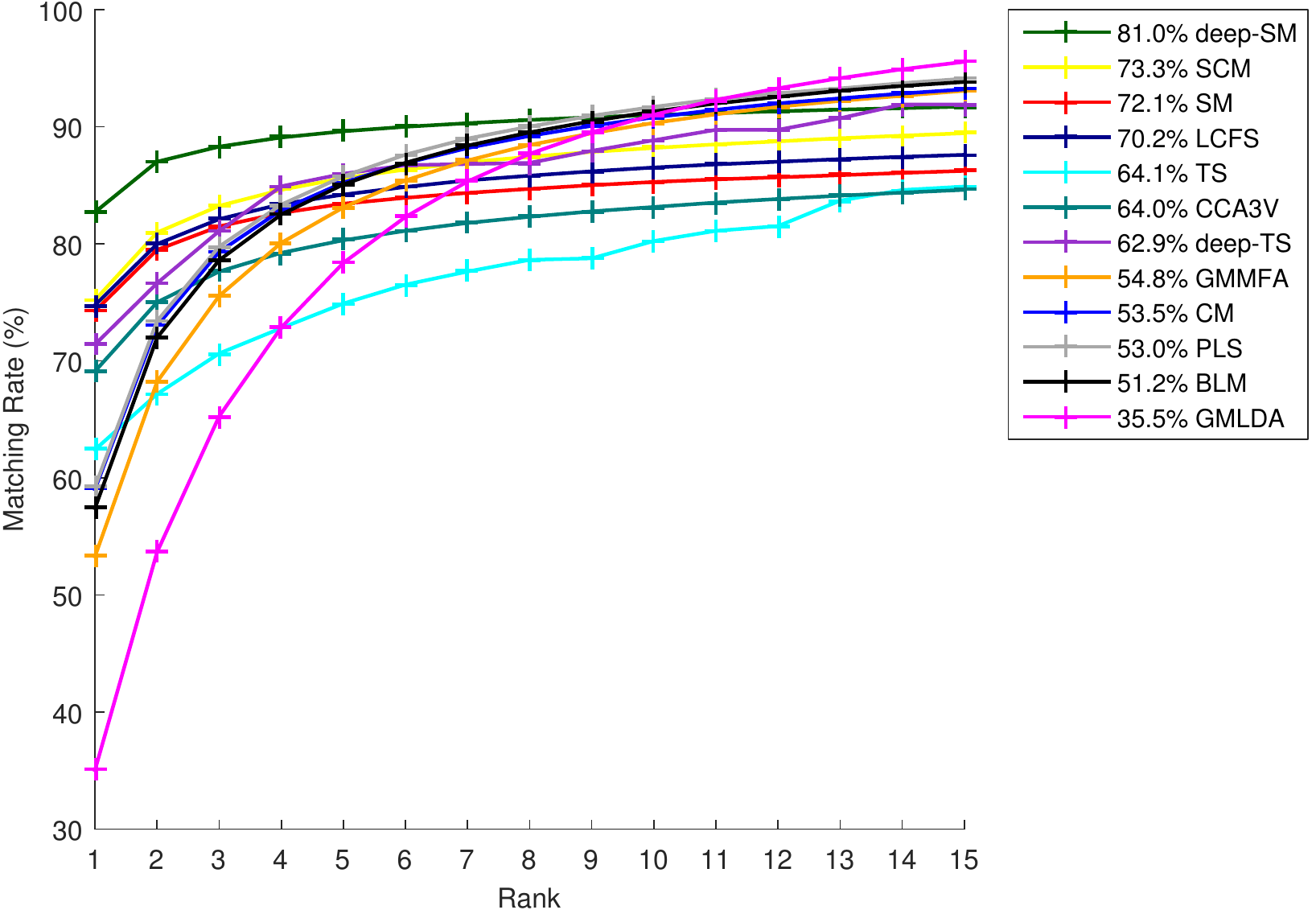}}
\subfigure[XTD retrieval: I$ \rightarrow $T]{\includegraphics[width=0.24\linewidth]{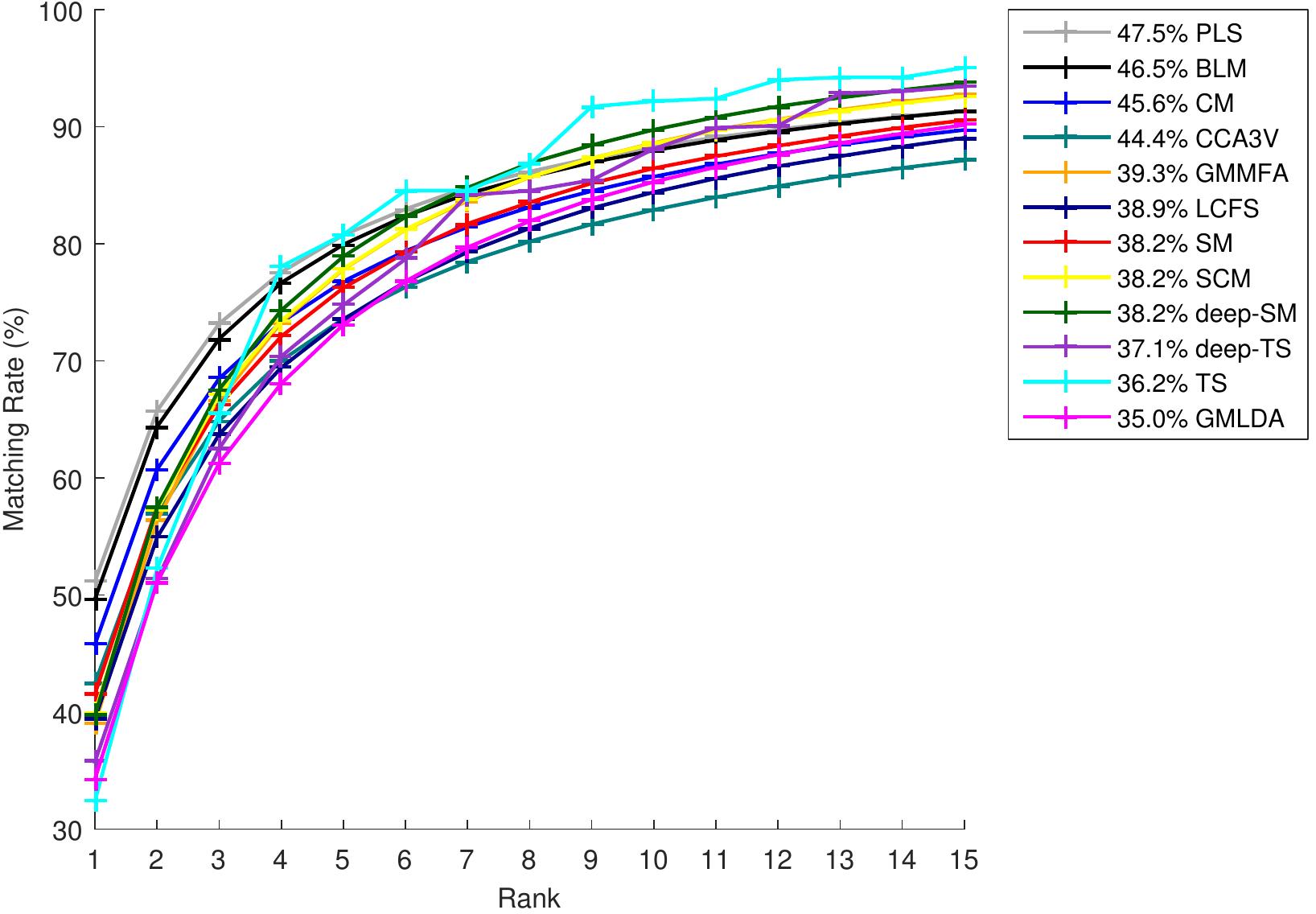}}
\subfigure[Non-XTD retrieval: T$ \rightarrow $I]{\includegraphics[width=0.24\linewidth]{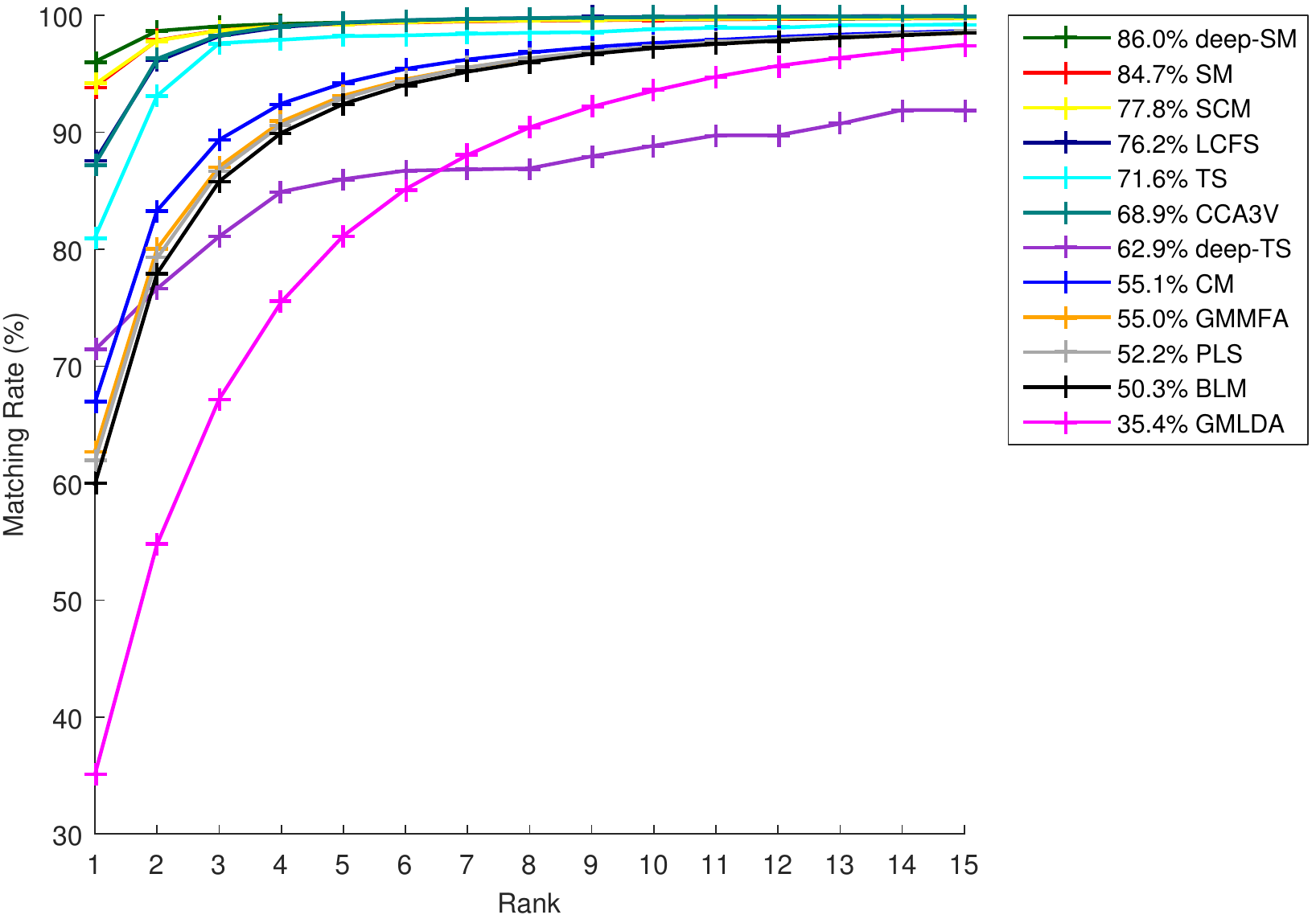}}
\subfigure[XTD retrieval: T$ \rightarrow $I]{\includegraphics[width=0.24\linewidth]{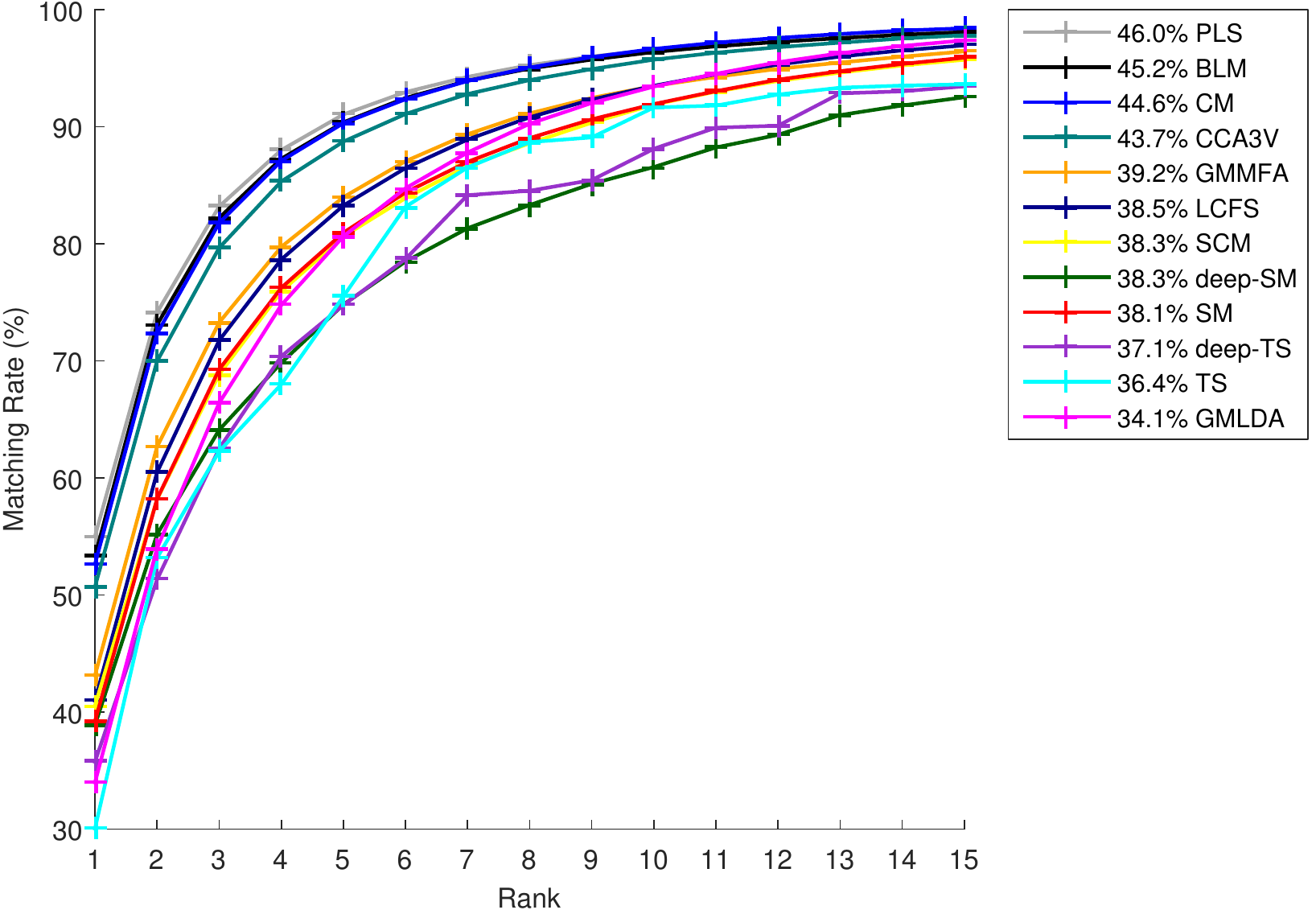}}
\end{center}
	\caption{Evaluation results of real-valued representations on NUS-WIDE. The MAP scores are shown before the names of the methods. Retrieval modes in (a)(b)(c)(d) are the same with Fig. \ref{fig:rlWiki} and Fig. \ref{fig:rlPascal}.}
	\label{fig:rlNUSWide}
\end{figure*}

\subsection{Image and Text Features} \label{sec:exp_features}
For images, we extract the $ 6^\text{th} $ layer (4096D) of CaffeNet~\cite{jia2014caffe}, which is a simplified version of AlexNet~\cite{krizhevsky2012imagenet} and is pre-trained the 1,000-class ImageNet~\cite{deng2009imagenet} dataset, as the image descriptors. For texts, we use the word2vec model pre-trained on the Google News dataset~\cite{mikolov2013distributed} to represent text documents. We cluster the whole word vectors into a dictionary and then encode each text document into a word-frequency vector (1024D), \ie, the Bag-of-Words (BoW) feature.

\subsection{Baselines}\label{sec:exp_baseline}
We compare the following real-valued and binary representations. The real-valued representations include:

\begin{itemize}
\item CM: correlation matching~\cite{rasiwasia2010new} employs canonical correlation analysis (CCA)~\cite{hotelling1936relations} to learn the uniform descriptors.
\item SM: semantic matching~\cite{rasiwasia2010new} learns two linear classifiers that map data into the semantic concept probabilities.
\item SCM: semantic correlation matching~\cite{rasiwasia2010new} is the combination of CM and SM.
\item PLS: partial least squares~\cite{sharma2011bypassing} is a method to learn common subspace to make corresponding data high correlated.
\item BLM: bilinear model~\cite{turk1991eigenfaces} is another common subspace learning method.
\item GMMFA: generalized multiview analysis (GMA)~\cite{sharma2012generalized} + marginal Fisher analysis (MFA)~\cite{yan2007graph} is a supervised extension of CCA and is used to extend MFA.
\item GMLDA: GMA~\cite{sharma2012generalized} + linear discriminant analysis (LDA)~\cite{belhumeur1997eigenfaces} is the same as GMMFA but is used to extend LDA.
\item CCA3V: three-view canonical correlation analysis~\cite{sharma2012generalized} incorporates a third view capturing high-level semantics that comes from supervised ground-truth labels or unsupervised clustering the tags.
\item LCFS: Learning coupled feature spaces~\cite{wang2013learning} learns two projection matrices to map multimodal data into a common feature space and imposes $ \ell_{21} $-norm penalties to select relevant and discriminative features.
\item deep-SM: deep semantic matching~\cite{wei2016cross} uses deep networks to replace the linear classifiers in SM.
\end{itemize}

\begin{figure*}[t]
\begin{center}
\subfigure[Image-to-text retrieval]{\includegraphics[width=0.49\linewidth]{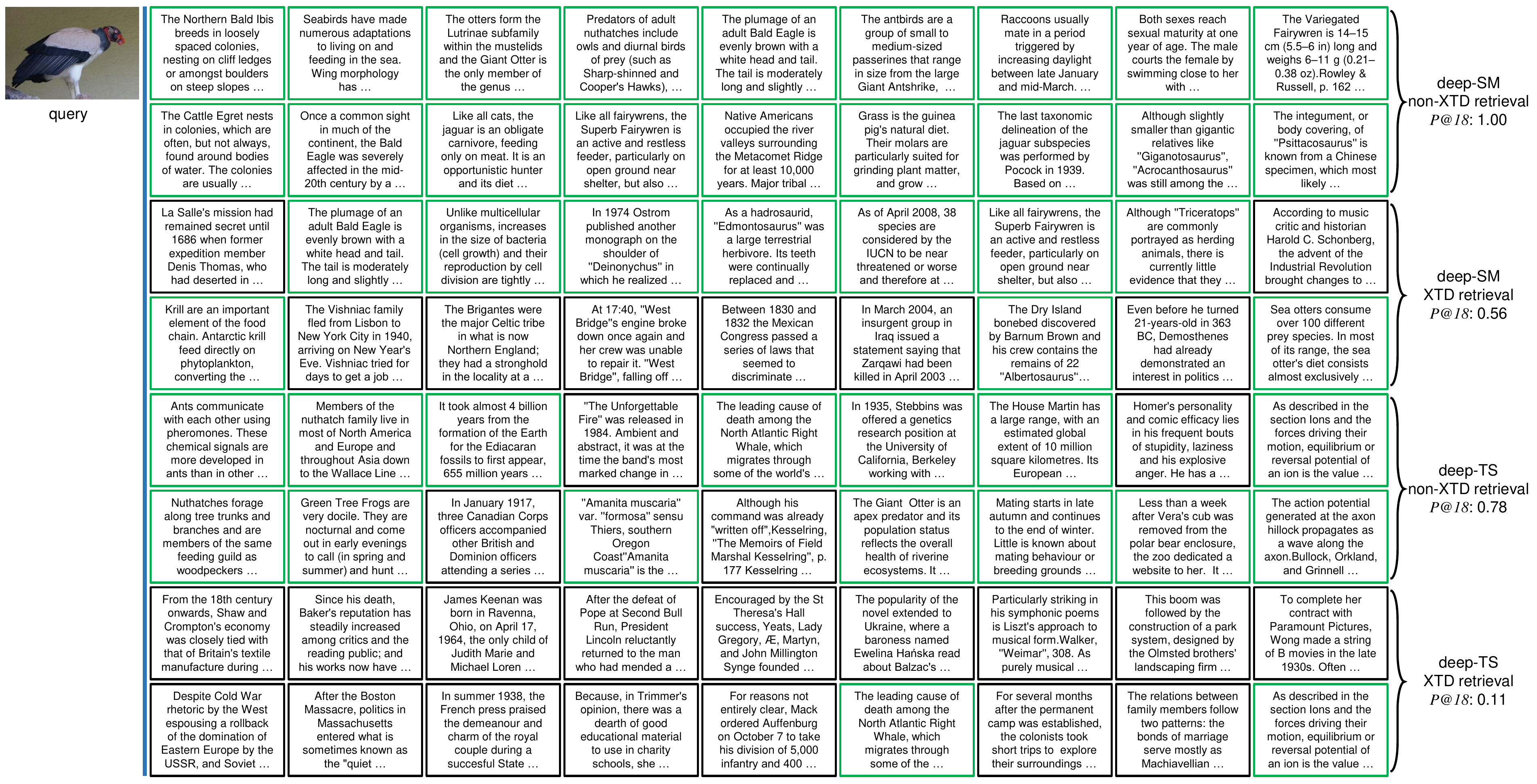}}
\subfigure[Text-to-image retrieval]{\includegraphics[width=0.49\linewidth]{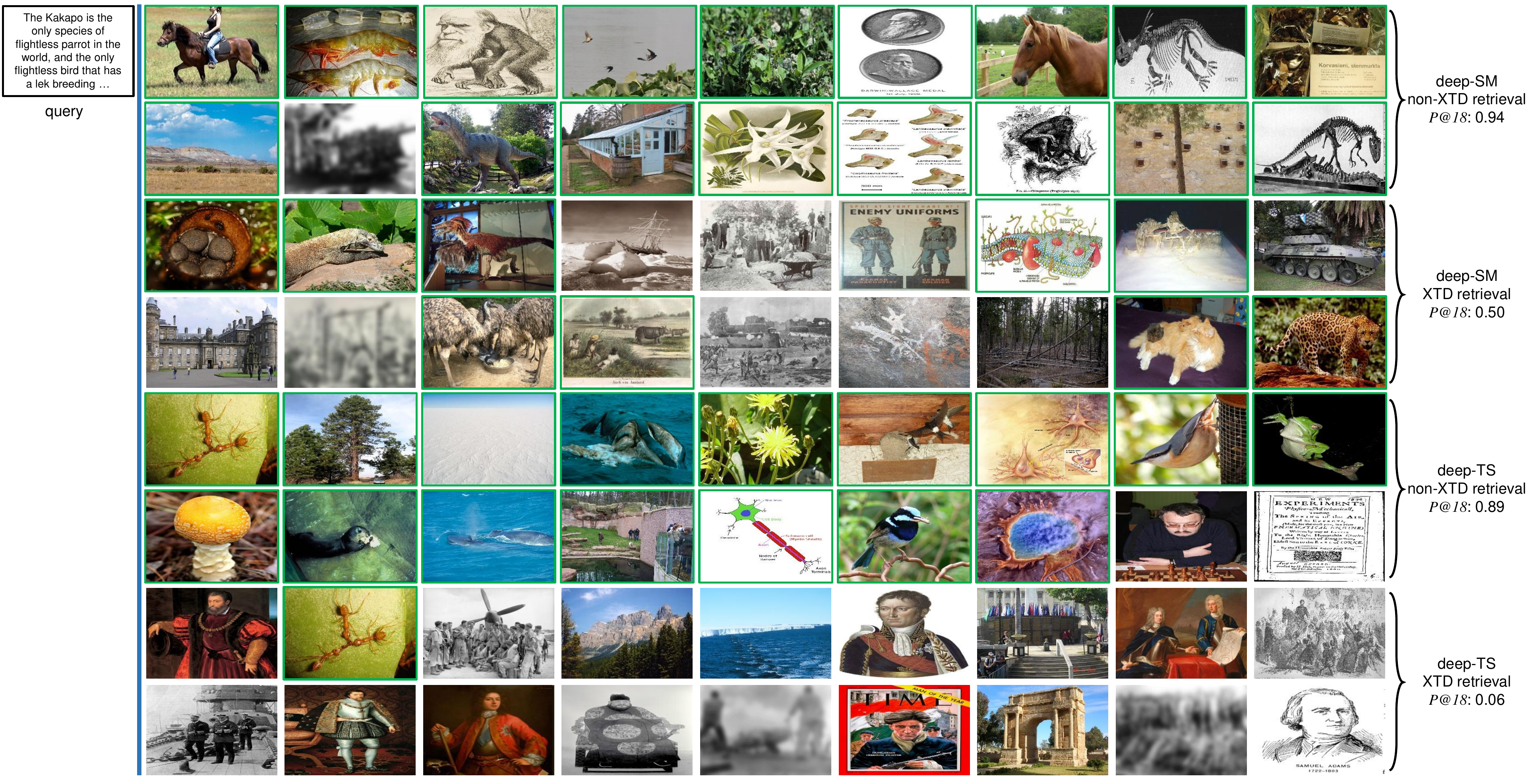}}
\end{center}
	\caption{Sample retrieval results of (a) image query 658 and (b) text query 9 on Wikipedia. The query is on the \textit{left}. The first two rows correspond to the non-extendable retrieval of deep-SM. The third and fourth rows correspond to the extendable retrieval of deep-SM. The fifth and sixth rows correspond to the non-extendable retrieval of deep-TS. The last two rows correspond to the extendable retrieval of deep-TS. The performance drop of deep-TS is much larger from non-extendable retrieval to extendable retrieval and it loses competitive accuracy under the extendable setting.}
	\label{fig:retrieval_res}
\end{figure*}

\begin{figure*}[!t]
\begin{center}
\subfigure[Non-XTD retrieval: I$ \rightarrow $T (8b)]{\includegraphics[width=0.24\linewidth]{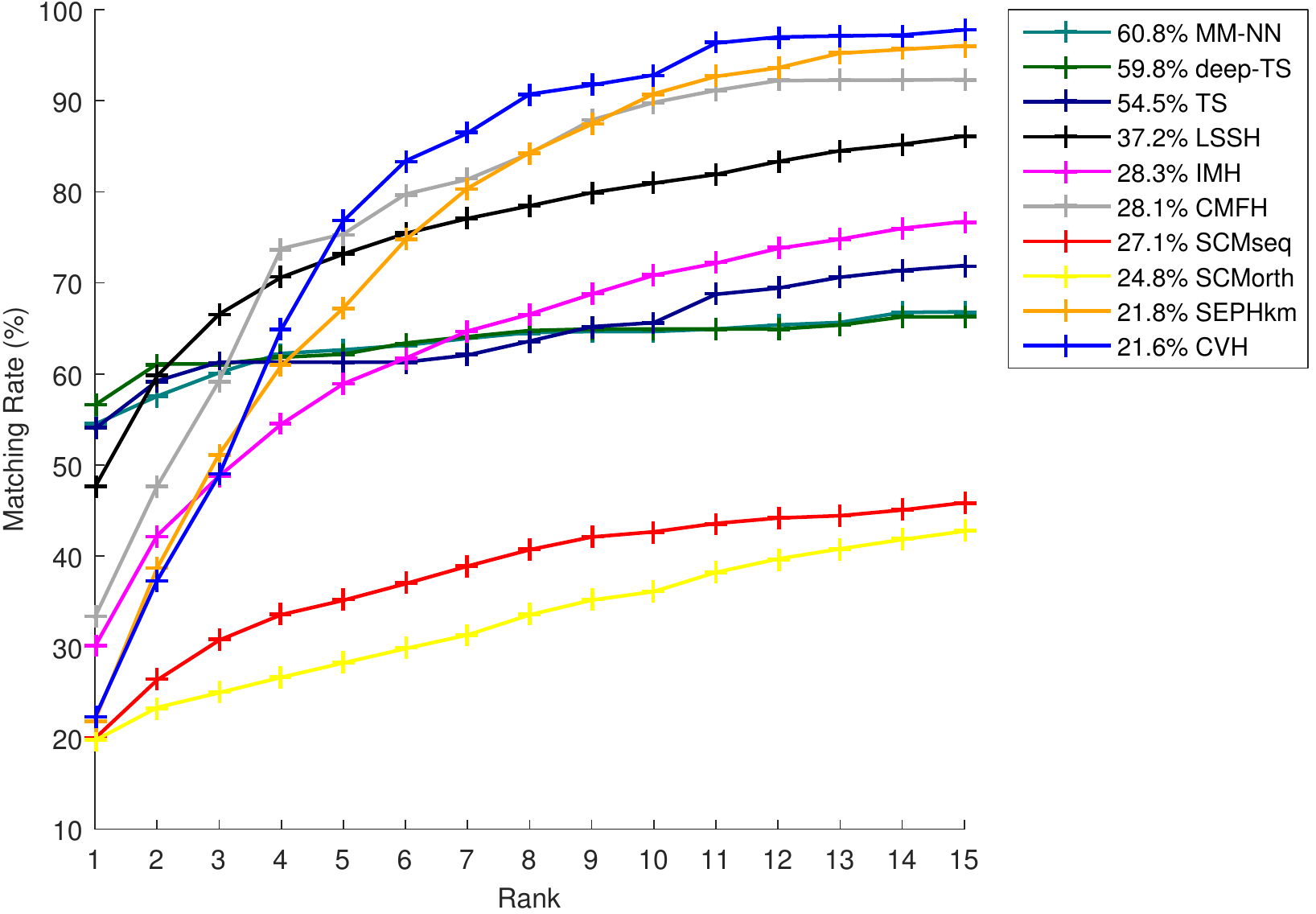}\label{subfig:wiki_i2t1_hs8}}
\subfigure[XTD retrieval: I$ \rightarrow $T (8b)]{\includegraphics[width=0.24\linewidth]{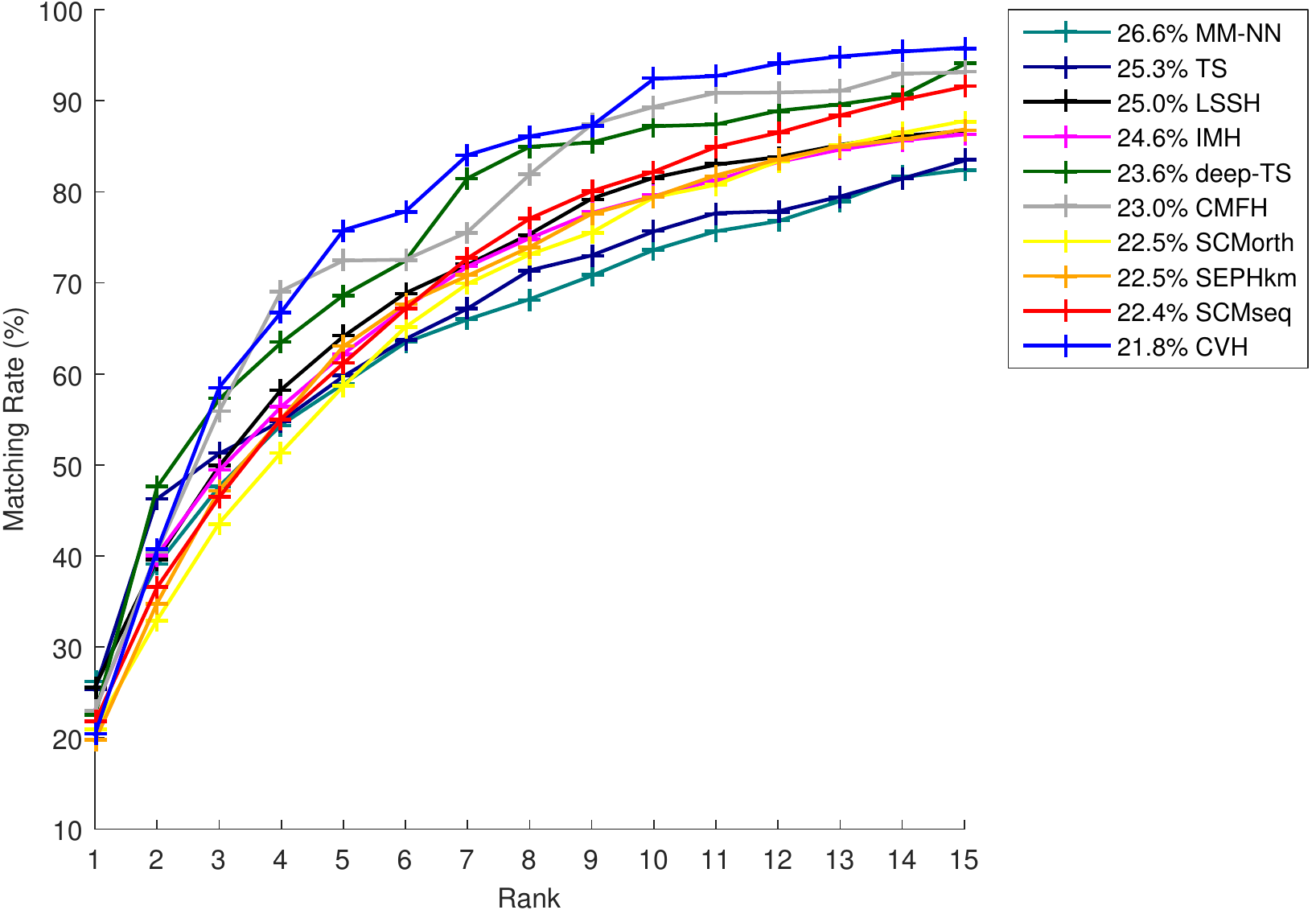}\label{subfig:wiki_i2t2_hs8}}
\subfigure[Non-XTD retrieval: T$ \rightarrow $I (8b)]{\includegraphics[width=0.24\linewidth]{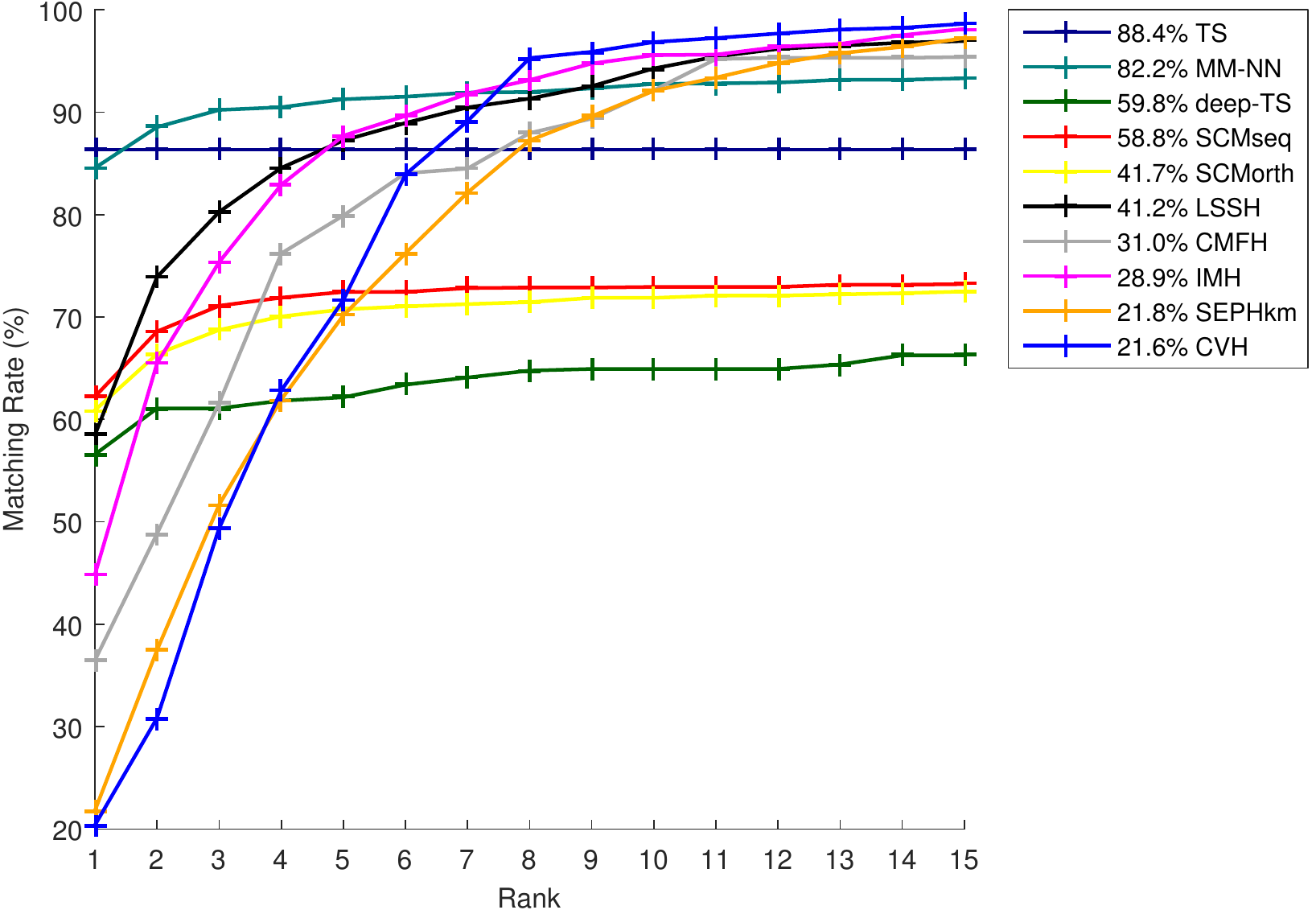}\label{subfig:wiki_t2i1_hs8}}
\subfigure[XTD retrieval: T$ \rightarrow $I (8b)]{\includegraphics[width=0.24\linewidth]{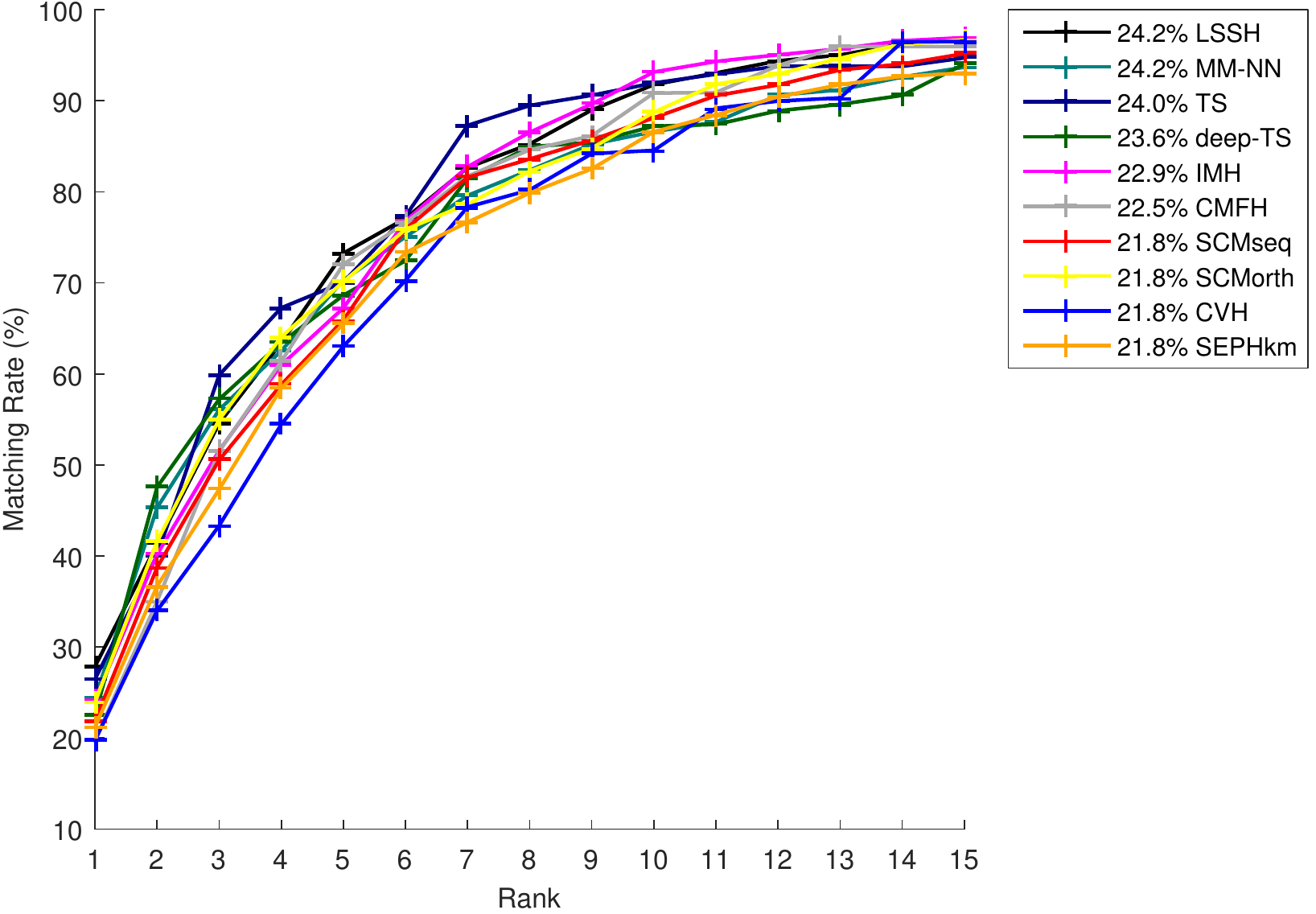}}
\subfigure[Non-XTD retrieval: I$ \rightarrow $T (16b)]{\includegraphics[width=0.24\linewidth]{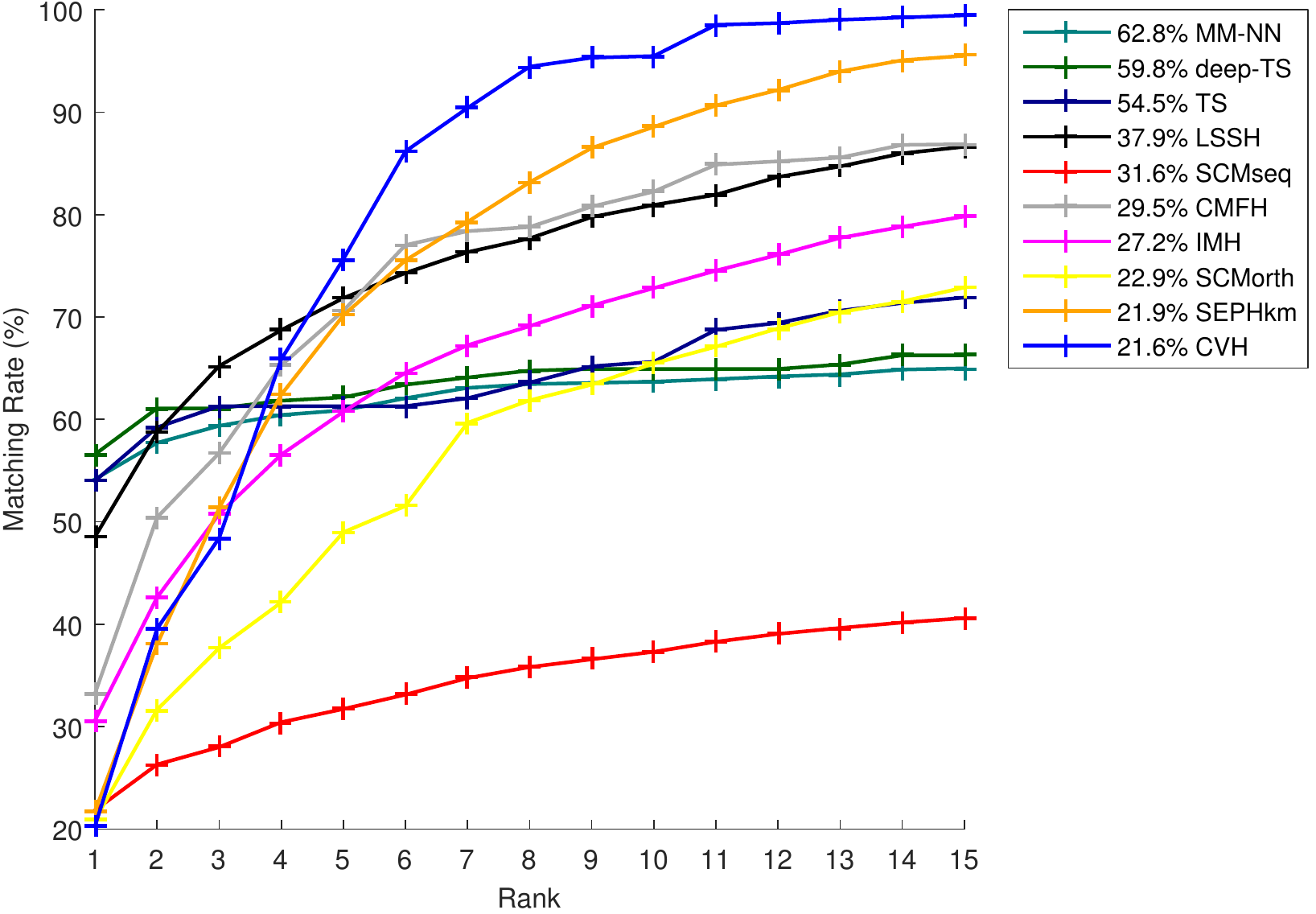}\label{subfig:wiki_i2t1_hs16}}
\subfigure[XTD retrieval: I$ \rightarrow $T (16b)]{\includegraphics[width=0.24\linewidth]{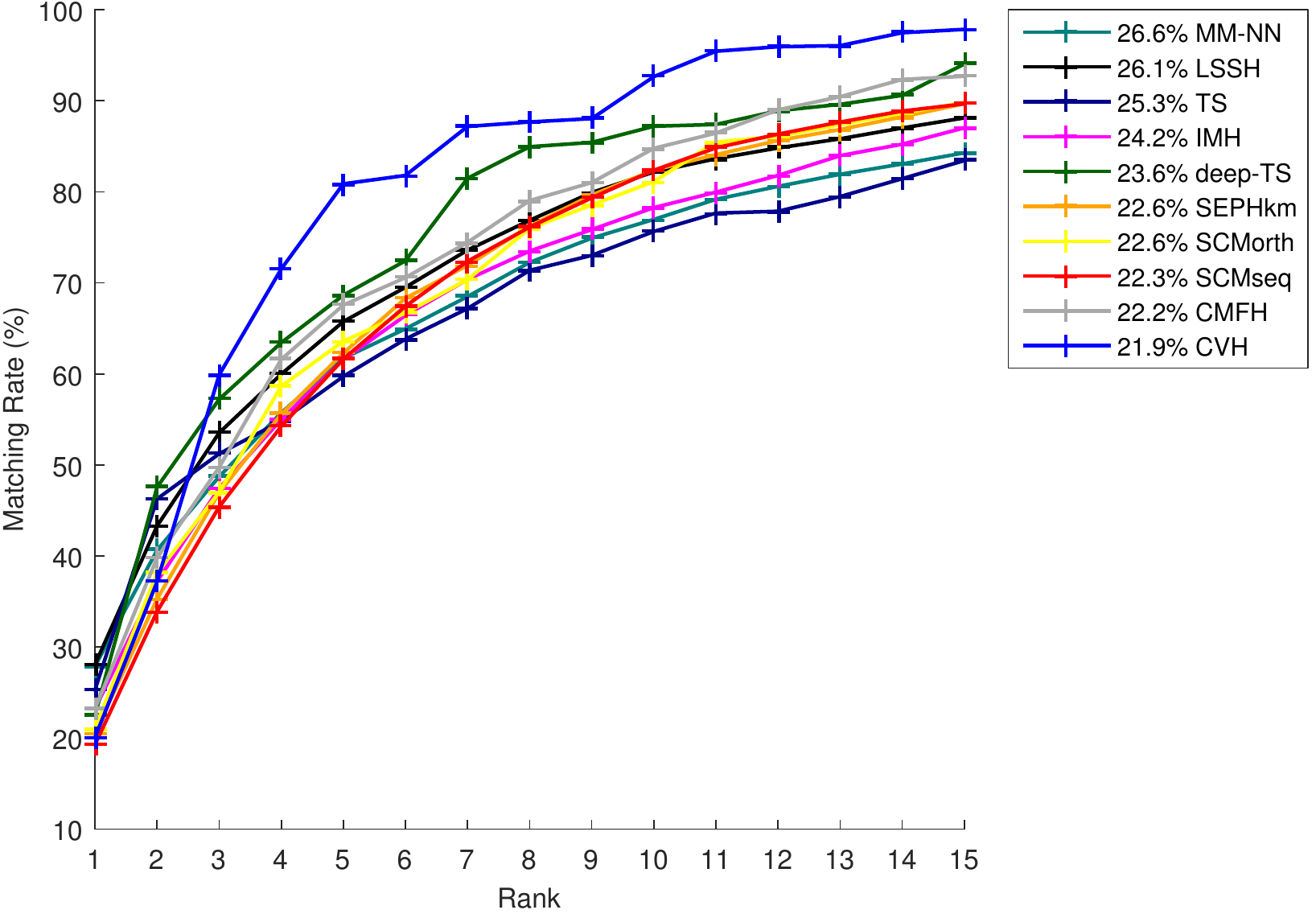}}
\subfigure[Non-XTD retrieval: T$ \rightarrow $I (16b)]{\includegraphics[width=0.24\linewidth]{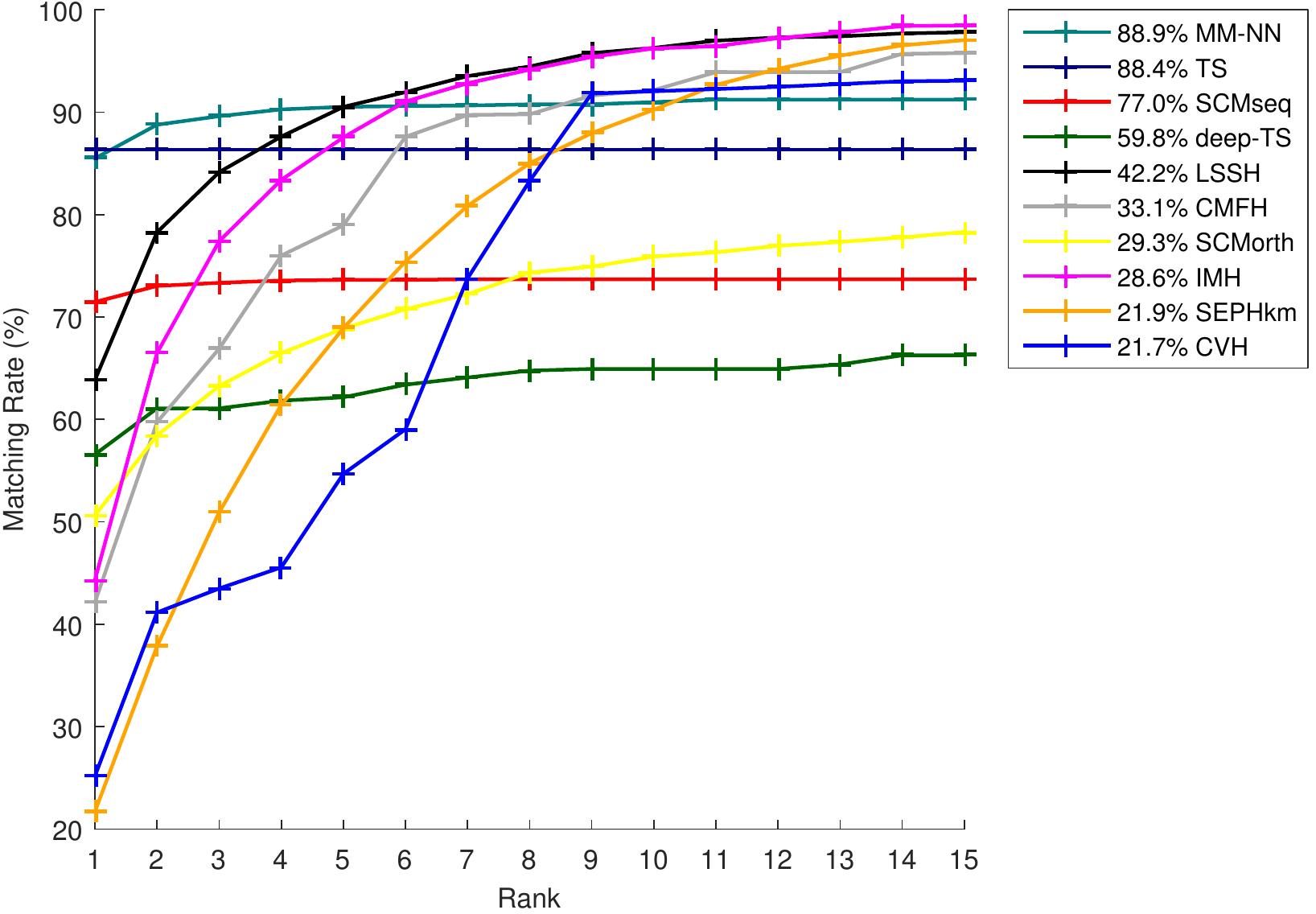}\label{subfig:wiki_t2i1_hs16}}
\subfigure[XTD retrieval: T$ \rightarrow $I (16b)]{\includegraphics[width=0.24\linewidth]{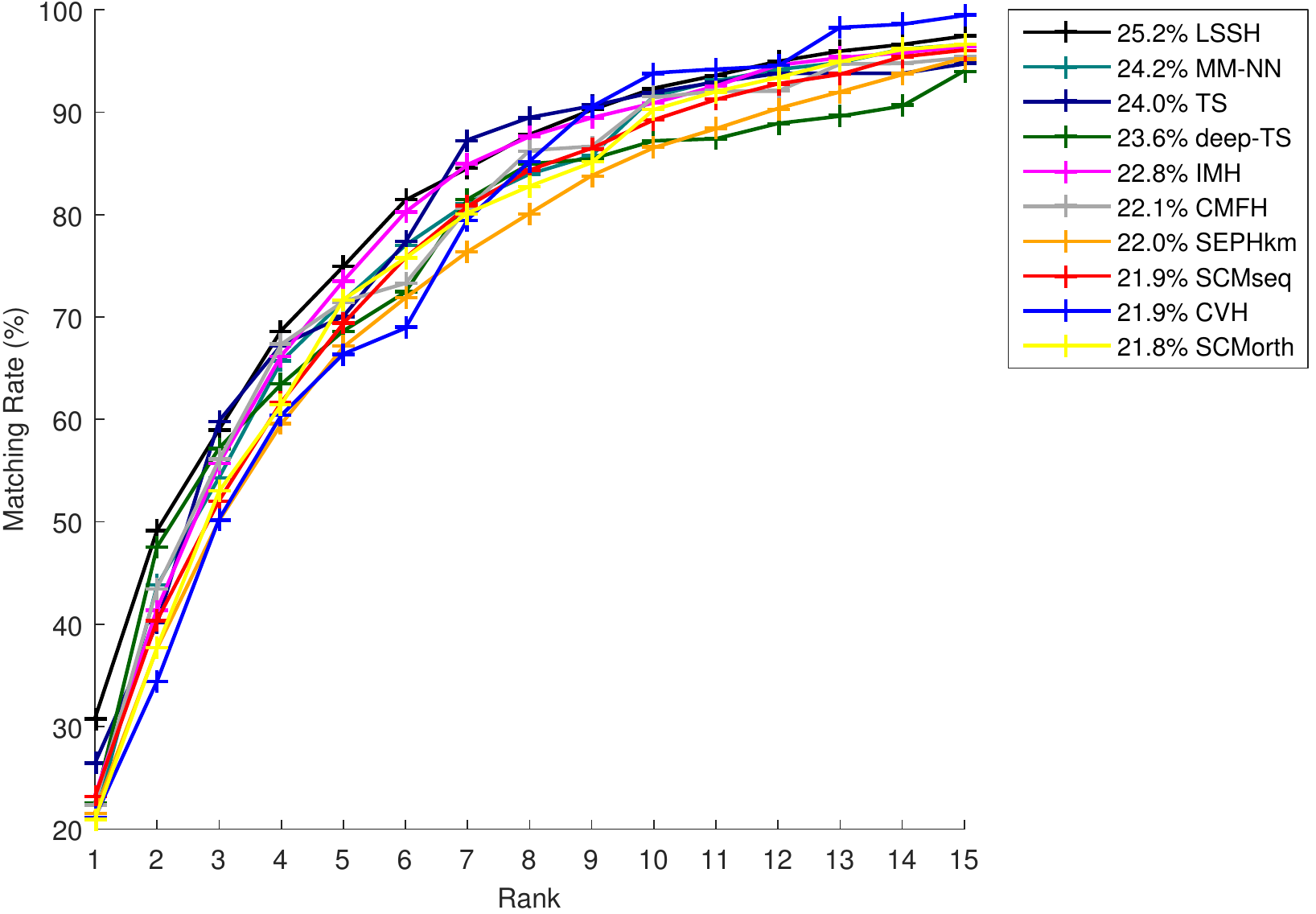}}
\subfigure[Non-XTD retrieval: I$ \rightarrow $T (32b)]{\includegraphics[width=0.24\linewidth]{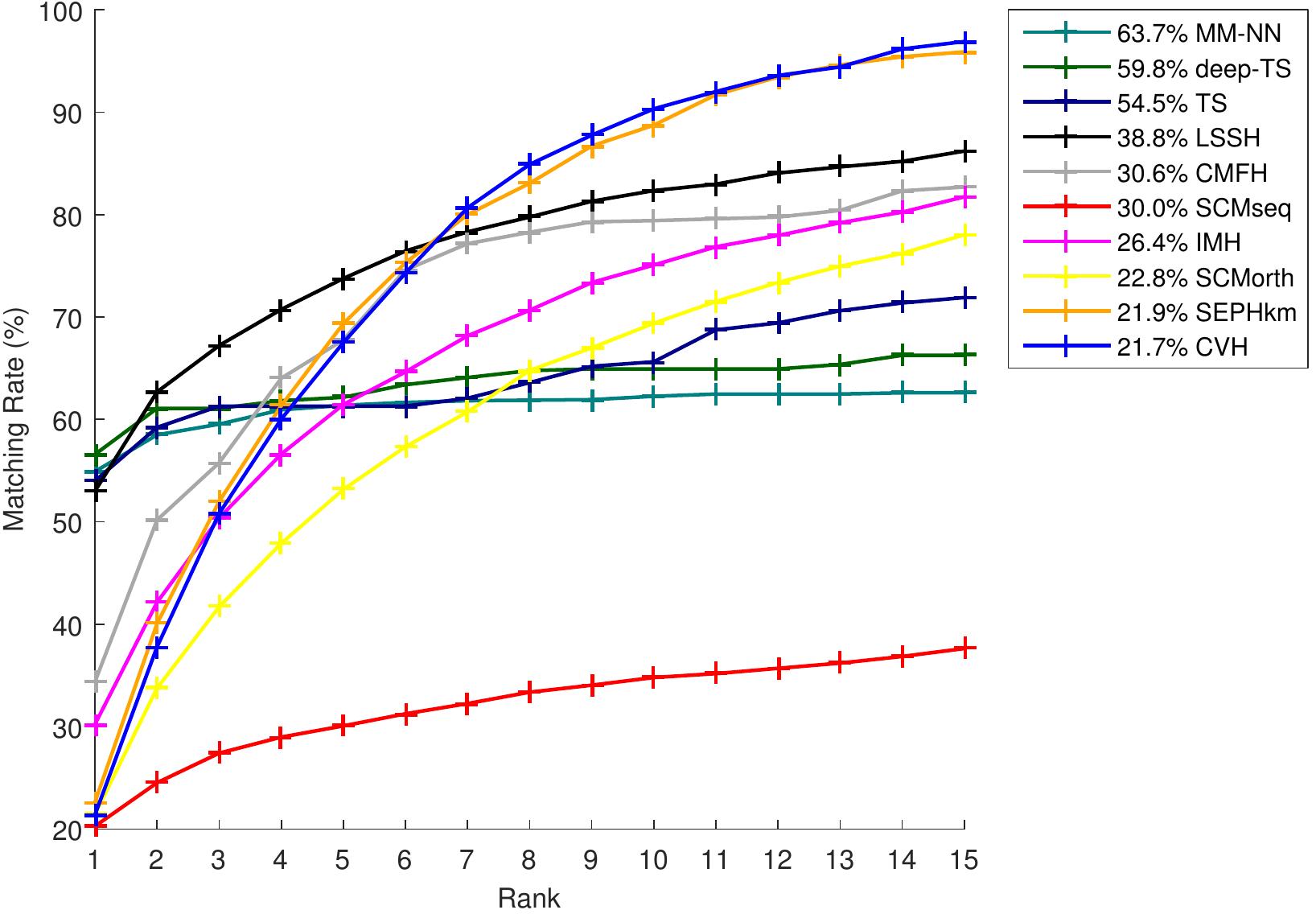}}
\subfigure[XTD retrieval: I$ \rightarrow $T (32b)]{\includegraphics[width=0.24\linewidth]{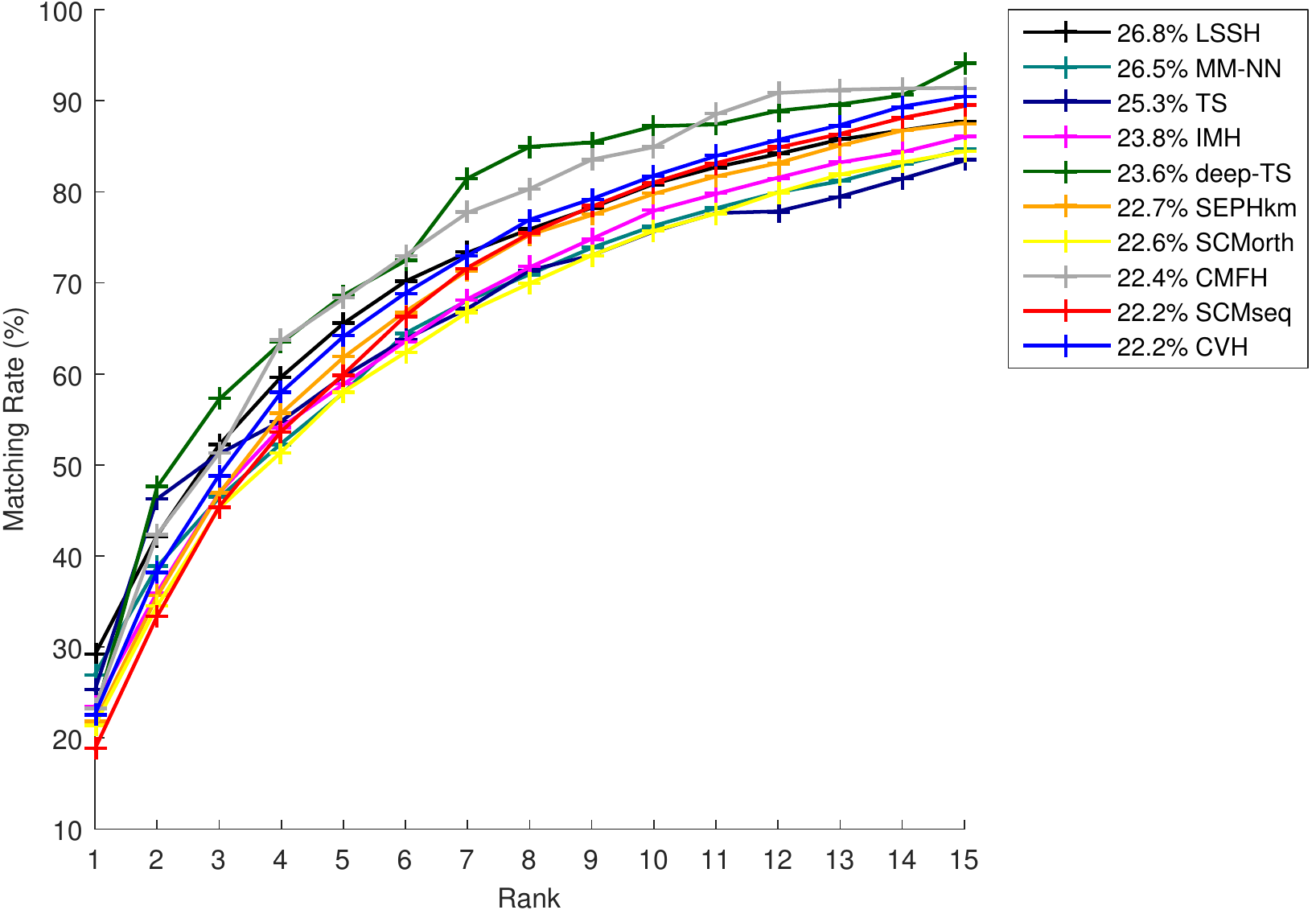}}
\subfigure[Non-XTD retrieval: T$ \rightarrow $I (32b)]{\includegraphics[width=0.24\linewidth]{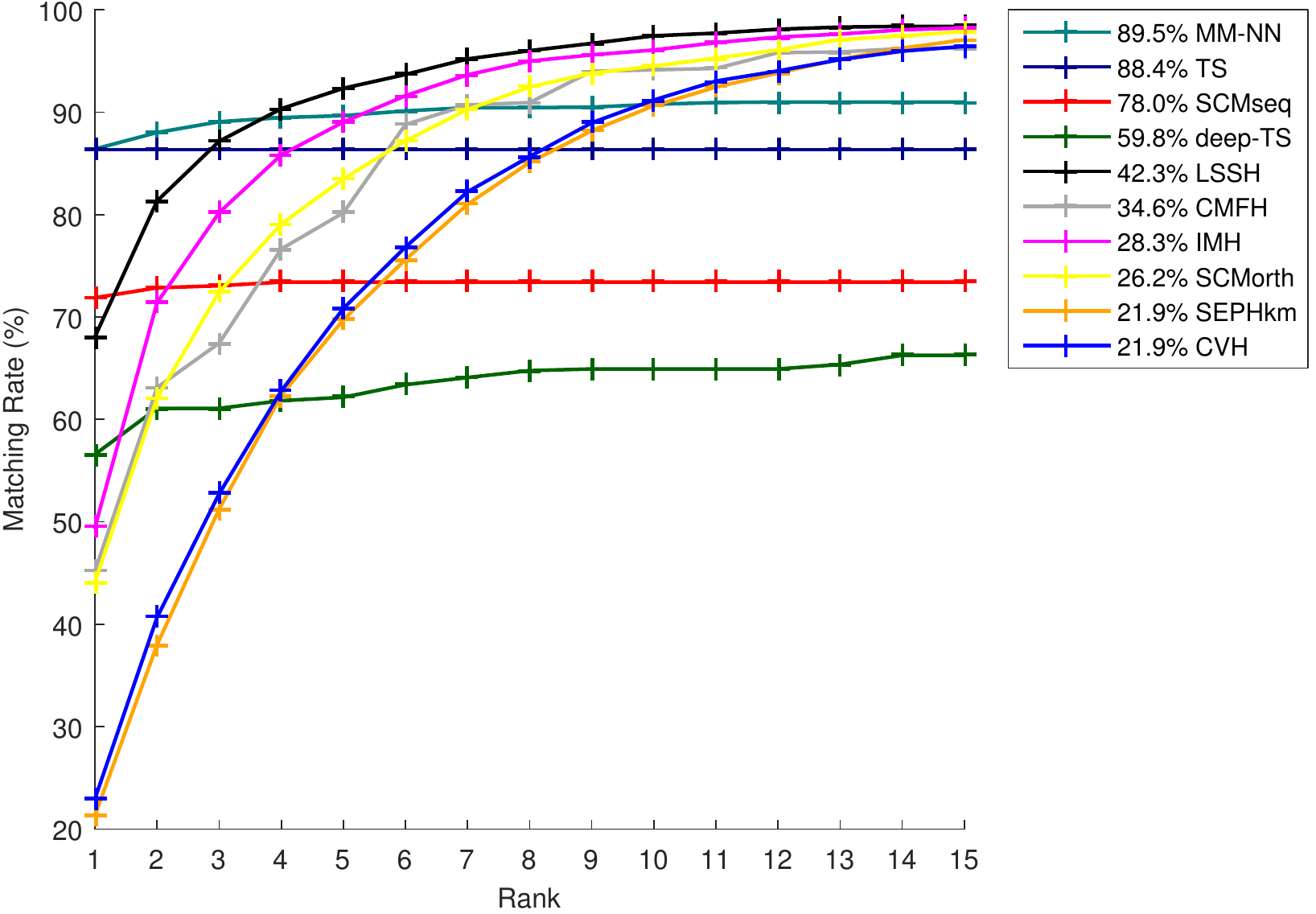}\label{subfig:wiki_t2i1_hs32}}
\subfigure[XTD retrieval: T$ \rightarrow $I (32b)]{\includegraphics[width=0.24\linewidth]{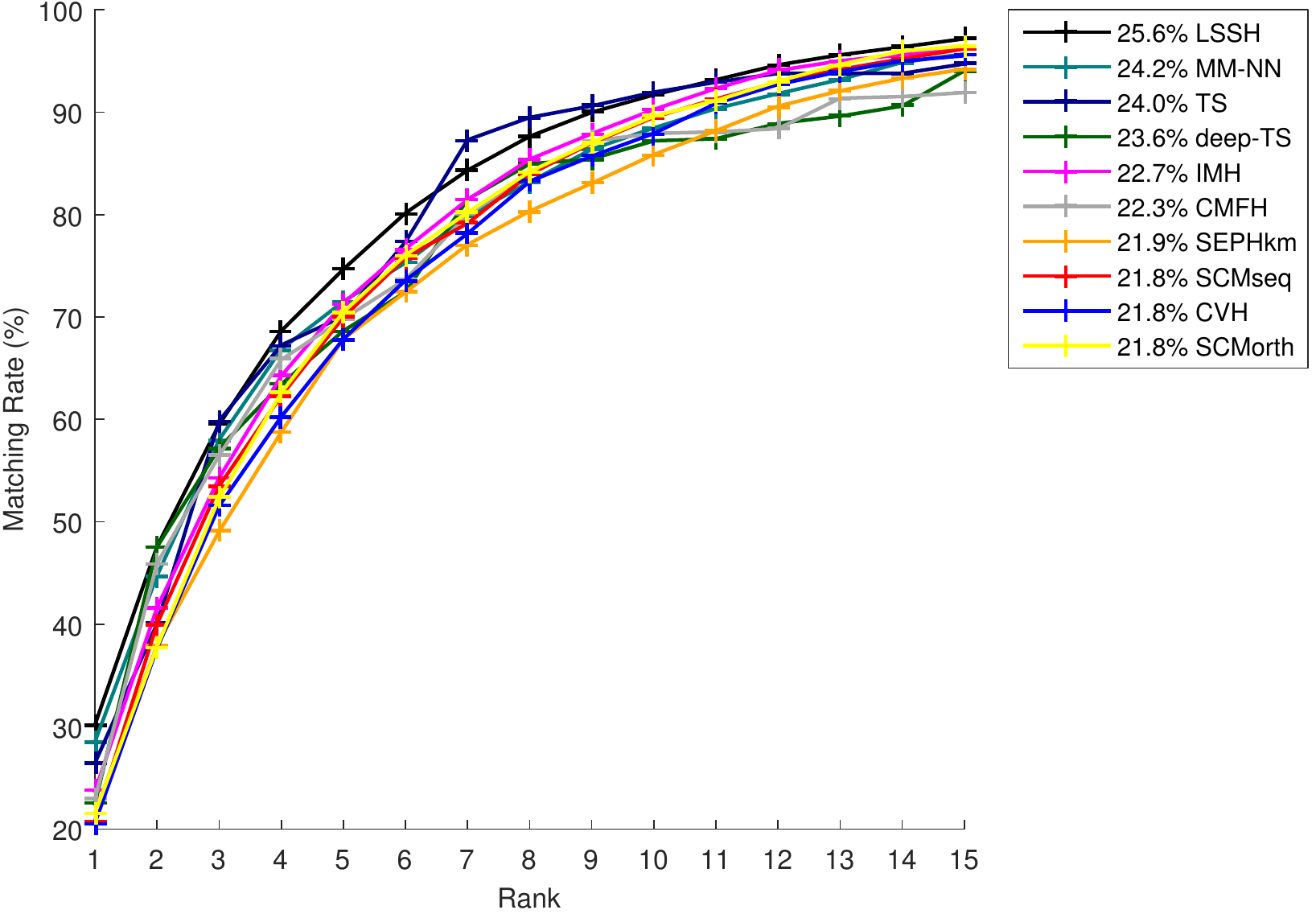}}
\end{center}
	\caption{Evaluation results of binary representations on Wikipedia. CMC curves are shown. MAP is shown before the name of each method. (a)(c)(e)(g)(i)(k) represent the non-extendable retrieval results, while (b)(d)(f)(h)(j)(l) are the extendable retrieval results. (a)(b)(e)(f)(i)(j) denote image-to-text retrieval, while (c)(d)(g)(h)(k)(l) denote text-to-image retrieval. (a)(b)(c)(d) are the results of 8-bit hash codes, (e)(f)(g)(h) are the results of 16-bit hash codes and (i)(j)(k)(l) are the results of 32-bit hash codes.}
	\label{fig:hsWiki}
\end{figure*}

\begin{figure*}[!t]
\begin{center}
\subfigure[Non-XTD retrieval: I$ \rightarrow $T (8b)]{\includegraphics[width=0.24\linewidth]{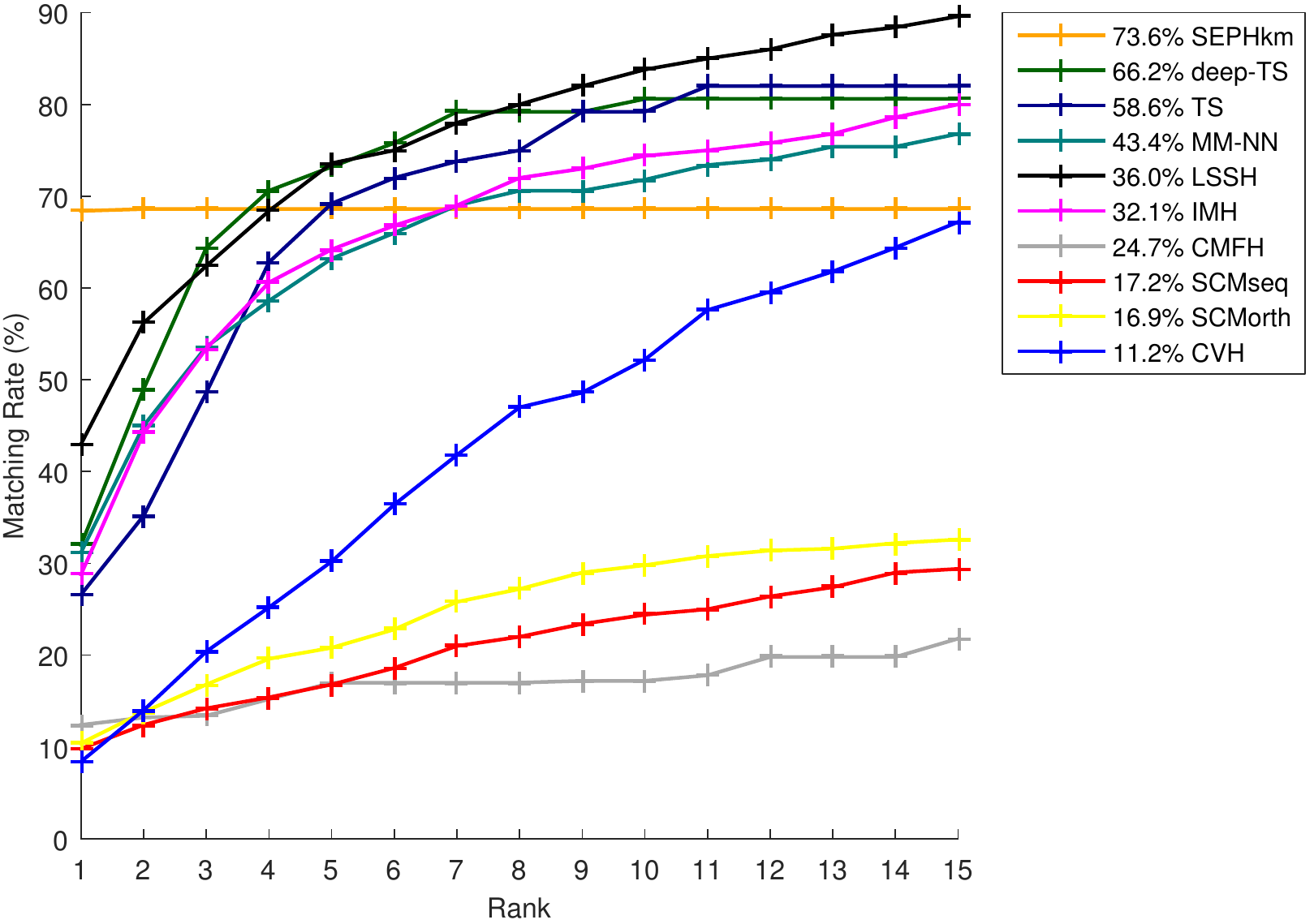}\label{subfig:pascal_i2t1_hs8}}
\subfigure[XTD retrieval: I$ \rightarrow $T (8b)]{\includegraphics[width=0.24\linewidth]{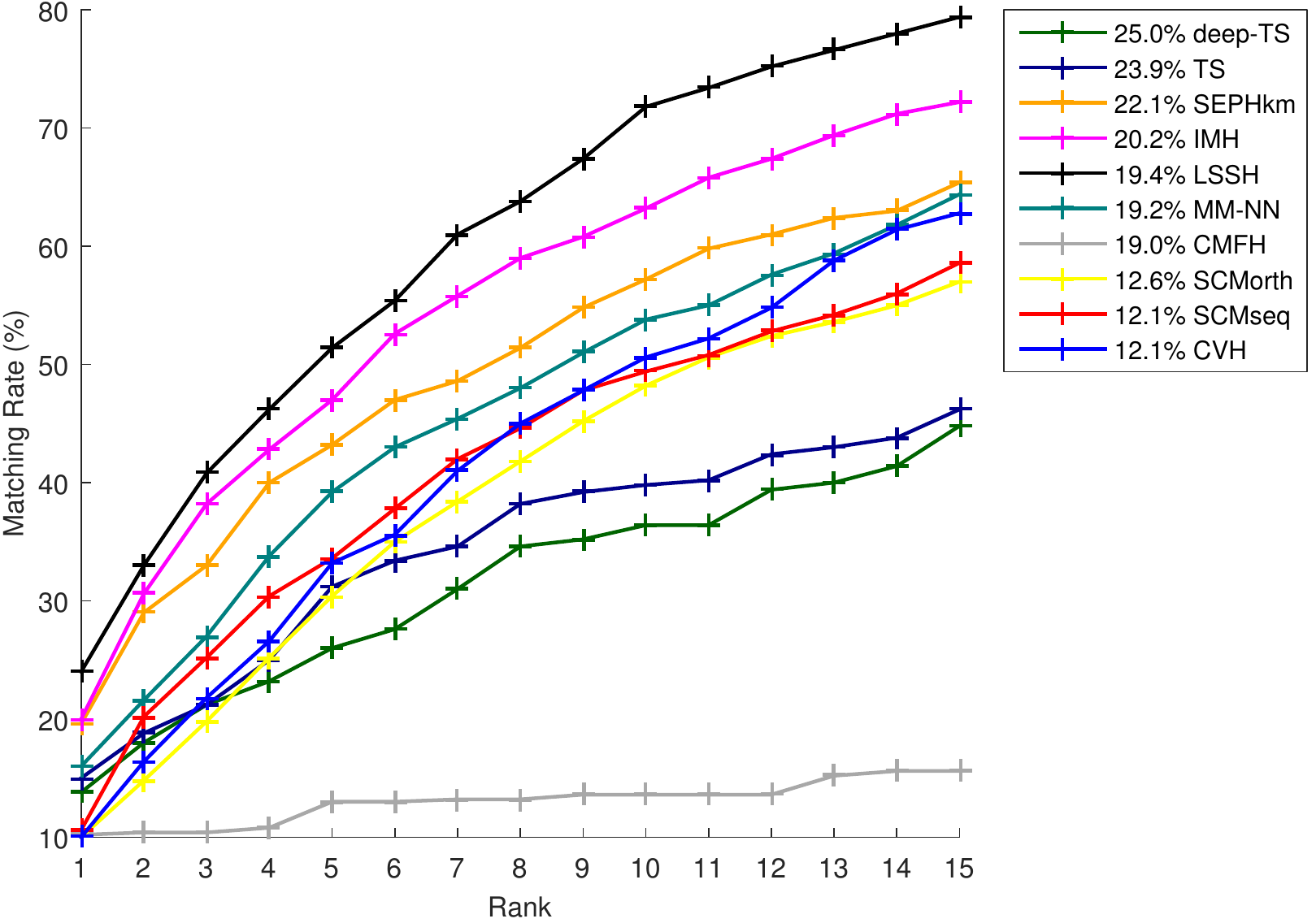}}
\subfigure[Non-XTD retrieval: T$ \rightarrow $I (8b)]{\includegraphics[width=0.24\linewidth]{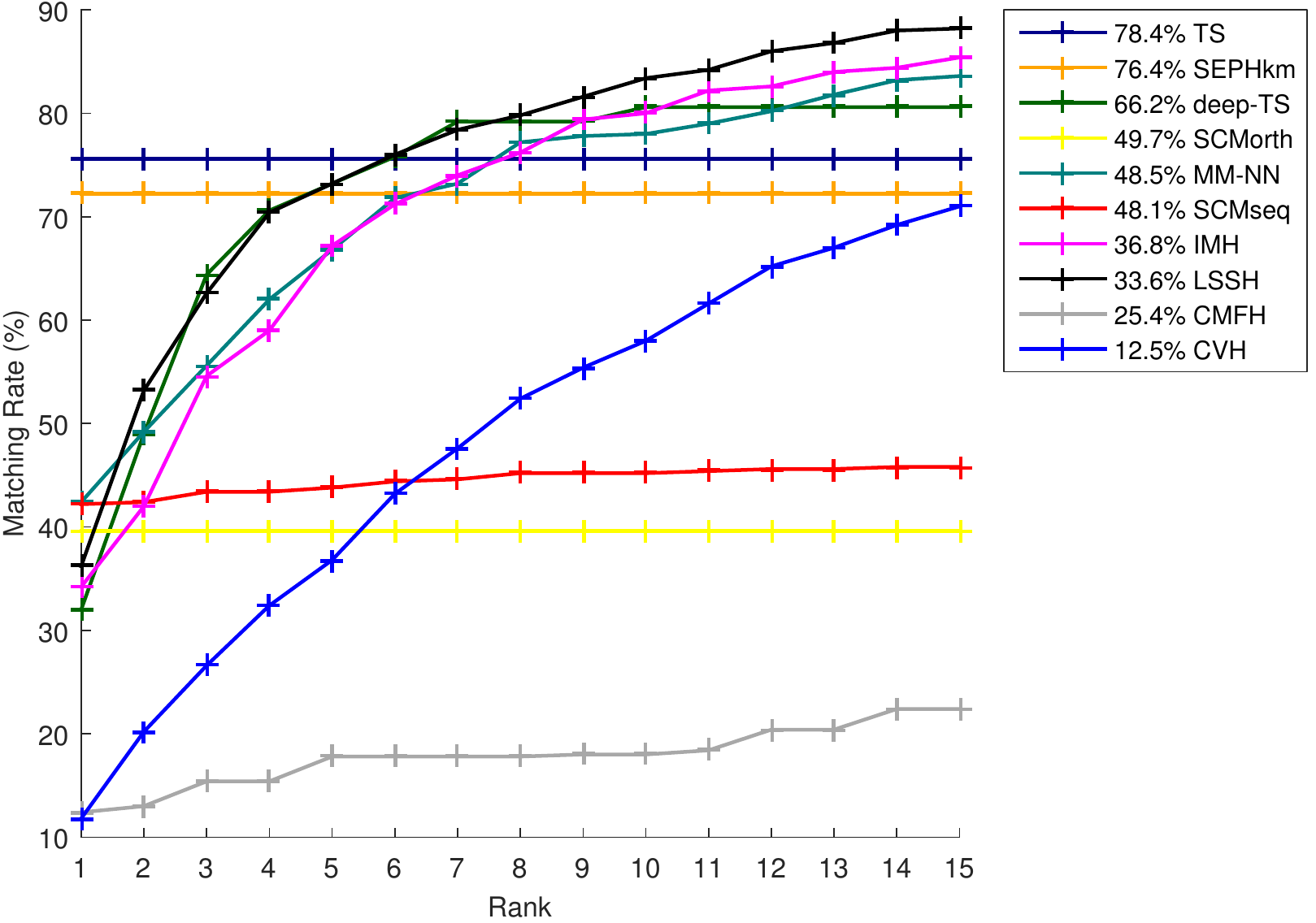}\label{subfig:pascal_t2i1_hs8}}
\subfigure[XTD retrieval: T$ \rightarrow $I (8b)]{\includegraphics[width=0.24\linewidth]{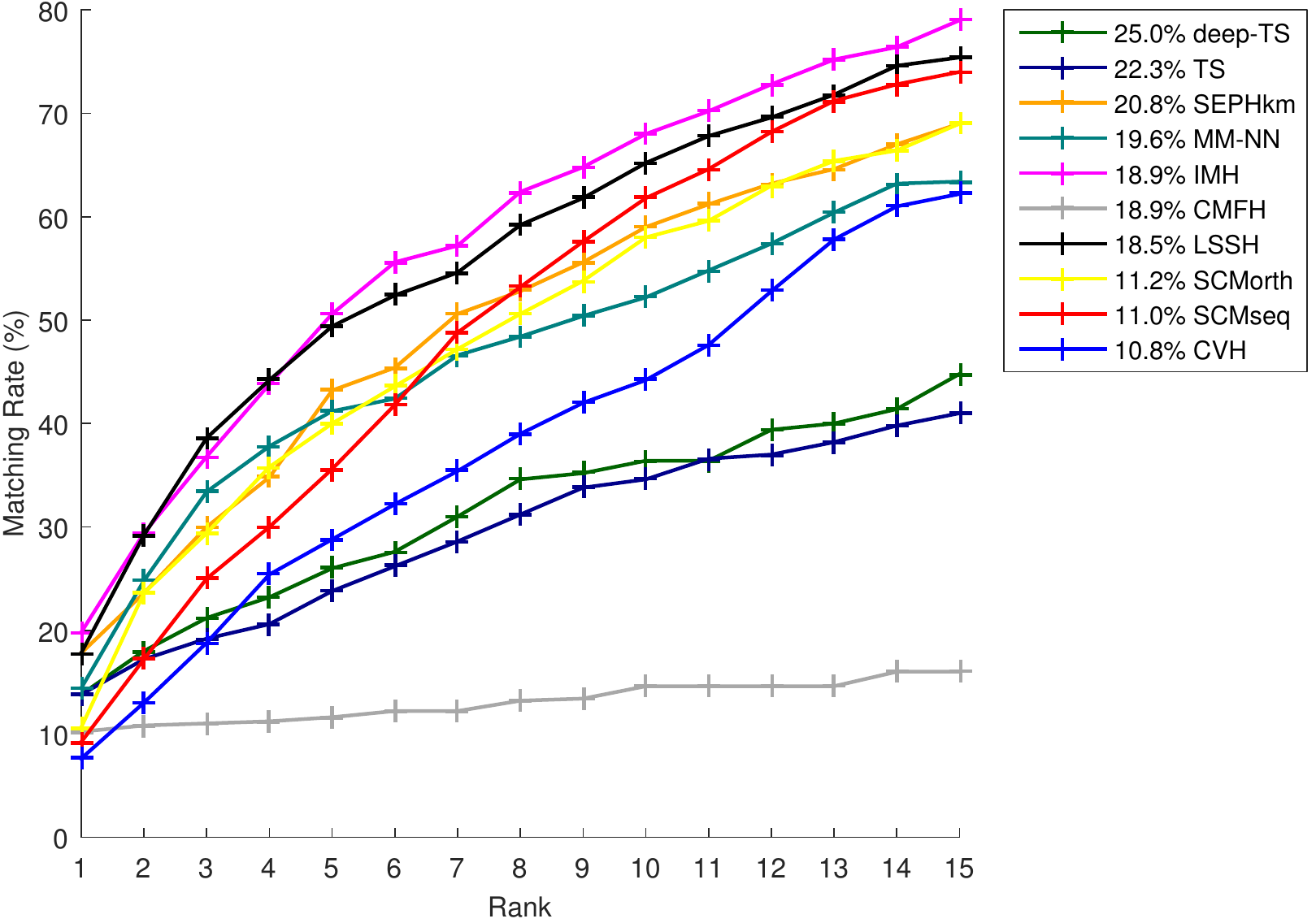}}
\subfigure[Non-XTD retrieval: I$ \rightarrow $T (16b)]{\includegraphics[width=0.24\linewidth]{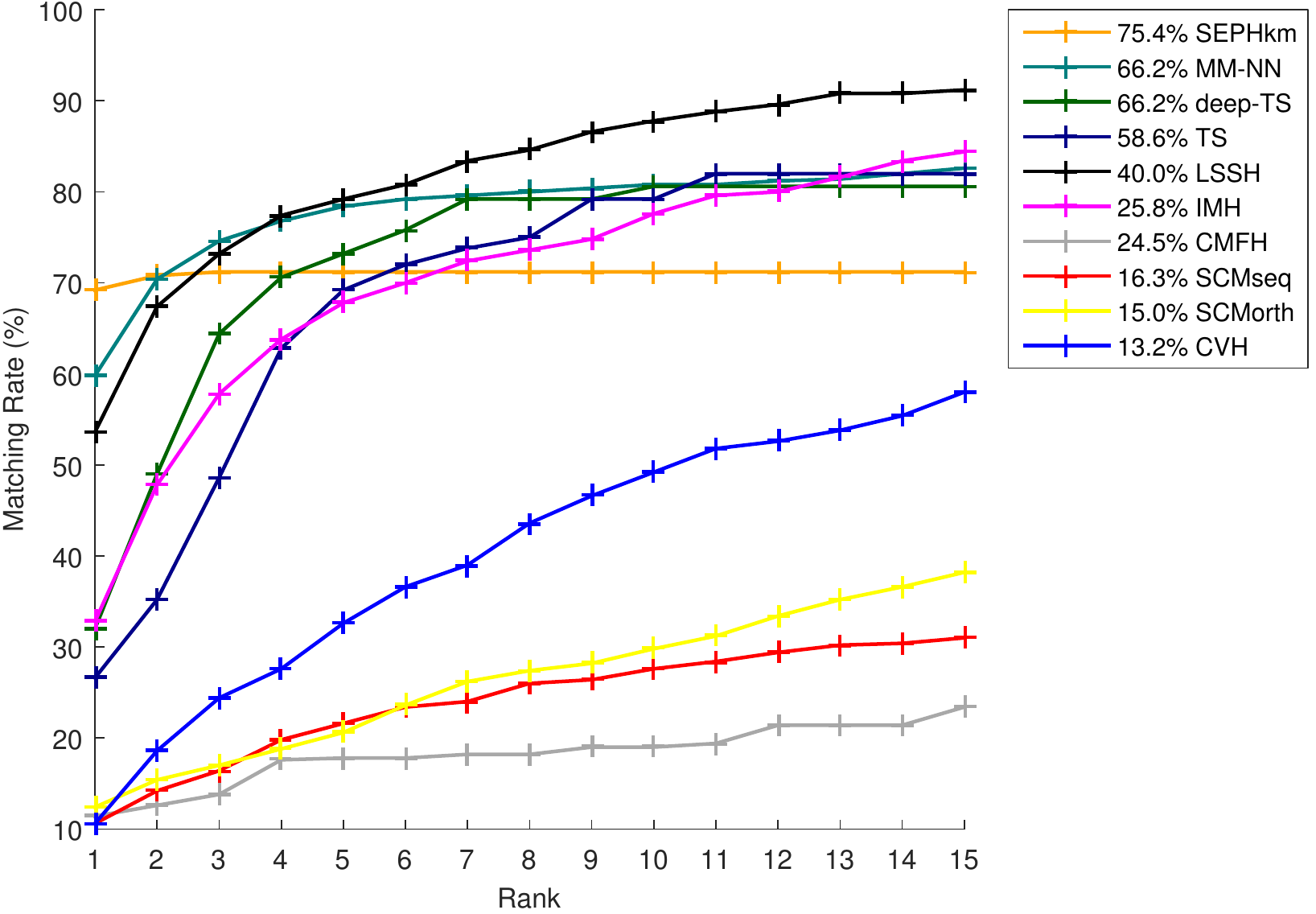}\label{subfig:pascal_i2t1_hs16}}
\subfigure[XTD retrieval: I$ \rightarrow $T (16b)]{\includegraphics[width=0.24\linewidth]{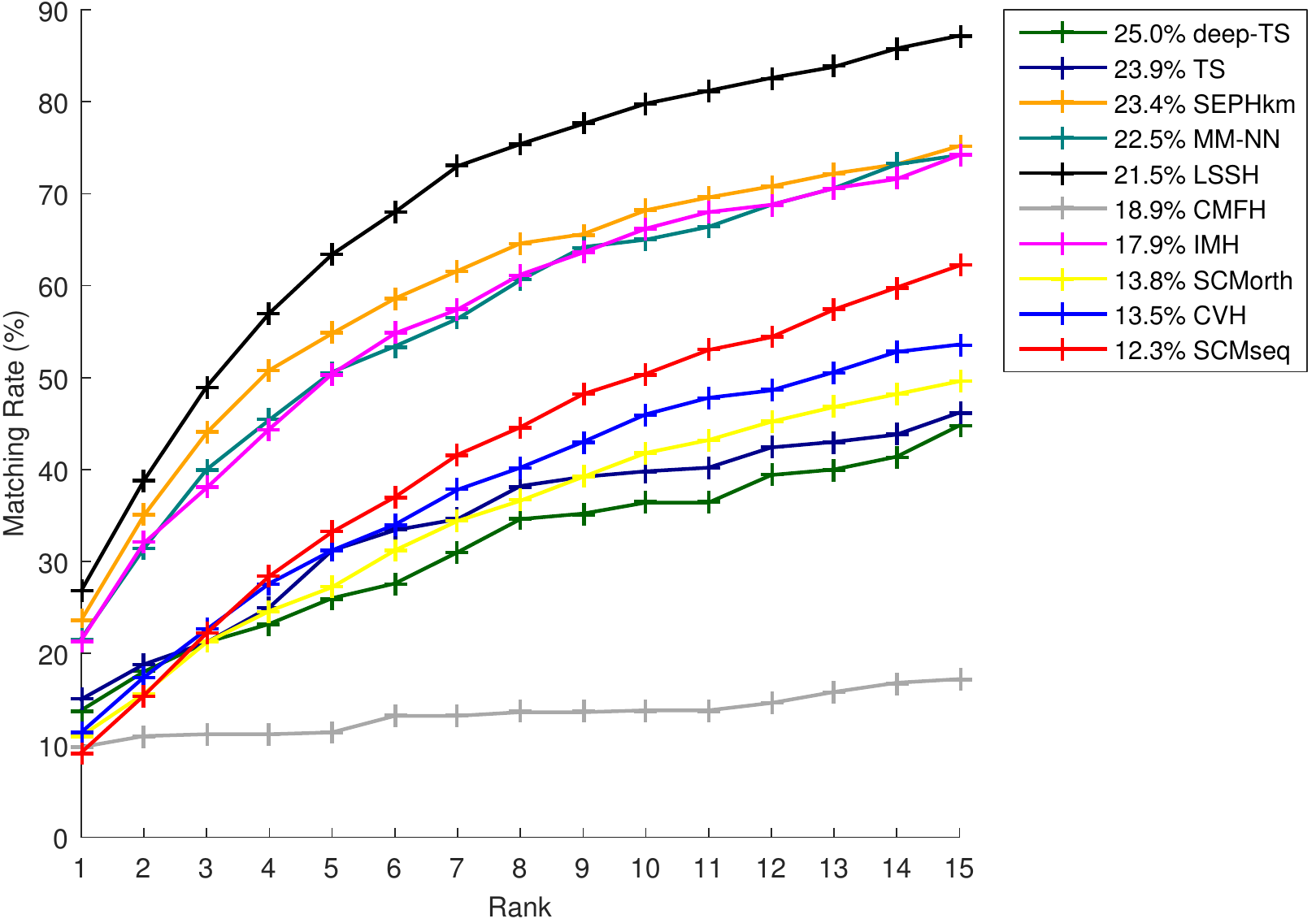}}
\subfigure[Non-XTD retrieval: T$ \rightarrow $I (16b)]{\includegraphics[width=0.24\linewidth]{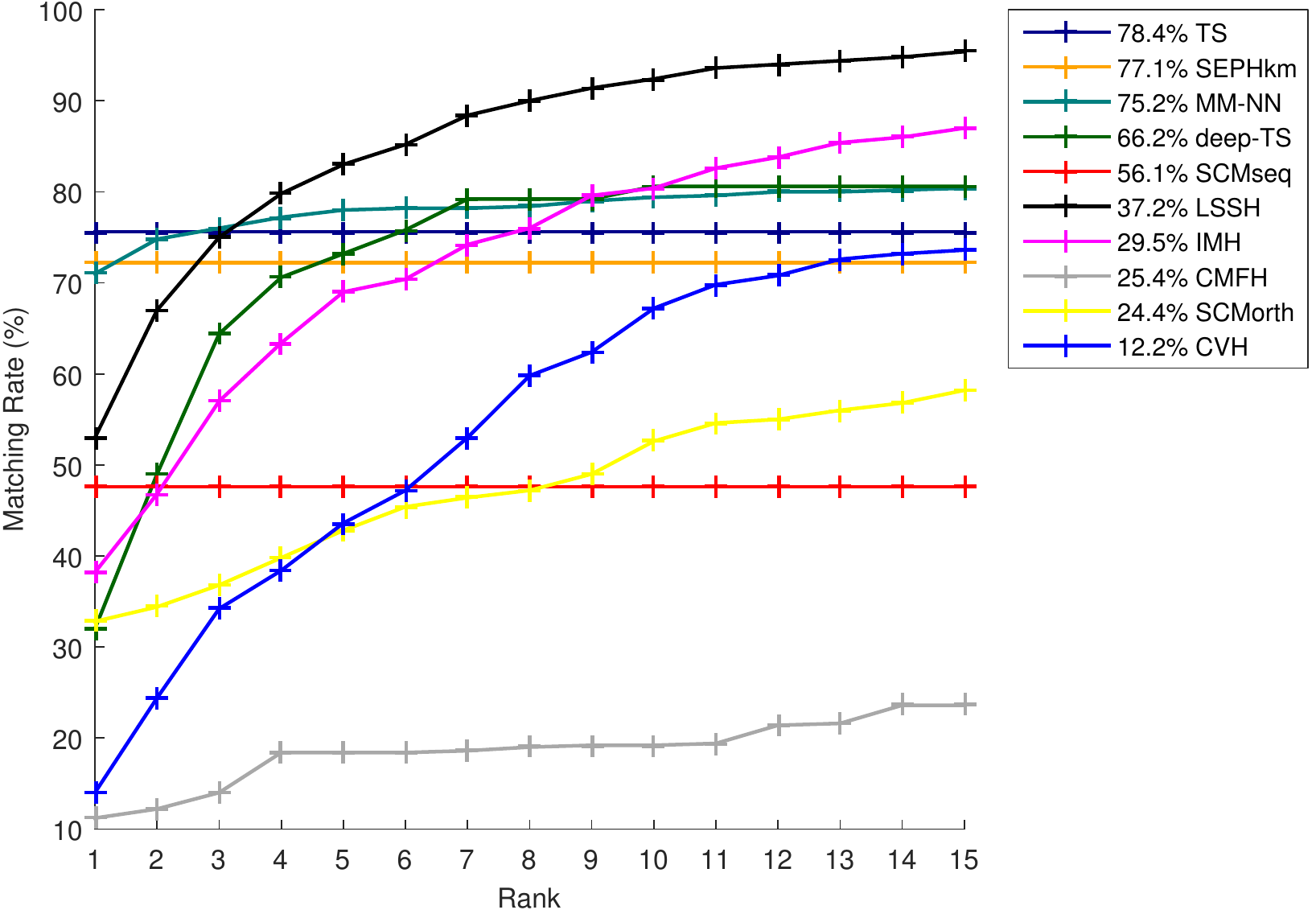}\label{subfig:pascal_t2i1_hs16}}
\subfigure[XTD retrieval: T$ \rightarrow $I (16b)]{\includegraphics[width=0.24\linewidth]{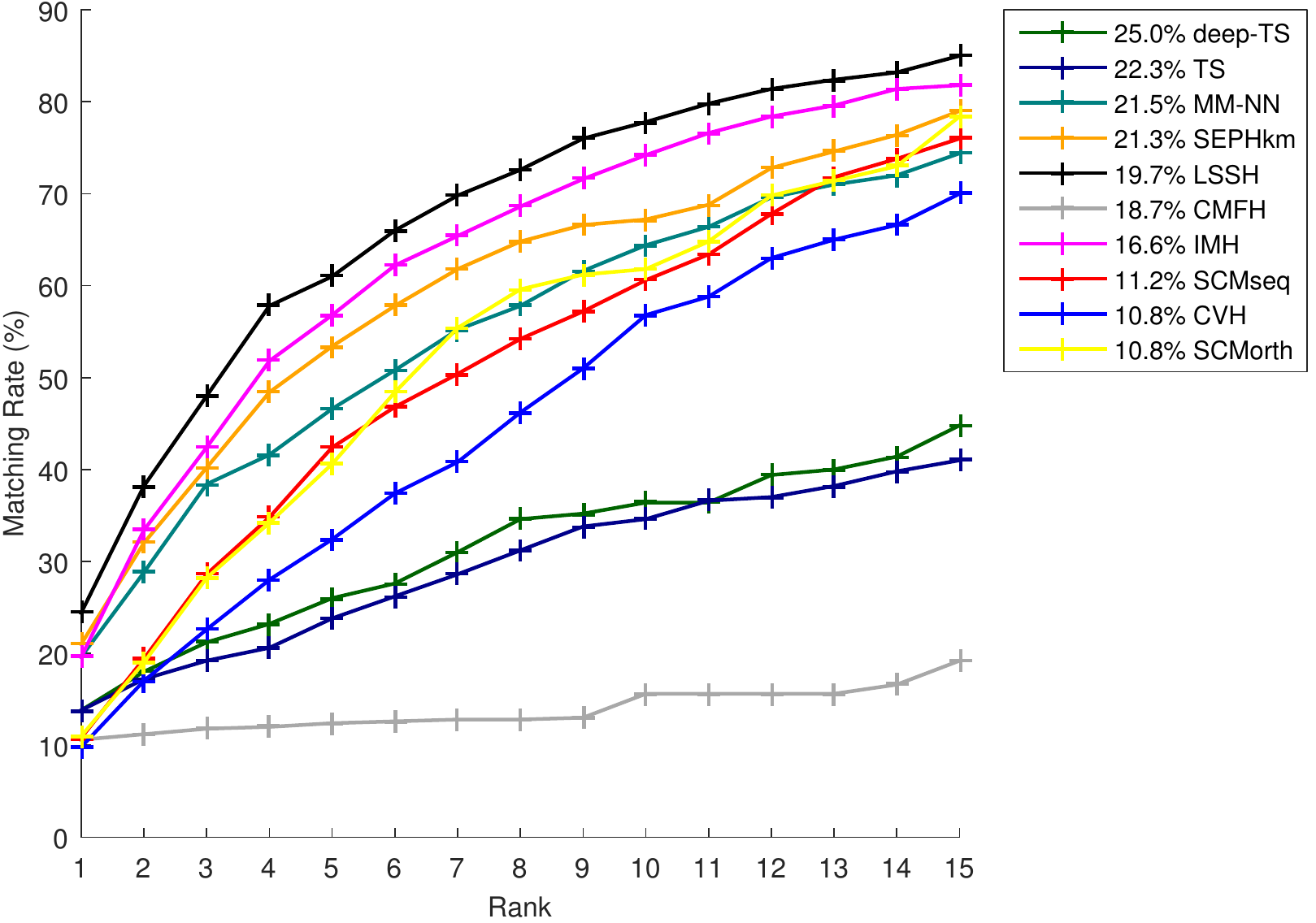}}
\subfigure[Non-XTD retrieval: I$ \rightarrow $T (32b)]{\includegraphics[width=0.24\linewidth]{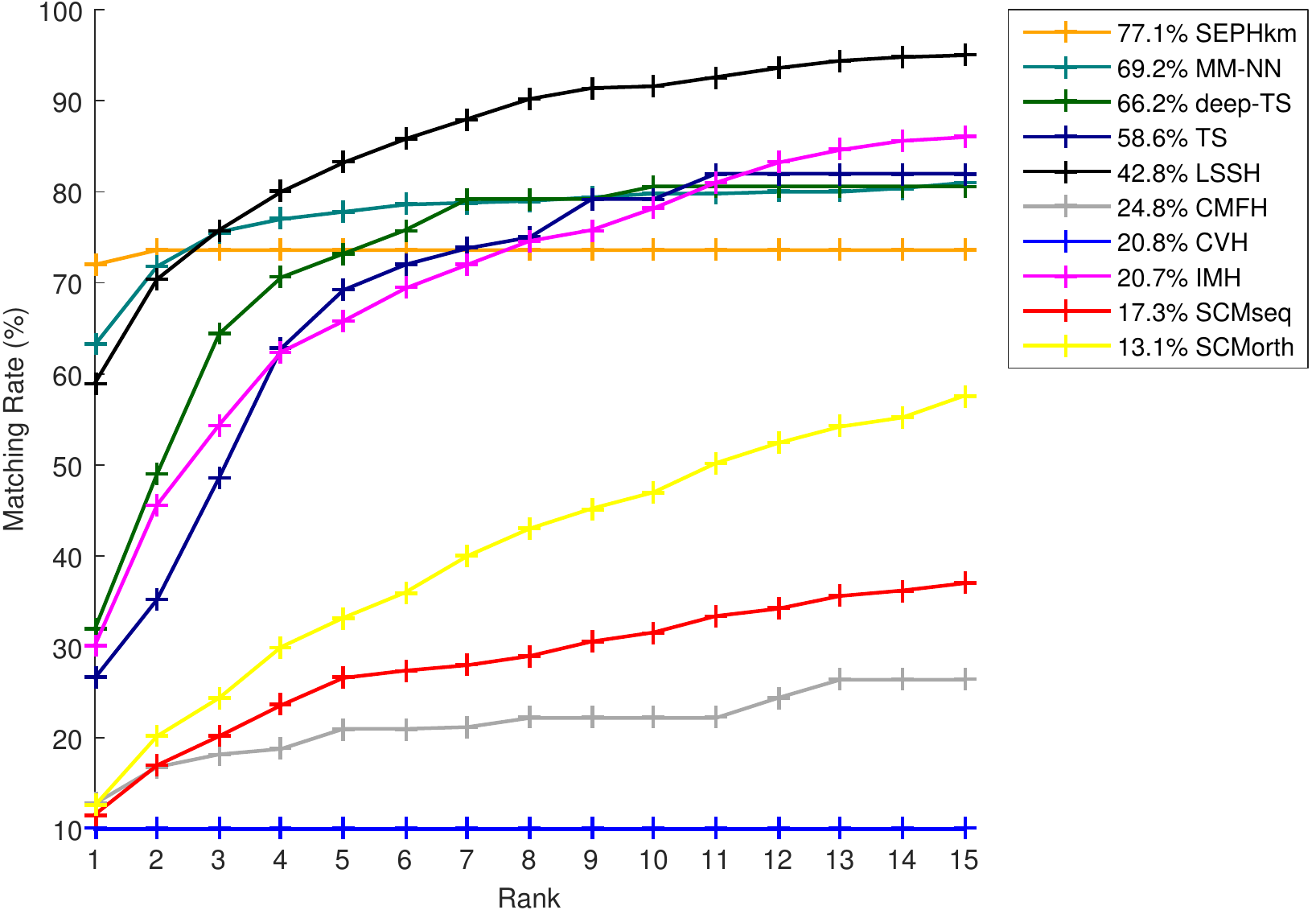}}
\subfigure[XTD retrieval: I$ \rightarrow $T (32b)]{\includegraphics[width=0.24\linewidth]{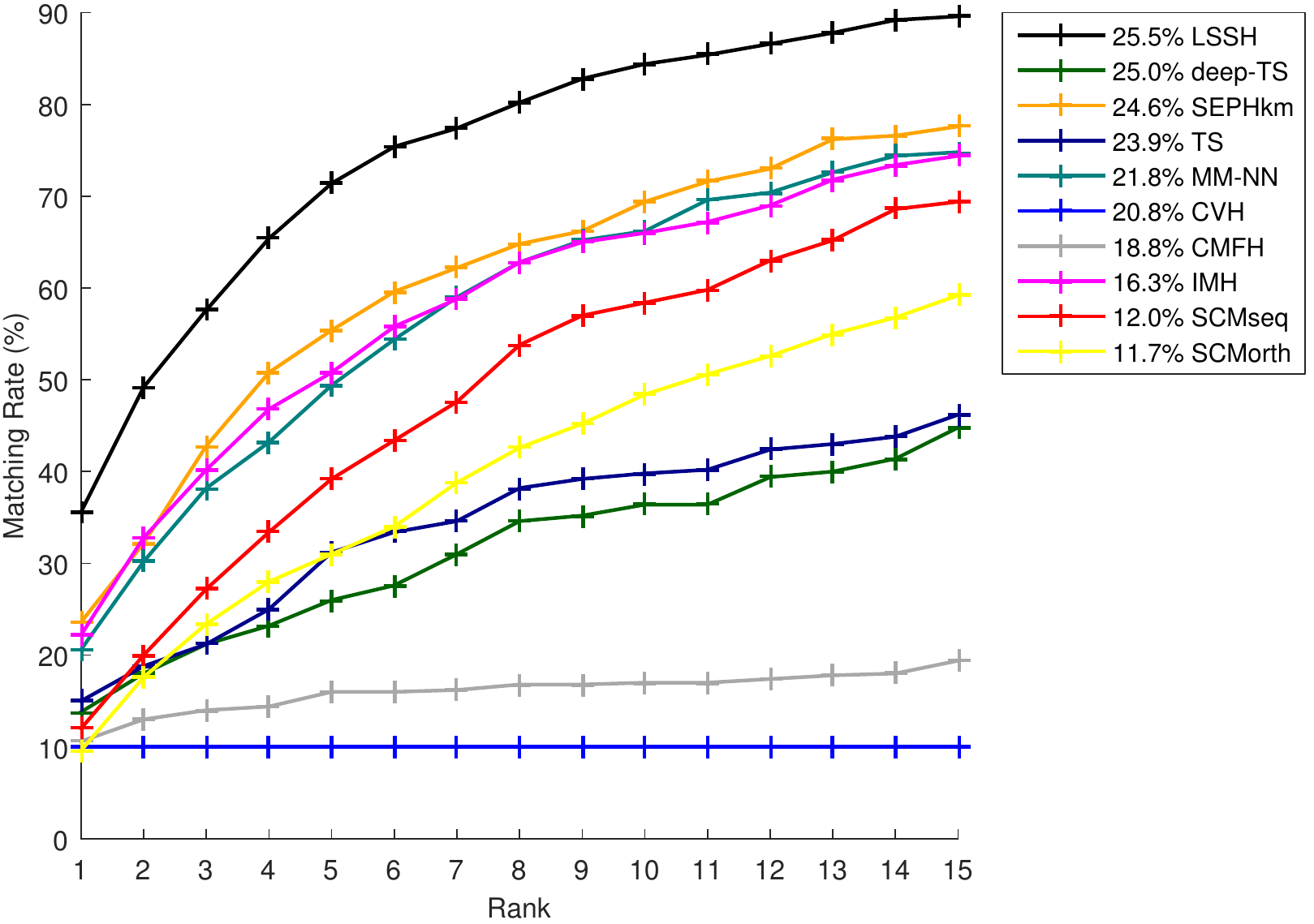}}
\subfigure[Non-XTD retrieval: T$ \rightarrow $I (32b)]{\includegraphics[width=0.24\linewidth]{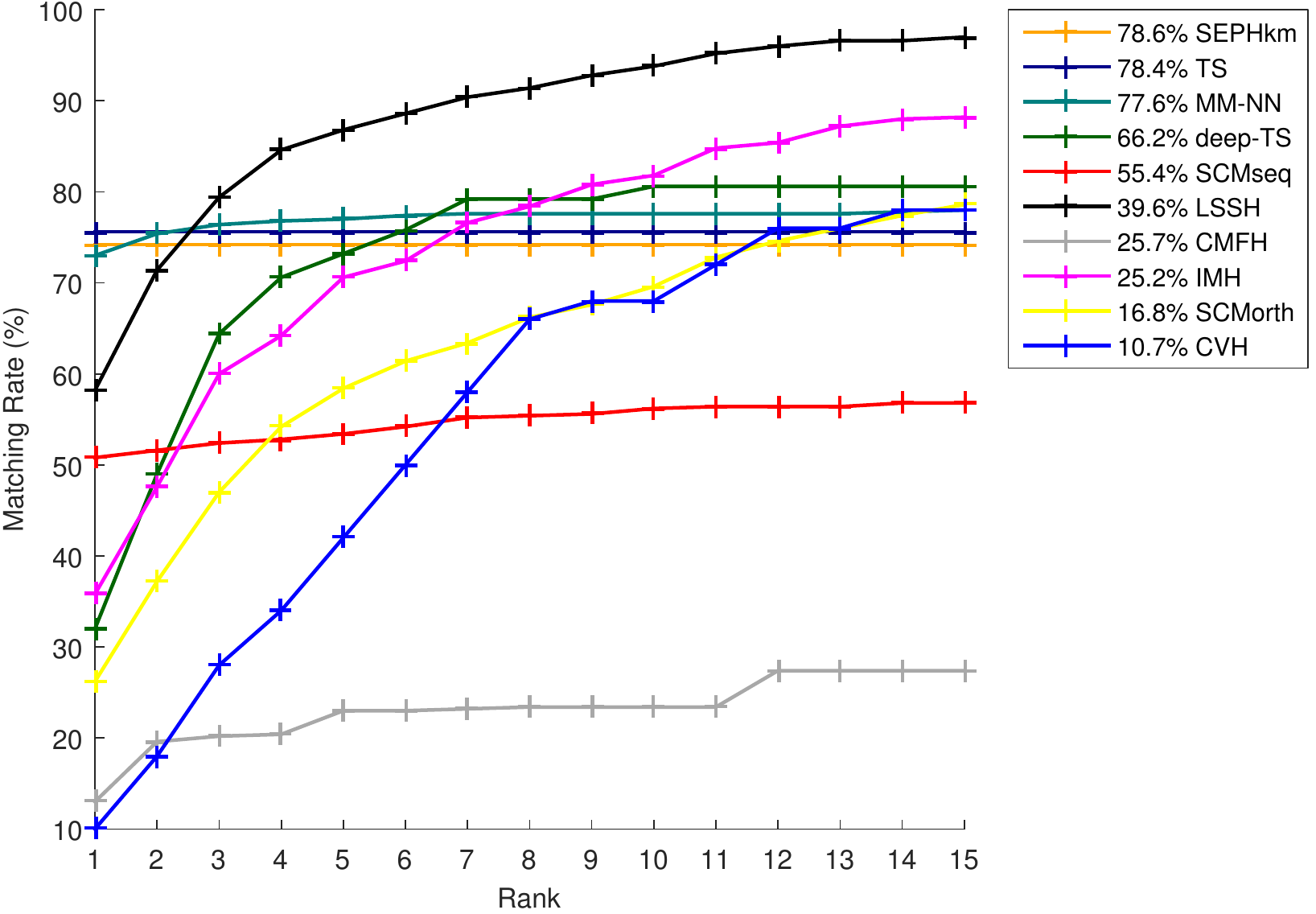}\label{subfig:pascal_t2i1_hs32}}
\subfigure[XTD retrieval: T$ \rightarrow $I (32b)]{\includegraphics[width=0.24\linewidth]{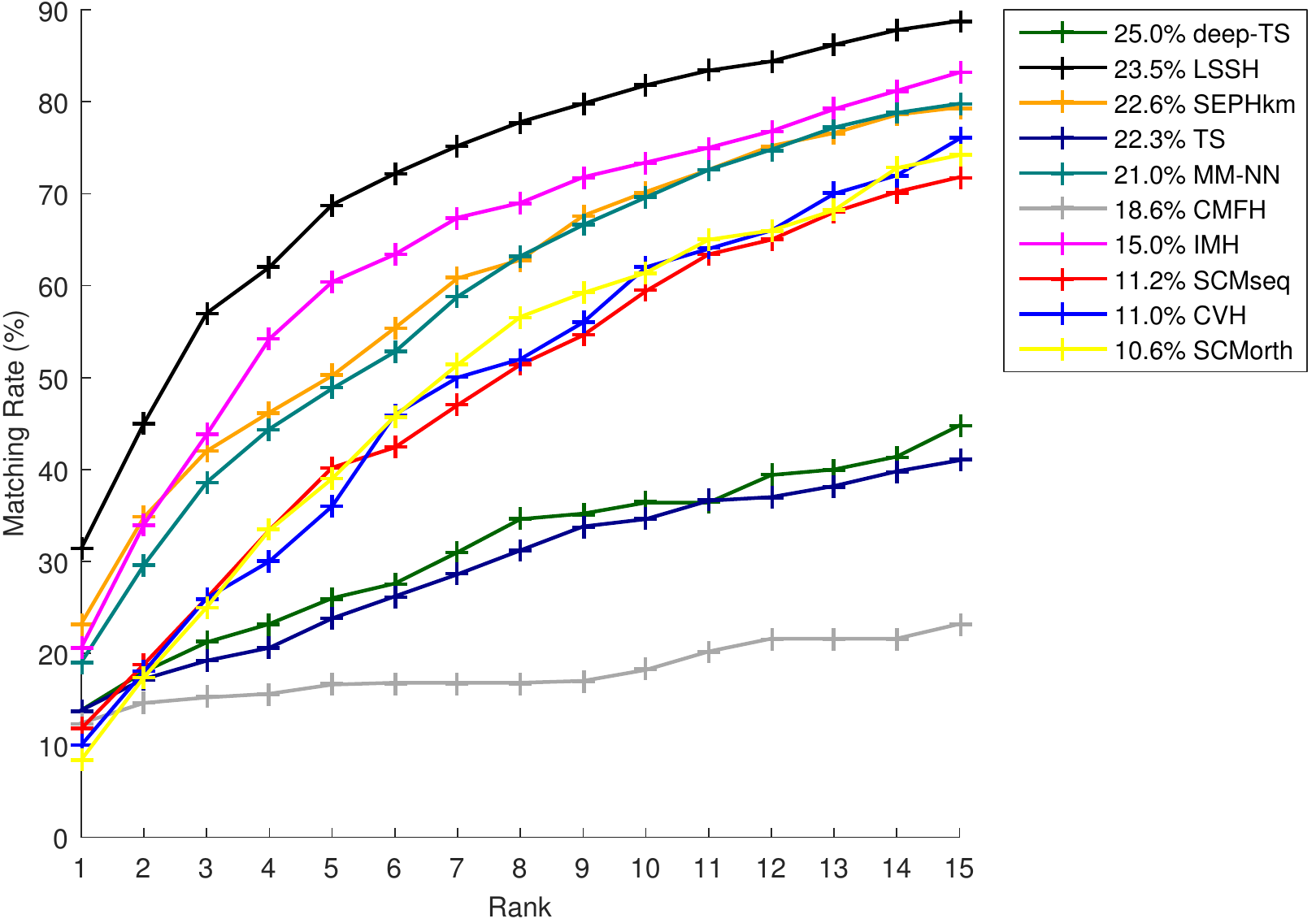}}
\end{center}
	\caption{Evaluation results of binary representations on Pascal Sentence. CMC curves are shown. MAP is shown before the name of each method. Retrieval modes in (a)(b)(c)(d)(e)(f)(g)(h)(i)(j)(k)(l) are the same with Fig. \ref{fig:hsWiki}.}
	\label{fig:hsPascal}
\end{figure*}

\begin{figure*}[!t]
\begin{center}
\subfigure[Non-xtd retrieval: I$ \rightarrow $T (8b)]{\includegraphics[width=0.24\linewidth]{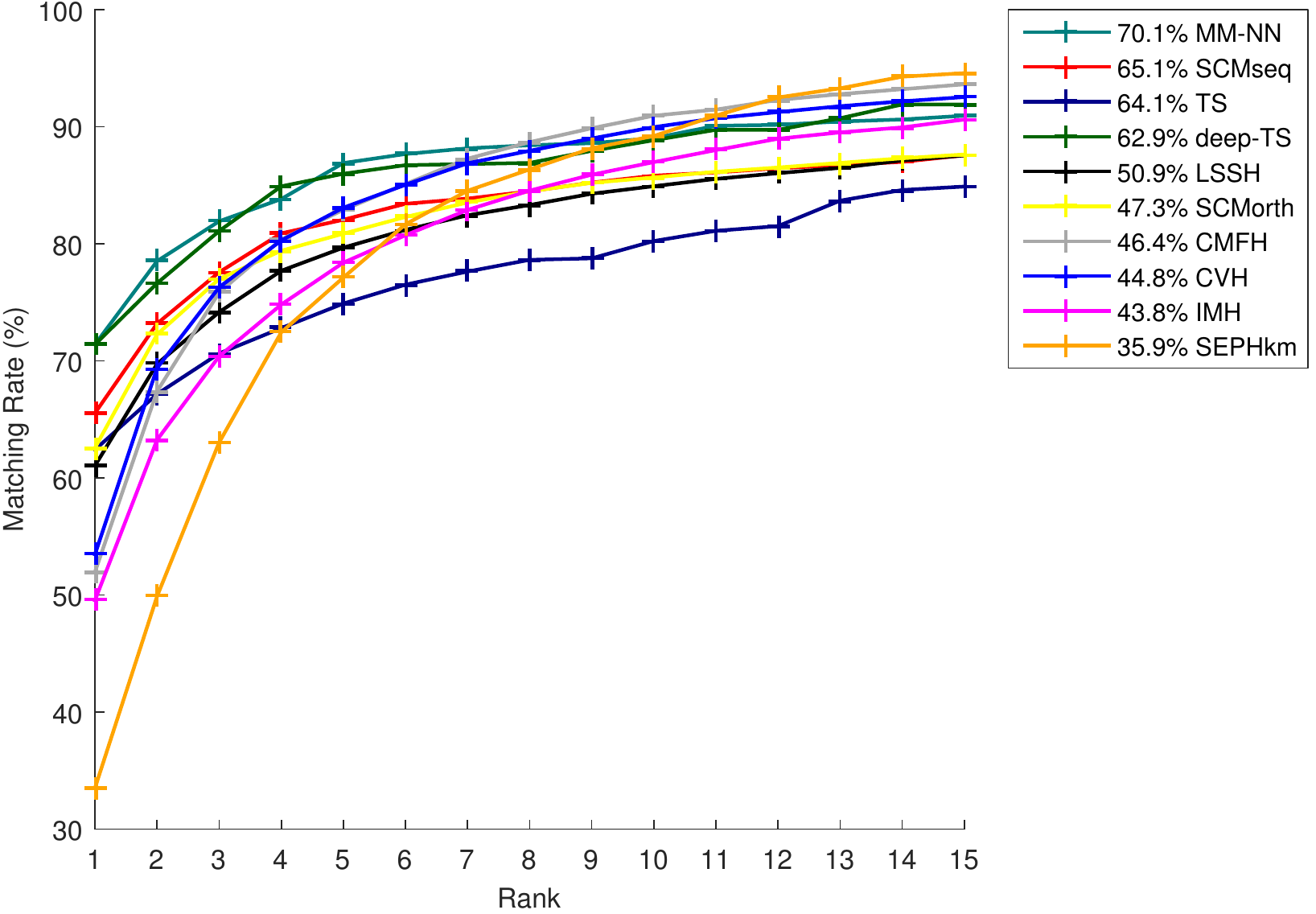}}
\subfigure[Extendable retrieval: I$ \rightarrow $T (8b)]{\includegraphics[width=0.24\linewidth]{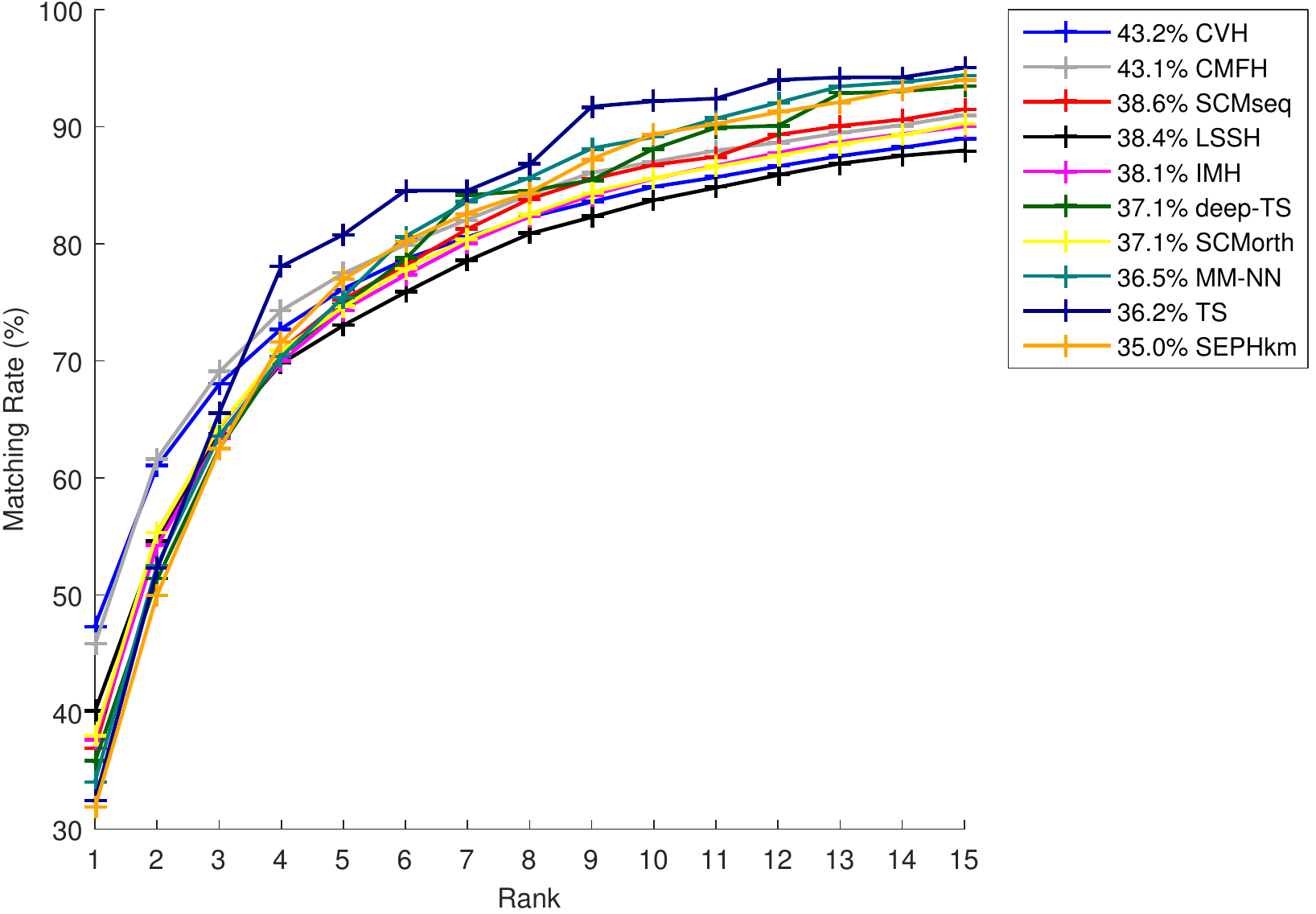}}
\subfigure[Non-xtd retrieval: T$ \rightarrow $I (8b)]{\includegraphics[width=0.24\linewidth]{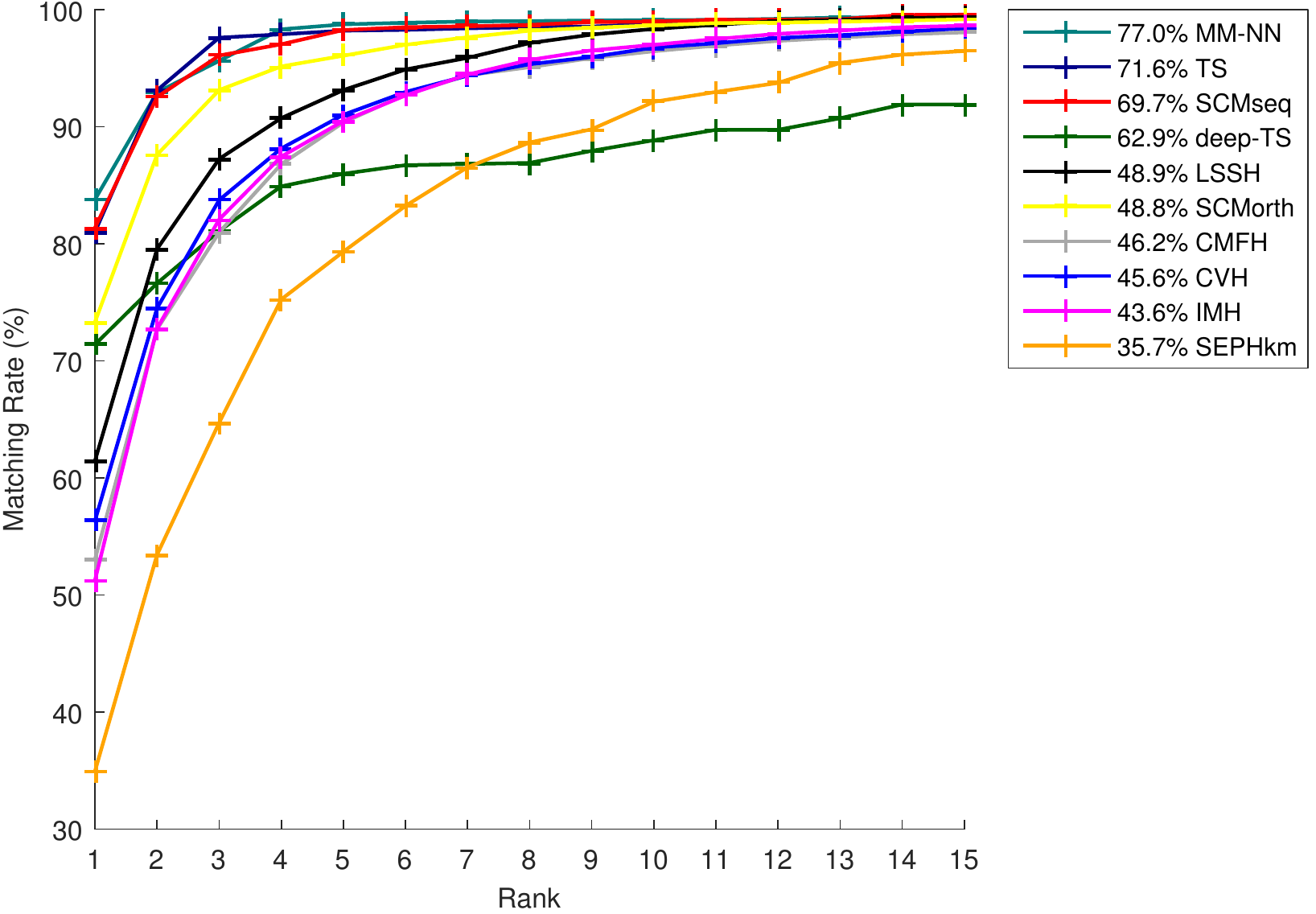}}
\subfigure[Extendable retrieval: T$ \rightarrow $I (8b)]{\includegraphics[width=0.24\linewidth]{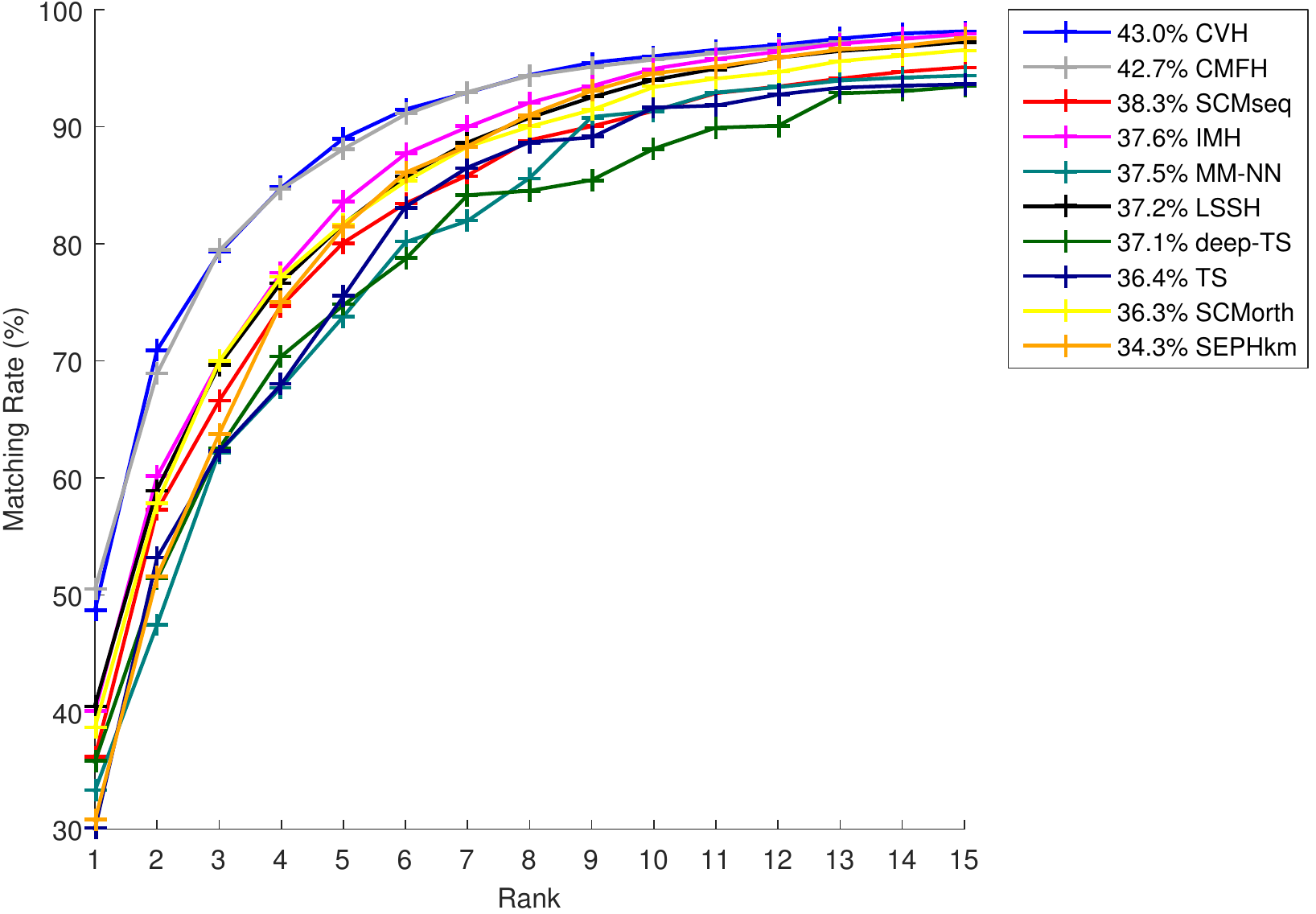}}
\subfigure[Non-xtd retrieval: I$ \rightarrow $T (16b)]{\includegraphics[width=0.24\linewidth]{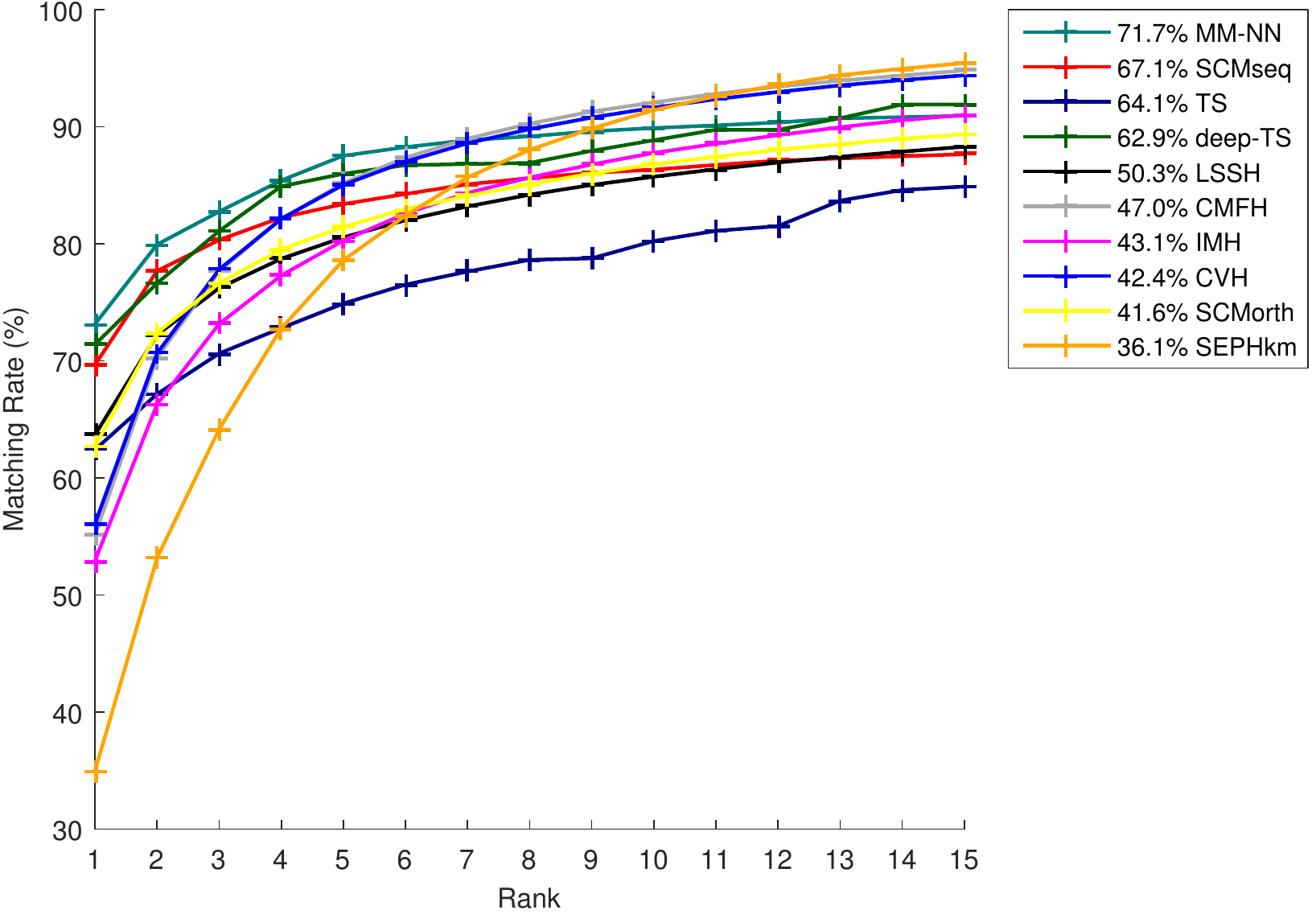}}
\subfigure[Extendable retrieval: I$ \rightarrow $T (16b)]{\includegraphics[width=0.24\linewidth]{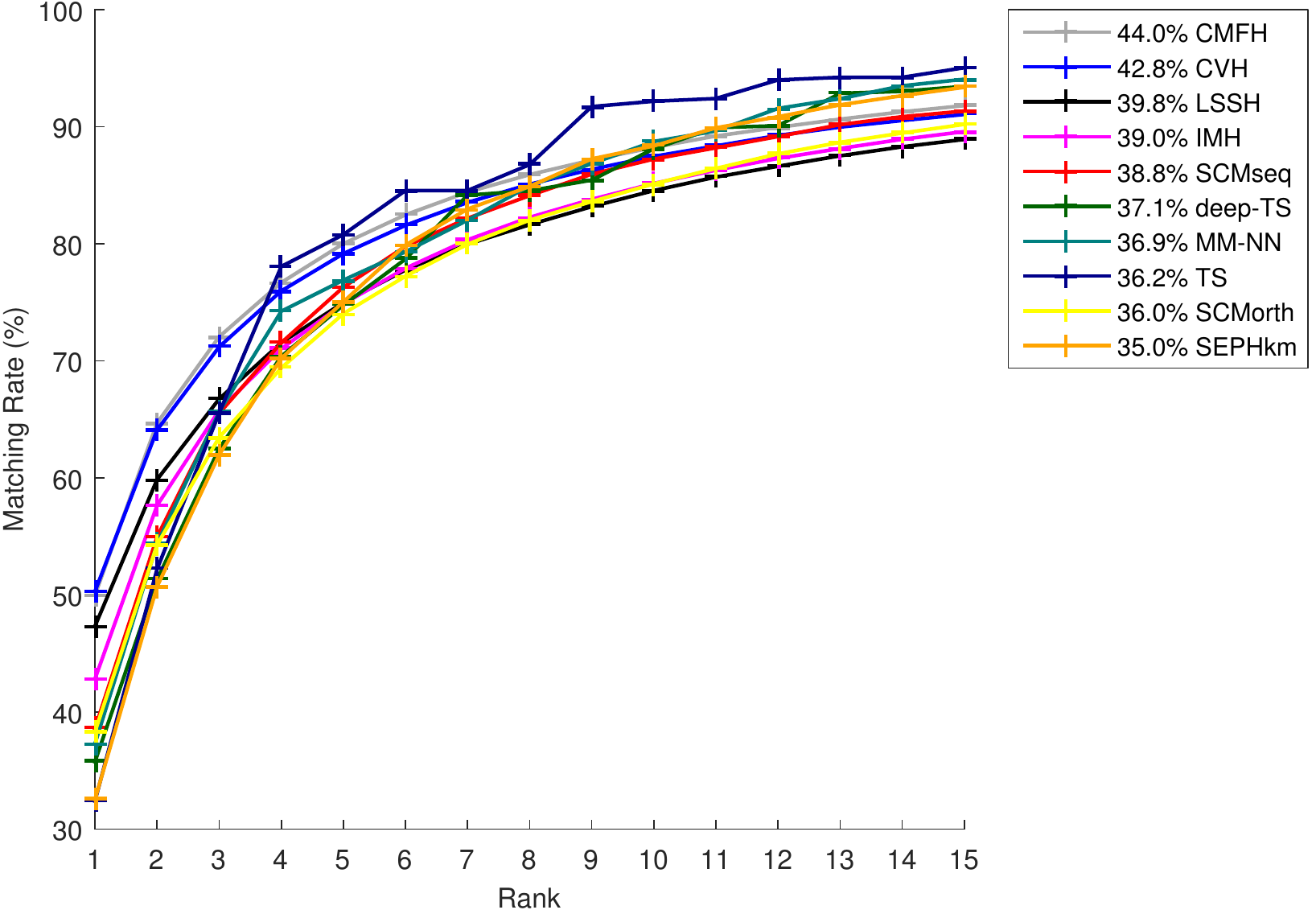}}
\subfigure[Non-xtd retrieval: T$ \rightarrow $I (16b)]{\includegraphics[width=0.24\linewidth]{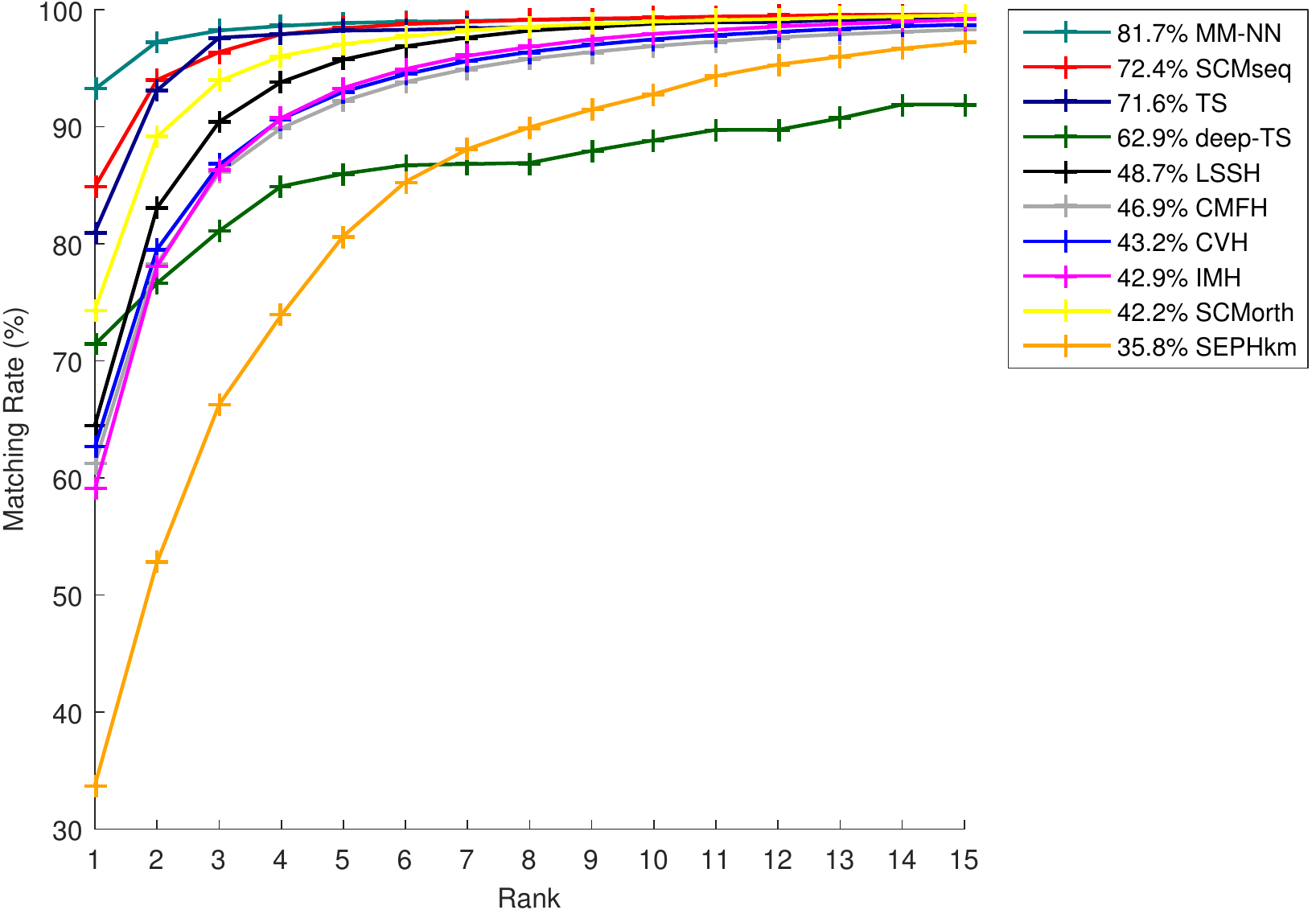}}
\subfigure[Extendable retrieval: T$ \rightarrow $I (16b)]{\includegraphics[width=0.24\linewidth]{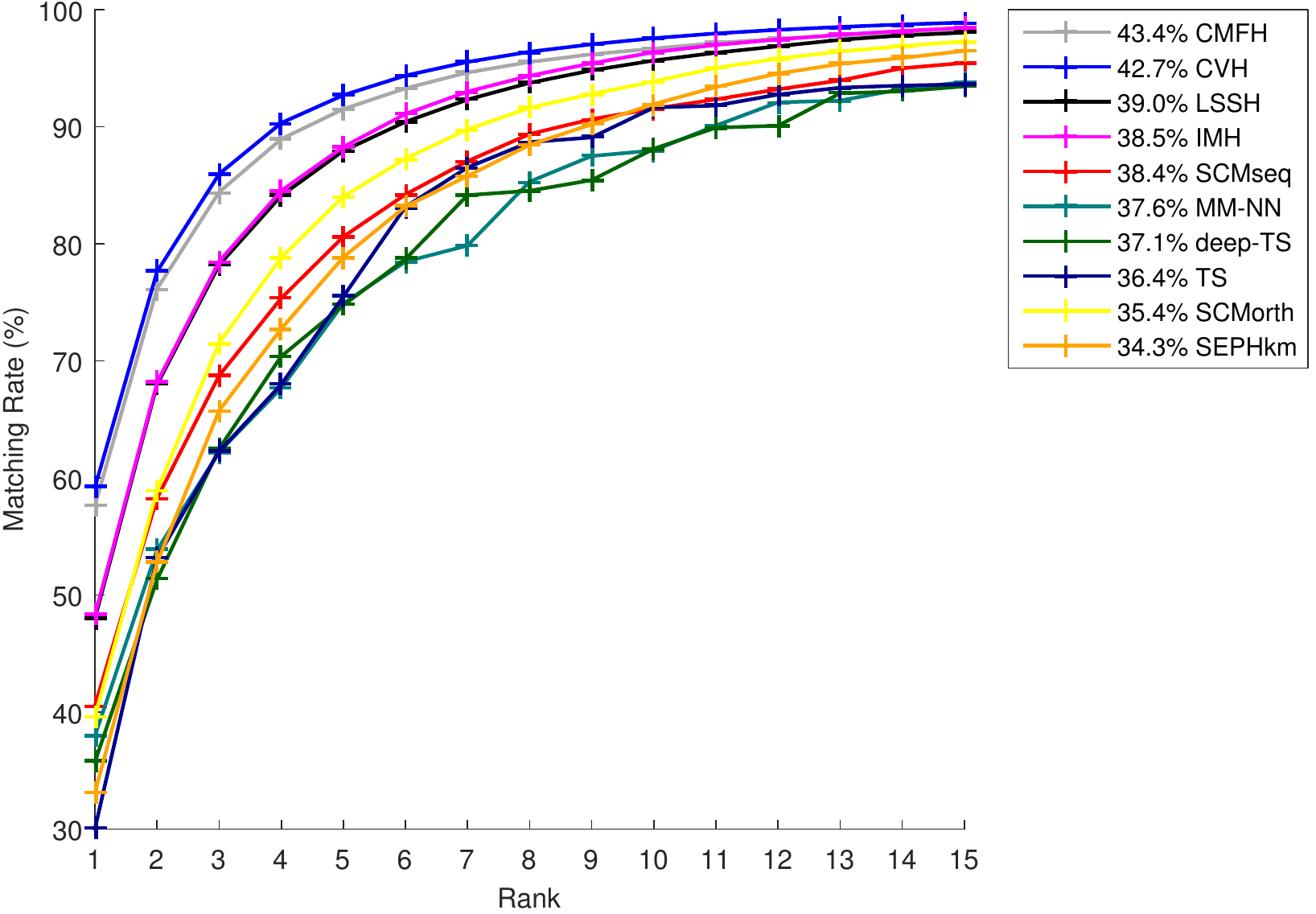}}
\subfigure[Non-xtd retrieval: I$ \rightarrow $T (32b)]{\includegraphics[width=0.24\linewidth]{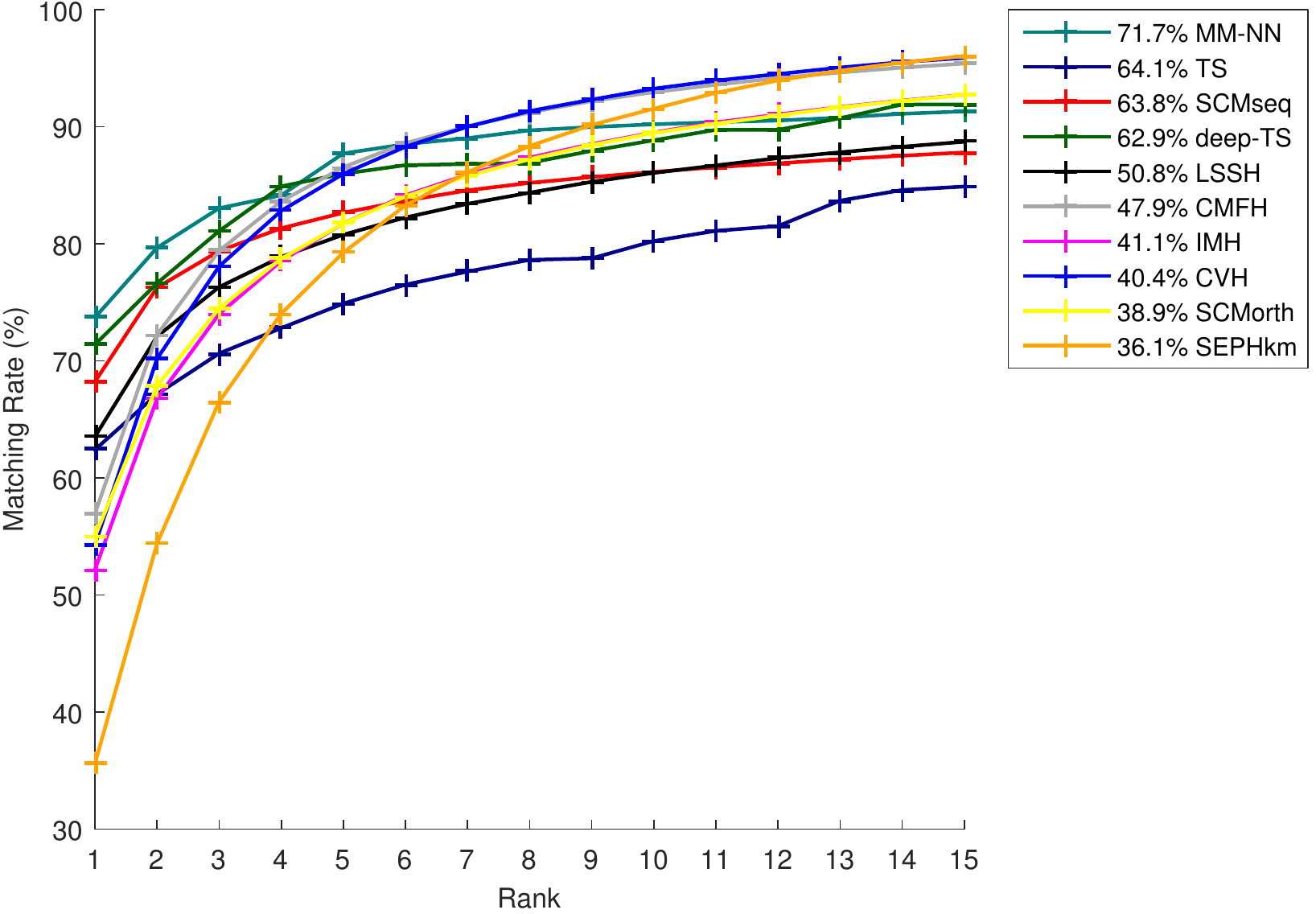}}
\subfigure[Extendable retrieval: I$ \rightarrow $T (32b)]{\includegraphics[width=0.24\linewidth]{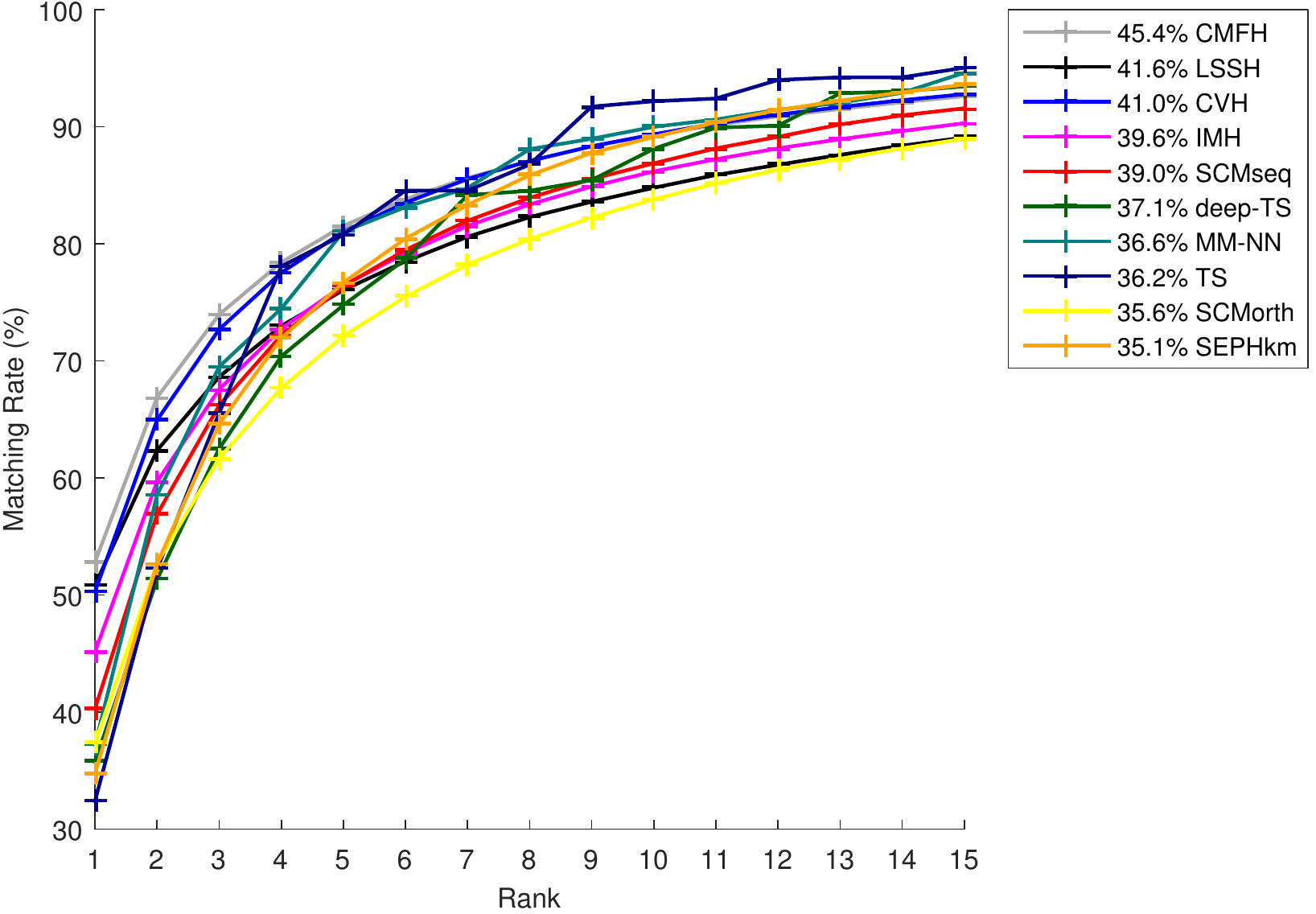}}
\subfigure[Non-xtd retrieval: T$ \rightarrow $I (32b)]{\includegraphics[width=0.24\linewidth]{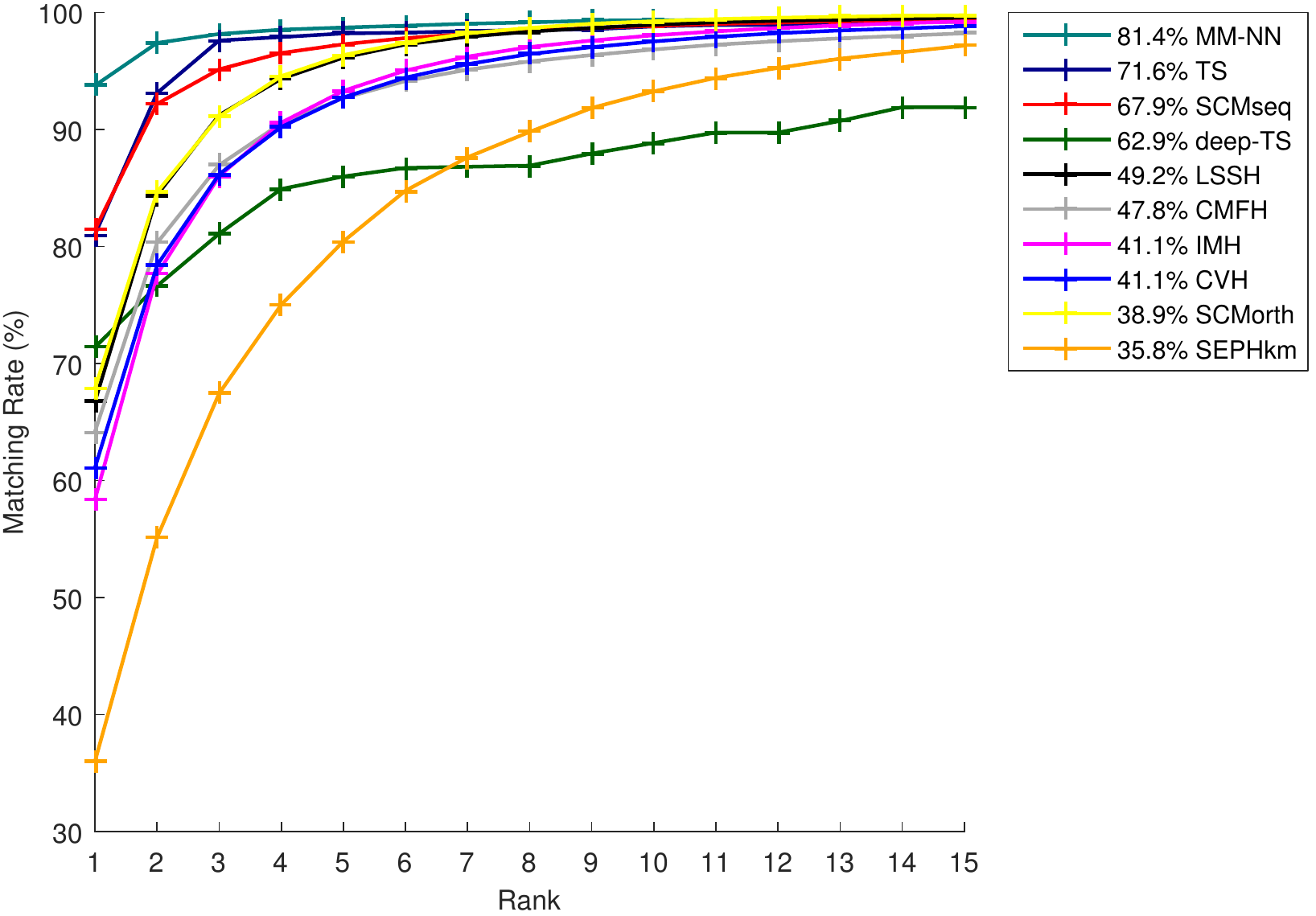}}
\subfigure[Extendable retrieval: T$ \rightarrow $I (32b)]{\includegraphics[width=0.24\linewidth]{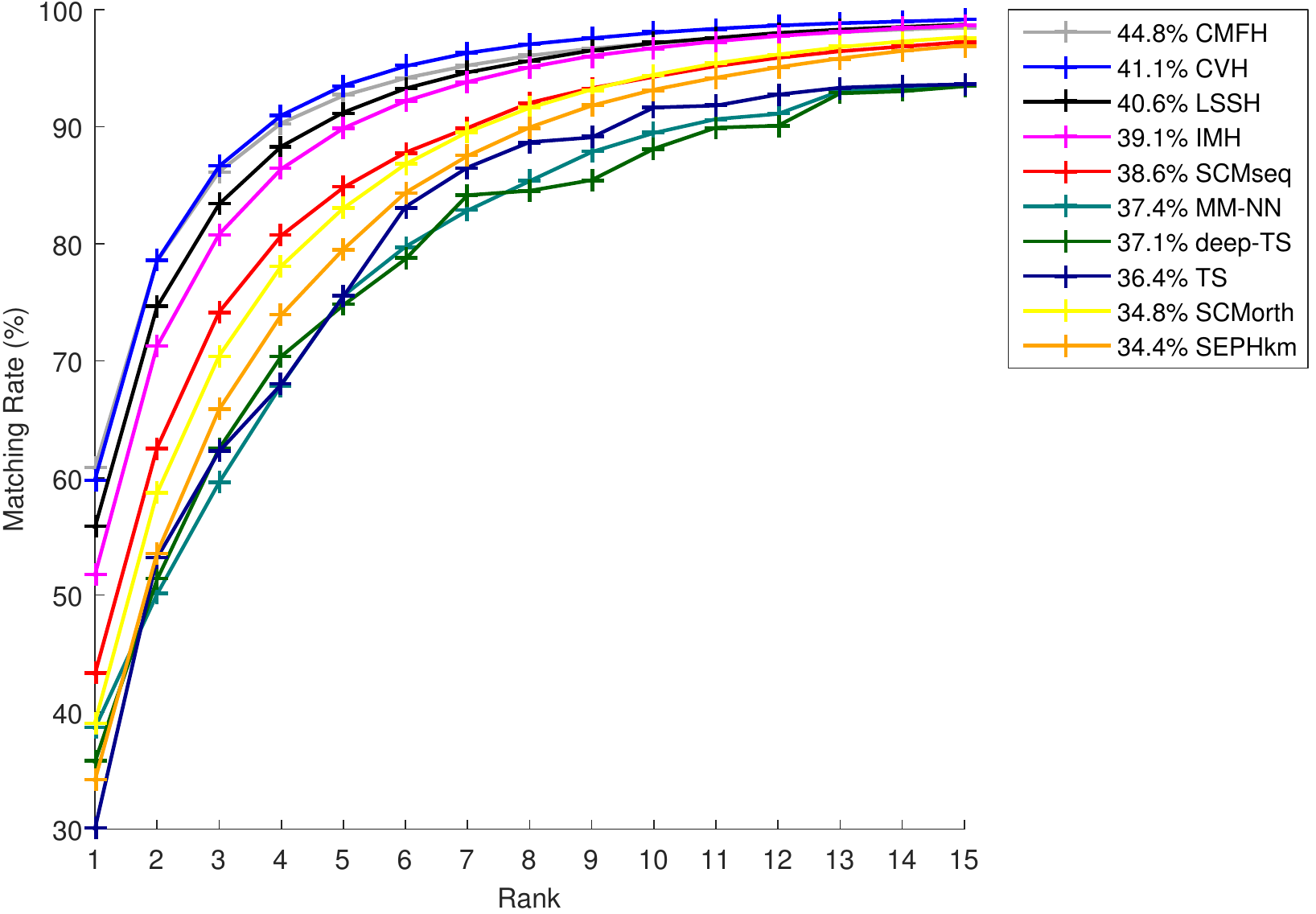}}
\end{center}
	\caption{Evaluation results of binary representations on Pascal Sentence. CMC curves are shown. MAP is shown before the name of each method. Retrieval modes in (a)(b)(c)(d)(e)(f)(g)(h)(i)(j)(k)(l) are the same with Fig. \ref{fig:hsWiki} and Fig. \ref{fig:hsPascal}.}
	\label{fig:hsNUSWide}
\end{figure*}

The binary methods include:
\begin{itemize}
\item CVH: cross-view hashing~\cite{kumar2011learning} is the extension of spectral hashing~\cite{weiss2009spectral} to the cross-media case.
\item SCMseq: the sequential learning of semantic correlation maximization hashing (SCM)~\cite{zhang2014large} aims to make the distances of the hash codes equal to the similarities of the label vectors and uses a sequential learning algorithm to learn the hash codes.
\item SCMorth: the orthogonal projection learning of SCM~\cite{zhang2014large} is the same as SCMseq, but uses an orthogonal projection learning algorithm to learn the hash codes.
\item CMFH: collective matrix factorization hashing~\cite{ding2014collective} is based on CCA and learns hash codes with latent factor model from different modalities.
\item LSSH: latent semantic sparse hashing~\cite{zhou2014latent} is based on CCA and captures high-level semantic information, \ie, sparse coding and matrix factorization for images and texts respectively.
\item SEPHkm: semantics-preserving hashing (SEPH)~\cite{song2013inter} + k-means transforms the semantic affinities of the training data into a probability distribution and approximates it with to-be-learned hash codes via minimizing the Kullback-Leibler divergence.
\item IMH: inter-media hashing~\cite{song2013inter} is another extension method of spectral hashing~\cite{weiss2009spectral}.
\item MM-NN: multimodal similarity-preserving hashing~\cite{ma2015multimodal} is based on a coupled Siamese neural network.
\end{itemize}

For the real-valued representations, the CNN and BoW features described in Section \ref{sec:exp_features} are projected onto a $K$-dim subspace, where $K$ denotes the number of training classes. For the binary representations, the CNN and BoW features are binarized to 8, 16 and 32 bits for evaluation.

\subsection{Trivial Solution}

We compare the trivial solution \cite{sablayrolles2016should} under the non-extendable and extendable retrieval settings. In a nutshell, the trivial solution learns a classifier in the training set. Then, the classifier assigns a class label to the gallery and query images/texts. The system thus returns those images/texts of the same class with the query. Note that, since there are a number of images/texts predicted into the same class, \ie, these images/texts form ties, and we randomly assign them a rank in the tie.

\begin{itemize}
\item TS: trivial solution~\cite{sablayrolles2016should} uses the multi-class logistic regression to learn classifiers and uses the classification results as the retrieval results. 
\item deep-TS: deep trivial solution~\cite{sablayrolles2016should} works in a similar manner with TS, but uses deep networks to learn the classifiers. The training process is the same with deep-SM~\cite{wei2016cross}.
\end{itemize}

We compare TS and deep-TS with both the real-valued and binary representation methods to evaluate the performance of these cross-media methods. Note that for NUS-WIDE and Wikipedia where 10 classes are used for training and testing, TS and deep-TS only consume less than 4 bits under the new protocol (5 class for training); for Pascal Sentence with 20 classes, TS and deep-TS only cost less than 5 bits under the new protocol (10 classes for training).

\subsection{Evaluation on Real-valued Methods}

We first present CMC curves and MAP scores obtained by the real-valued baseline methods detailed in Section \ref{sec:exp_baseline}. The results of the trivial solutions (TS and deep-TS) are also drawn.
Note that in the train/test splitting (Fig. \ref{fig:DatasetSeperation}), we ensure that the non-extendable retrieval and extendable retrieval have the same number of training classes and roughly the same number of training samples. So their numbers are directly comparable.
The results on Wikipedia, Pascal Sentence, and NUS-WIDE are shown in Fig. \ref{fig:rlWiki}, Fig. \ref{fig:rlPascal} and Fig. \ref{fig:rlNUSWide}, respectively. From these results, we arrive at three major conclusions.

(1) \textbf{The CMC and MAP scores of non-extendable retrieval are higher than those of extendable retrieval.}  For example, MAP of semantic matching (SM) \cite{rasiwasia2010new} decreases from 60.7\% (Fig. \ref{subfig:wiki_i2t1_rl}) to 29.4\% (Fig. \ref{subfig:wiki_i2t2_rl}), and its rank-1 accuracy on the CMC curve drops from 56.5\% (Fig. \ref{subfig:wiki_i2t1_rl}) to 28.8\% (Fig. \ref{subfig:wiki_i2t2_rl}). A similarly considerable performance drop can be seen of other methods such as correlation matching (CM) \cite{rasiwasia2010new}, bilinear model (BLM) \cite{turk1991eigenfaces}, \emph{etc.}
This observation is expected because the distributions of the testing data are more different from the training data in extendable retrieval. In comparison, for the non-extendable retrieval, the training and testing data come from very similar distributions because they share the same set of classes. We therefore speculate that testing on data of another domain is a challenging and meaningful task since the benchmarking methods behave poorly compared to their performance of non-extendable retrieval. Regarding this point, we think that transfer learning should be an effective strategy in the future. 

(2) \textbf{While the trivial solution yields competitive accuracy under the non-extendable retrieval settings, its advantage is significantly reduced under the extendable retrieval settings.} For example, MAP rank of deep trivial solution (deep-TS) \cite{sablayrolles2016should} drops from 3 (59.8\% in Fig. \ref{subfig:wiki_i2t1_rl}) to 8 (23.6\% in Fig. \ref{subfig:wiki_i2t2_rl}). Similarly, trivial solution (TS) \cite{sablayrolles2016should} drops from 4 (54.5\% in Fig. \ref{subfig:wiki_i2t1_rl}) to 7 (25.3\% in Fig. \ref{subfig:wiki_i2t2_rl}). The possible reason is that TS and deep-TS are based on classification and they fit the class distribution tightly. In extendable retrieval it is quite possible that the training and testing data come from very dissimilar class distributions. Then, the learned classifiers have a big chance to misclassify the testing data. We therefore think that there is no trivial solution in extendable retrieval. In Fig. \ref{fig:retrieval_res} we present some sample query results of deep semantic matching (deep-SM) \cite{wei2016cross} and deep trivial solution \cite{sablayrolles2016should} (deep-TS) in both non-extendable retrieval and extendable retrieval. The performance drop of deep-TS is much more obvious.

(3) \textbf{The performance of different methods \emph{w.r.t} the CMC curves is not consistent with MAP, but the rank-1 accuracy is somehow positively related with MAP.} For example, correlation matching (CM) \cite{rasiwasia2010new} shows the second best performance \emph{w.r.t} its CMC curve, but it has the lowest MAP (21.6\% in Fig. \ref{subfig:wiki_i2t1_rl}). The rank-1 accuracy of CM is the third lowest and is somehow consistent with its MAP. Similar situation is observed for other methods such as GMLDA \cite{sharma2012generalized} and bilinear model (BLM) \cite{turk1991eigenfaces}. The difference is caused by the definition of the two evaluation metrics. MAP is a global metric and it averages the precision of the whole matches, its score reflects the distribution of the matching documents in the returned list. A high MAP can be gotten even if most matches rank medially in the returned list. However, CMC curve only counts the first match of each query and reflects the possibility to find the (first) match at each rank. We think the two metrics reflect different aspects of performance, which are complementary to each other.

\subsection{Evaluation on Binary Representations}

We evaluate the binary methods with three code lengths: 8, 16 and 32. The experimental results on the three datasets are illustrated in Figures~\ref{fig:hsWiki},~\ref{fig:hsPascal} and~\ref{fig:hsNUSWide}, respectively. The numbers in the brackets denote the code lengths.

The findings obtained from the real-valued representations still hold for the binary cases. Here we add some additional observations specifically observed for cross-media hashing.

(1) \textbf{As code length increases, the CMC curves and MAP both undergo improvement for most methods, and vice versa.} The variations of the CMC curves are coincident with the changes of the MAP scores between different code lengths. For example, MAP of multimodal similarity-preserving hashing (MM-NN) \cite{ma2015multimodal} improves from $ 60.8\% $ (Fig. \ref{subfig:wiki_i2t1_hs8}) to $ 62.8\% $ (Fig. \ref{subfig:wiki_i2t1_hs16}), the CMC curve also has improvement in accuracy. Exceptions exits. For example, MAP of SCMorth \cite{zhang2014large} drops from $ 41.7\% $ (Fig. \ref{subfig:wiki_t2i1_hs8}) to $ 29.3\% $ (Fig. \ref{subfig:wiki_t2i1_hs16}) but CMC curve is improved. A similar situation is observed for other methods such as inter-media hashing (IMH) \cite{song2013inter} and cross-view hashing (CVH) \cite{kumar2011learning}. The reason is that these methods are based on CCA \cite{hotelling1936relations} and produce limited effective bits for the global accuracy (MAP). We think that a longer code improves the top-rank accuracy in general.

(2) \textbf{The CMC curve may serve as a good discriminator between methods under the cases of similar MAP scores.} The CMC curve can illustrate the differences of performance even when with equal MAP scores. For example, MAP of CVH is similar in Figs. \ref{subfig:wiki_t2i1_hs16} (21.6\%), \ref{subfig:wiki_t2i1_hs16} (21.7\%) and \ref{subfig:wiki_t2i1_hs32} (21.9\%), but the CMC curves are more discriminative. The reason is that, compared to MAP, CMC curve reflects more details of the search results. Regard this point, and we think that CMC curve is a good supplementary metric for MAP.

(3) \textbf{In most cases, the difference in performance (CMC and MAP) of various methods is much smaller under the new protocol.} For example, in Fig. \ref{subfig:wiki_i2t2_hs8}, the difference of CMC and MAP between the methods is much smaller compared to Fig. \ref{subfig:wiki_i2t1_hs8}. Similar situation is observed for both the real-valued and binary representation methods. This observation is expected because the discriminative knowledge learned from the training data cannot be directly transfered to the testing data under the new protocol. The drops of performance reduce the difference between the methods. We think the next challenge of cross-media retrieval is to solve the knowledge transfer problem.

\section{Conclusion}
This paper introduces a new evaluation protocol and extensive benchmarking results for extendable cross-media retrieval. The new protocol involves 1) a complete separation of the training the testing sets and 2) a complete separation of the training and testing classes. This protocol thus reflects the extendable settings that have been largely ignored in the cross-media retrieval community. Through the benchmarking results, we demonstrate a significant performance drop from the non-extendable retrieval settings to the extendable retrieval settings. Moreover, we find that the classification trivial solution works under the non-extendable retrieval but is less effective in the extendable retrieval. These observations indicate that the two evaluation protocols (the existing protocol and the new protocol) are entirely different, and the practical usage seems to favor the new protocol through our analysis. 

In the future, we point out two critical research directions apart from the common feature/subspace learning techniques:

\textbf{First, transfer learning within each dataset should be proposed.} The model trained from a part of dataset is expected to be effective for another part of data with different class distribution, both non-extendable retrieval and extendable retrieval should be evaluated.

\textbf{Second, generically applicable models that work on different cross-media datasets are to be investigated.} It is expected that a generic model is trained on a large-scale training dataset and the learned model is effective in all the other testing datasets.  

\ifCLASSOPTIONcaptionsoff
  \newpage
\fi

\bibliographystyle{IEEEtran}
\bibliography{IEEEabrv,bare_jrnl}

\begin{thebibliography}{10}
\providecommand{\url}[1]{#1}
\csname url@samestyle\endcsname
\providecommand{\newblock}{\relax}
\providecommand{\bibinfo}[2]{#2}
\providecommand{\BIBentrySTDinterwordspacing}{\spaceskip=0pt\relax}
\providecommand{\BIBentryALTinterwordstretchfactor}{4}
\providecommand{\BIBentryALTinterwordspacing}{\spaceskip=\fontdimen2\font plus
\BIBentryALTinterwordstretchfactor\fontdimen3\font minus
  \fontdimen4\font\relax}
\providecommand{\BIBforeignlanguage}[2]{{%
\expandafter\ifx\csname l@#1\endcsname\relax
\typeout{** WARNING: IEEEtran.bst: No hyphenation pattern has been}%
\typeout{** loaded for the language `#1'. Using the pattern for}%
\typeout{** the default language instead.}%
\else
\language=\csname l@#1\endcsname
\fi
#2}}
\providecommand{\BIBdecl}{\relax}
\BIBdecl

\bibitem{zheng2014packing}
L.~Zheng, S.~Wang, Z.~Liu, and Q.~Tian, ``Packing and padding: Coupled
  multi-index for accurate image retrieval,'' in \emph{CVPR}, 2014, pp.
  1939--1946.

\bibitem{zheng2016accurate}
L.~Zheng, S.~Wang, J.~Wang, and Q.~Tian, ``Accurate image search with
  multi-scale contextual evidences,'' \emph{IJCV}, vol. 120, no.~1, pp. 1--13,
  2016.

\bibitem{zheng2016sift}
L.~Zheng, Y.~Yang, and Q.~Tian, ``Sift meets cnn: a decade survey of instance
  retrieval,'' \emph{arXiv preprint arXiv:1608.01807}, 2016.

\bibitem{song2013effective}
J.~Song, Y.~Yang, Z.~Huang, H.~T. Shen, and J.~Luo, ``Effective multiple
  feature hashing for large-scale near-duplicate video retrieval,'' \emph{TMM},
  vol.~15, no.~8, pp. 1997--2008, 2013.

\bibitem{wu2005understanding}
F.~Wu, Y.~Yang, Y.~Zhuang, and Y.~Pan, ``Understanding multimedia document
  semantics for cross-media retrieval,'' in \emph{PCM}.\hskip 1em plus 0.5em
  minus 0.4em\relax Springer Berlin Heidelberg, 2005, pp. 993--1004.

\bibitem{zhuang2008mining}
Y.-T. Zhuang, Y.~Yang, and F.~Wu, ``Mining semantic correlation of
  heterogeneous multimedia data for cross-media retrieval,'' \emph{TMM},
  vol.~10, no.~2, pp. 221--229, 2008.

\bibitem{yang2008harmonizing}
Y.~Yang, Y.-T. Zhuang, F.~Wu, and Y.-H. Pan, ``Harmonizing hierarchical
  manifolds for multimedia document semantics understanding and cross-media
  retrieval,'' \emph{TMM}, vol.~10, no.~3, pp. 437--446, 2008.

\bibitem{yang2009ranking}
Y.~Yang, D.~Xu, F.~Nie, J.~Luo, and Y.~Zhuang, ``Ranking with local regression
  and global alignment for cross media retrieval,'' in \emph{ACMMM}.\hskip 1em
  plus 0.5em minus 0.4em\relax ACM, 2009, pp. 175--184.

\bibitem{yang2010cross}
Y.~Yang, F.~Wu, D.~Xu, Y.~Zhuang, and L.-T. Chia, ``Cross-media retrieval using
  query dependent search methods,'' \emph{Pattern Recognition}, vol.~43, no.~8,
  pp. 2927--2936, 2010.

\bibitem{rasiwasia2010new}
N.~Rasiwasia, J.~Costa~Pereira, E.~Coviello, G.~Doyle, G.~R. Lanckriet,
  R.~Levy, and N.~Vasconcelos, ``A new approach to cross-modal multimedia
  retrieval,'' in \emph{ACMMM}.\hskip 1em plus 0.5em minus 0.4em\relax ACM,
  2010, pp. 251--260.

\bibitem{hotelling1936relations}
H.~Hotelling, ``Relations between two sets of variates,'' \emph{Biometrika},
  vol.~28, no. 3/4, pp. 321--377, 1936.

\bibitem{sharma2012generalized}
A.~Sharma, A.~Kumar, H.~Daume, and D.~W. Jacobs, ``Generalized multiview
  analysis: A discriminative latent space,'' in \emph{CVPR}.\hskip 1em plus
  0.5em minus 0.4em\relax IEEE, 2012, pp. 2160--2167.

\bibitem{wang2013learning}
K.~Wang, R.~He, W.~Wang, L.~Wang, and T.~Tan, ``Learning coupled feature spaces
  for cross-modal matching,'' in \emph{ICCV}, 2013, pp. 2088--2095.

\bibitem{gong2014multi}
Y.~Gong, Q.~Ke, M.~Isard, and S.~Lazebnik, ``A multi-view embedding space for
  modeling internet images, tags, and their semantics,'' \emph{IJCV}, vol. 106,
  no.~2, pp. 210--233, 2014.

\bibitem{rasiwasia2014cluster}
N.~Rasiwasia, D.~Mahajan, V.~Mahadevan, and G.~Aggarwal, ``Cluster canonical
  correlation analysis.'' in \emph{AISTATS}, 2014, pp. 823--831.

\bibitem{andrew2013deep}
G.~Andrew, R.~Arora, J.~A. Bilmes, and K.~Livescu, ``Deep canonical correlation
  analysis.'' in \emph{ICML}, 2013, pp. 1247--1255.

\bibitem{wang2016learning}
L.~Wang, Y.~Li, and S.~Lazebnik, ``Learning deep structure-preserving
  image-text embeddings,'' in \emph{CVPR}, 2016, pp. 5005--5013.

\bibitem{yan2015deep}
F.~Yan and K.~Mikolajczyk, ``Deep correlation for matching images and text,''
  in \emph{CVPR}, 2015, pp. 3441--3450.

\bibitem{wei2016cross}
Y.~Wei, Y.~Zhao, C.~Lu, S.~Wei, L.~Liu, Z.~Zhu, and S.~Yan, ``Cross-modal
  retrieval with cnn visual features: A new baseline,'' \emph{IEEE Trans. on
  Cybernetics}, 2016, {P}reprint.

\bibitem{he2016cross}
Y.~He, S.~Xiang, C.~Kang, J.~Wang, and C.~Pan, ``Cross-modal retrieval via deep
  and bidirectional representation learning,'' \emph{TMM}, vol.~18, no.~7, pp.
  1363--1377, 2016.

\bibitem{socher2014grounded}
R.~Socher, A.~Karpathy, Q.~V. Le, C.~D. Manning, and A.~Y. Ng, ``Grounded
  compositional semantics for finding and describing images with sentences,''
  \emph{TACL}, vol.~2, pp. 207--218, 2014.

\bibitem{kiros2014unifying}
R.~Kiros, R.~Salakhutdinov, and R.~S. Zemel, ``Unifying visual-semantic
  embeddings with multimodal neural language models,'' \emph{arXiv preprint
  arXiv:1411.2539}, 2014.

\bibitem{feng2014cross}
F.~Feng, X.~Wang, and R.~Li, ``Cross-modal retrieval with correspondence
  autoencoder,'' in \emph{ACMMM}.\hskip 1em plus 0.5em minus 0.4em\relax ACM,
  2014, pp. 7--16.

\bibitem{vukotic2016bidirectional}
V.~Vukoti{\'c}, C.~Raymond, and G.~Gravier, ``Bidirectional joint
  representation learning with symmetrical deep neural networks for multimodal
  and crossmodal applications,'' in \emph{ICMR}.\hskip 1em plus 0.5em minus
  0.4em\relax ACM, 2016, pp. 343--346.

\bibitem{park2016image}
G.~Park and W.~Im, ``Image-text multi-modal representation learning by
  adversarial backpropagation,'' \emph{arXiv preprint arXiv:1612.08354}, 2016.

\bibitem{kumar2011learning}
S.~Kumar and R.~Udupa, ``Learning hash functions for cross-view similarity
  search,'' in \emph{IJCAI}, vol.~22, no.~1, 2011, pp. 1360--1365.

\bibitem{song2013inter}
J.~Song, Y.~Yang, Y.~Yang, Z.~Huang, and H.~T. Shen, ``Inter-media hashing for
  large-scale retrieval from heterogeneous data sources,'' in
  \emph{SIGMOD}.\hskip 1em plus 0.5em minus 0.4em\relax ACM, 2013, pp.
  785--796.

\bibitem{ding2014collective}
G.~Ding, Y.~Guo, and J.~Zhou, ``Collective matrix factorization hashing for
  multimodal data,'' in \emph{CVPR}, 2014, pp. 2075--2082.

\bibitem{zhang2014large}
D.~Zhang and W.-J. Li, ``Large-scale supervised multimodal hashing with
  semantic correlation maximization.'' in \emph{AAAI}, vol.~1, no.~2, 2014,
  p.~7.

\bibitem{zhou2014latent}
J.~Zhou, G.~Ding, and Y.~Guo, ``Latent semantic sparse hashing for cross-modal
  similarity search,'' in \emph{SIGIR}.\hskip 1em plus 0.5em minus 0.4em\relax
  ACM, 2014, pp. 415--424.

\bibitem{yu2014discriminative}
Z.~Yu, F.~Wu, Y.~Yang, Q.~Tian, J.~Luo, and Y.~Zhuang, ``Discriminative coupled
  dictionary hashing for fast cross-media retrieval,'' in \emph{SIGIR}.\hskip
  1em plus 0.5em minus 0.4em\relax ACM, 2014, pp. 395--404.

\bibitem{xu2016discriminant}
X.~Xu, F.~Shen, Y.~Yang, and H.~T. Shen, ``Discriminant cross-modal hashing,''
  in \emph{ICMR}.\hskip 1em plus 0.5em minus 0.4em\relax ACM, 2016, pp.
  305--308.

\bibitem{bronstein2010data}
M.~M. Bronstein, A.~M. Bronstein, F.~Michel, and N.~Paragios, ``Data fusion
  through cross-modality metric learning using similarity-sensitive hashing,''
  in \emph{CVPR}.\hskip 1em plus 0.5em minus 0.4em\relax IEEE, 2010, pp.
  3594--3601.

\bibitem{zhen2012probabilistic}
Y.~Zhen and D.-Y. Yeung, ``A probabilistic model for multimodal hash function
  learning,'' in \emph{SIGKDD}.\hskip 1em plus 0.5em minus 0.4em\relax ACM,
  2012, pp. 940--948.

\bibitem{lin2015semantics}
Z.~Lin, G.~Ding, M.~Hu, and J.~Wang, ``Semantics-preserving hashing for
  cross-view retrieval,'' in \emph{CVPR}, 2015, pp. 3864--3872.

\bibitem{masci2014multimodal}
J.~Masci, M.~M. Bronstein, A.~M. Bronstein, and J.~Schmidhuber, ``Multimodal
  similarity-preserving hashing,'' \emph{TPAMI}, vol.~36, no.~4, pp. 824--830,
  2014.

\bibitem{jiang2016deep}
Q.-Y. Jiang and W.-J. Li, ``Deep cross-modal hashing,'' \emph{arXiv preprint
  arXiv:1602.02255}, 2016.

\bibitem{cao2016deep}
Y.~Cao, M.~Long, J.~Wang, Q.~Yang, and P.~S. Yu, ``Deep visual-semantic hashing
  for cross-modal retrieval,'' in \emph{SIGKDD}.\hskip 1em plus 0.5em minus
  0.4em\relax ACM, 2016, pp. 1445--1454.

\bibitem{frome2013devise}
A.~Frome, G.~S. Corrado, J.~Shlens, S.~Bengio, J.~Dean, T.~Mikolov
  \emph{et~al.}, ``Devise: A deep visual-semantic embedding model,'' in
  \emph{NIPS}, 2013, pp. 2121--2129.

\bibitem{mao2014explain}
J.~Mao, W.~Xu, Y.~Yang, J.~Wang, and A.~L. Yuille, ``Explain images with
  multimodal recurrent neural networks,'' \emph{arXiv preprint
  arXiv:1410.1090}, 2014.

\bibitem{karpathy2014deep}
A.~Karpathy, A.~Joulin, and F.~F. Li, ``Deep fragment embeddings for
  bidirectional image sentence mapping,'' in \emph{NIPS}, 2014, pp. 1889--1897.

\bibitem{ma2015multimodal}
L.~Ma, Z.~Lu, L.~Shang, and H.~Li, ``Multimodal convolutional neural networks
  for matching image and sentence,'' in \emph{ICCV}, 2015, pp. 2623--2631.

\bibitem{chen2015mind}
X.~Chen and C.~Lawrence~Zitnick, ``Mind's eye: A recurrent visual
  representation for image caption generation,'' in \emph{CVPR}, 2015, pp.
  2422--2431.

\bibitem{karpathy2015deep}
A.~Karpathy and L.~Fei-Fei, ``Deep visual-semantic alignments for generating
  image descriptions,'' in \emph{CVPR}, 2015, pp. 3128--3137.

\bibitem{chua2009nus}
T.-S. Chua, J.~Tang, R.~Hong, H.~Li, Z.~Luo, and Y.~Zheng, ``Nus-wide: a
  real-world web image database from national university of singapore,'' in
  \emph{CIVR}.\hskip 1em plus 0.5em minus 0.4em\relax ACM, 2009, pp. 1--9.

\bibitem{huiskes2008mir}
M.~J. Huiskes and M.~S. Lew, ``The mir flickr retrieval evaluation,'' in
  \emph{MIR}.\hskip 1em plus 0.5em minus 0.4em\relax ACM, 2008, pp. 39--43.

\bibitem{everingham2010pascal}
M.~Everingham, L.~Van~Gool, C.~K. Williams, J.~Winn, and A.~Zisserman, ``The
  pascal visual object classes (voc) challenge,'' \emph{IJCV}, vol.~88, no.~2,
  pp. 303--338, 2010.

\bibitem{krapac2010improving}
J.~Krapac, M.~Allan, J.~Verbeek, and F.~Juried, ``Improving web image search
  results using query-relative classifiers.''

\bibitem{rashtchian2010collecting}
C.~Rashtchian, P.~Young, M.~Hodosh, and J.~Hockenmaier, ``Collecting image
  annotations using amazon's mechanical turk,'' in \emph{NAACL Workshop}.\hskip
  1em plus 0.5em minus 0.4em\relax Association for Computational Linguistics,
  2010, pp. 139--147.

\bibitem{yan2016image}
Y.~Yan, F.~Nie, W.~Li, C.~Gao, Y.~Yang, and D.~Xu, ``Image classification by
  cross-media active learning with privileged information,'' \emph{TMM},
  vol.~18, no.~12, pp. 2494--2502, 2016.

\bibitem{hodosh2013framing}
M.~Hodosh, P.~Young, and J.~Hockenmaier, ``Framing image description as a
  ranking task: Data, models and evaluation metrics,'' \emph{JAIR}, vol.~47,
  pp. 853--899, 2013.

\bibitem{young2014image}
P.~Young, A.~Lai, M.~Hodosh, and J.~Hockenmaier, ``From image descriptions to
  visual denotations: New similarity metrics for semantic inference over event
  descriptions,'' \emph{TACL}, vol.~2, pp. 67--78, 2014.

\bibitem{lin2014microsoft}
T.-Y. Lin, M.~Maire, S.~Belongie, J.~Hays, P.~Perona, D.~Ramanan,
  P.~Doll{\'a}r, and C.~L. Zitnick, ``Microsoft coco: Common objects in
  context,'' in \emph{ECCV}.\hskip 1em plus 0.5em minus 0.4em\relax Springer,
  2014, pp. 740--755.

\bibitem{ordonez2011im2text}
V.~Ordonez, G.~Kulkarni, and T.~L. Berg, ``Im2text: Describing images using 1
  million captioned photographs,'' in \emph{NIPS}, 2011, pp. 1143--1151.

\bibitem{zhai2014learning}
X.~Zhai, Y.~Peng, and J.~Xiao, ``Learning cross-media joint representation with
  sparse and semisupervised regularization,'' \emph{TCSVT}, vol.~24, no.~6, pp.
  965--978, 2014.

\bibitem{sablayrolles2016should}
A.~Sablayrolles, M.~Douze, H.~J{\'e}gou, and N.~Usunier, ``How should we
  evaluate supervised hashing?'' \emph{arXiv preprint arXiv:1609.06753}, 2016.

\bibitem{radenovic2016cnn}
F.~Radenovi{\'c}, G.~Tolias, and O.~Chum, ``Cnn image retrieval learns from
  bow: Unsupervised fine-tuning with hard examples,'' in \emph{ECCV}.\hskip 1em
  plus 0.5em minus 0.4em\relax Springer, 2016, pp. 3--20.

\bibitem{zheng2015scalable}
L.~Zheng, L.~Shen, L.~Tian, S.~Wang, J.~Wang, and Q.~Tian, ``Scalable person
  re-identification: A benchmark,'' in \emph{ICCV}, 2015, pp. 1116--1124.

\bibitem{zheng2016person}
L.~Zheng, Y.~Yang, and A.~G. Hauptmann, ``Person re-identification: Past,
  present and future,'' \emph{arXiv preprint arXiv:1610.02984}, 2016.

\bibitem{liu2016deep}
X.~Liu, W.~Liu, T.~Mei, and H.~Ma, ``A deep learning-based approach to
  progressive vehicle re-identification for urban surveillance,'' in
  \emph{ECCV}.\hskip 1em plus 0.5em minus 0.4em\relax Springer, 2016, pp.
  869--884.

\bibitem{liu2016large}
X.~Liu, W.~Liu, H.~Ma, and H.~Fu, ``Large-scale vehicle re-identification in
  urban surveillance videos,'' in \emph{ICME}, 2016, pp. 1--6.

\bibitem{jia2014caffe}
Y.~Jia, E.~Shelhamer, J.~Donahue, S.~Karayev, J.~Long, R.~Girshick,
  S.~Guadarrama, and T.~Darrell, ``Caffe: Convolutional architecture for fast
  feature embedding,'' 2014.

\bibitem{krizhevsky2012imagenet}
A.~Krizhevsky, I.~Sutskever, and G.~E. Hinton, ``Imagenet classification with
  deep convolutional neural networks,'' in \emph{NIPS}, 2012, pp. 1097--1105.

\bibitem{deng2009imagenet}
J.~Deng, W.~Dong, R.~Socher, L.-J. Li, K.~Li, and L.~Fei-Fei, ``Imagenet: A
  large-scale hierarchical image database,'' in \emph{CVPR}.\hskip 1em plus
  0.5em minus 0.4em\relax IEEE, 2009, pp. 248--255.

\bibitem{mikolov2013distributed}
T.~Mikolov, I.~Sutskever, K.~Chen, G.~S. Corrado, and J.~Dean, ``Distributed
  representations of words and phrases and their compositionality,'' in
  \emph{NIPS}, 2013, pp. 3111--3119.

\bibitem{sharma2011bypassing}
A.~Sharma and D.~W. Jacobs, ``Bypassing synthesis: Pls for face recognition
  with pose, low-resolution and sketch,'' in \emph{CVPR}.\hskip 1em plus 0.5em
  minus 0.4em\relax IEEE, 2011, pp. 593--600.

\bibitem{turk1991eigenfaces}
M.~Turk and A.~Pentland, ``Eigenfaces for recognition,'' \emph{JOCN}, vol.~3,
  no.~1, pp. 71--86, 1991.

\bibitem{yan2007graph}
S.~Yan, D.~Xu, B.~Zhang, H.-J. Zhang, Q.~Yang, and S.~Lin, ``Graph embedding
  and extensions: A general framework for dimensionality reduction,''
  \emph{TPAMI}, vol.~29, no.~1, 2007.

\bibitem{belhumeur1997eigenfaces}
P.~N. Belhumeur, J.~P. Hespanha, and D.~J. Kriegman, ``Eigenfaces vs.
  fisherfaces: Recognition using class specific linear projection,''
  \emph{TPAMI}, vol.~19, no.~7, pp. 711--720, 1997.

\bibitem{weiss2009spectral}
Y.~Weiss, A.~Torralba, and R.~Fergus, ``Spectral hashing,'' in \emph{NIPS},
  2009, pp. 1753--1760.

\end{thebibliography}

\end{document}